\documentclass[acmsmall,manuscript,nonacm]{acmart}
\setcopyright{none}

\usepackage{mathtools}
\usepackage{algorithmic}
\usepackage{enumitem}
\usepackage{tabularx}
\usepackage[longtable]{multirow}
\usepackage{caption}
\usepackage{textcomp}
\usepackage{xcolor}
\usepackage{makecell}
\usepackage{siunitx}
\usepackage{algorithm}
\usepackage[font=footnotesize]{subcaption}
\usepackage[flushleft]{threeparttable} %
\usepackage{rotating}
\usepackage{pdflscape}
\usepackage{ltablex}

\AtBeginDocument{%
  }

\begin{document}

\title[Federated Learning Survey]{Federated Learning Survey: A Multi-Level Taxonomy of Aggregation Techniques, Experimental Insights, and Future Frontiers}
\titlenote{Author-Accepted Manuscript (AAM). Published in ACM Transactions on Intelligent Systems and Technology (TIST). DOI: 10.1145/3678182.}
\author{Meriem ARBAOUI}

\affiliation{
  \institution{LabRi-SBA Laboratory, Algeria \& CESI LINEACT UR 7527}
  \streetaddress{}
  \city{Strasbourg}
  \country{France}
}
\email{m.arbaoui@esi-sba.dz}
\email{marbaoui@cesi.fr}
\orcid{0009-0002-1896-8637}

\author{Mohamed-el-Amine BRAHMIA}
\affiliation{%
  \institution{CESI LINEACT UR 7527}
  \streetaddress{}
  \city{Strasbourg}
  \country{France}}
\email{abrahmia@cesi.fr}
\orcid{0000-0003-0114-210X}

\author{Abdellatif RAHMOUN}
\affiliation{%
  \institution{LabRi-SBA Laboratory}
    \streetaddress{}
  \city{Sidi Bel-Abbes}
  \country{Algeria}  
}
\email{a.rahmoun@esi-sba.dz}
\orcid{0000-0002-8227-3322}

\author{Mourad ZGHAL}
\affiliation{%
  \institution{CESI LINEACT UR 7527}
  \streetaddress{}
  \city{Strasbourg}
  \country{France}}
\email{mzghal@cesi.fr}
\orcid{0000-0003-1329-8574}
\renewcommand{\shortauthors}{Arbaoui et al.}
\begin{abstract}
The emerging integration of IoT (Internet of Things) and AI (Artificial Intelligence) has unlocked numerous opportunities for innovation across diverse industries. However, growing privacy concerns and data isolation issues have inhibited this promising advancement. Unfortunately, traditional centralized machine learning (ML) methods have demonstrated their limitations in addressing these hurdles. In response to this ever-evolving landscape, Federated Learning (FL) has surfaced as a cutting-edge machine learning paradigm, enabling collaborative training across decentralized devices. FL allows users to jointly construct AI models without sharing their local raw data, ensuring data privacy, network scalability, and minimal data transfer. One essential aspect of FL revolves around proficient knowledge aggregation within a heterogeneous environment. Yet, the inherent characteristics of FL have amplified the complexity of its practical implementation compared to centralized ML. This survey delves into three prominent clusters of FL research contributions: personalization, optimization, and robustness. The objective is to provide a well-structured and fine-grained classification scheme related to these research areas through a unique methodology for selecting related work. Unlike other survey papers, we employed a hybrid approach that amalgamates bibliometric analysis and systematic scrutinizing to find the most influential work in the literature. Therefore, we examine challenges and contemporary techniques related to heterogeneity, efficiency, security, and privacy. Another valuable asset of this study is its comprehensive coverage of FL aggregation strategies, encompassing architectural features, synchronization methods, and several federation motivations. To further enrich our investigation, we provide practical insights into evaluating novel FL proposals and conduct experiments to assess and compare aggregation methods under IID and non-IID data distributions. Finally, we present a compelling set of research avenues that call for further exploration to open up a treasure of advancement.  
\end{abstract}


\maketitle

\section{Introduction}
\label{intro}
In today's data-centric world, the widespread adoption of IoT devices has led to tremendous generated data, enabling access to intelligent and high-quality services. This wealth of data has fueled an unprecedented AI expansion across countless application domains. Specifically, it is essential to feed substantial amounts of data into Deep Learning (DL) models to achieve impressive accuracy results, paving the way for advanced services development \cite{pandya2023federated}. Historically, the storage and analysis of such massive data have been entrusted to the Cloud, owing to its immense capacities. However, the considerable drawbacks of centralized machine learning techniques, which rely on cloud-only-based solutions, have become evident in the face of sophisticated human needs.

The first limitation in this context stems from the undesirable processing latency introduced when offloading massive IoT data to remote servers, as the location of these data centers is typically far from data owners. As a result, this long-distance communication architecture incurs high computation costs \cite{wahab2021federated}. The second problem originates from the need for users to sacrifice their data privacy in exchange for improved AI services \cite{imteaj2021survey}. While data holders are increasingly becoming wary about sharing their data with third parties, regardless of their trustworthiness and reputation. This growing awareness has prompted organizations and governments to implement strict privacy regulations to safeguard data ownership and control.

For instance, within the European Commission’s General Data Protection Regulation (GDPR), articles 5 and 6 introduce two ethical concepts: \textit{data minimization} and \textit{purpose limitation}. The former stresses collecting only relevant and meaningful data for a study, while the latter restricts using collected data for future research purposes beyond its original intent. In order to comply with these legal legislations, researchers have employed pseudonymization or de-identification techniques, which involve removing identifiable details such as names, addresses, and social security numbers from collected datasets and replacing them with pseudonyms \cite{pfitzner2021federated}. Nonetheless, it is worth noting that these approaches may not provide absolute privacy protection. In some scenarios, re-identification mechanisms can potentially link the pseudonyms back to their associated entities, compromising the privacy of individuals \cite{culnane2017health, rocher2019estimating}.

In light of these concerns, federated learning emerges as an innovative solution to distribute the computational workload of training ML models across multiple nodes, while ensuring the data privacy of locally held data at each site \cite{mcmahan2017communication}. In other words, federated learning allows participating nodes to collaboratively train a robust AI model by harnessing the collective knowledge within their local data without uploading it to a remote server, as in centralized machine learning. In the typical FL setup, this paradigm operates with a central server that receives model weight updates from participating clients and aggregates them to create a global model, potentially enhancing performance.

To fully leverage the attractive benefits of FL in many application domains, it is essential to meet a set of requirements that we highlighted below so that the data holder will be encouraged to join this collaborative paradigm:
\begin{itemize}
    \item \textit{Available high-quality data at each client:}
     Data is the lifeblood of machine learning. Although each participating entity may independently decide how it collects, extracts, and organizes its data according to its environment and preferences, it is imperative to gather high-quality data and prepare them adequately to flourish the FL training procedure.

     \item \textit{Computation capacity at each client:}
     Data alone is insufficient to fuel the FL process; power must accompany knowledge. To uncover the FL advantages, each client should participate in multiple communication rounds before achieving the desirable performance, whether it is a mobile device or a specialized company, and as such, a formidable computation capacity is mandatory, enabling each client to remain active in the long run. 

     \item \textit{Reliable communication between actors:}
     The third vital necessity that underpins FL's success is the establishment of reliable communication between the clients and the aggregator node. The exchange of local model updates and the global model weights, whether orchestrated through a central server or in a decentralized fashion, must occur with the utmost safety, security, and efficiency. Without such dependable communication channels, malicious actors could tamper with client models in transit or steal sensitive dataset information.  

     \item \textit{Reliable aggregation method:}
      Careful contemplating of the algorithm that combines the knowledge harvested from individual clients during each communication round is indispensable to propel FL adoption worldwide. An aggregation method must not only be fair and reliable but also robust and capable of weaving together diverse insights despite the variations among clients.
\end{itemize}

\begin{figure}[htbp]
    \centering
    \includegraphics[scale=0.4]{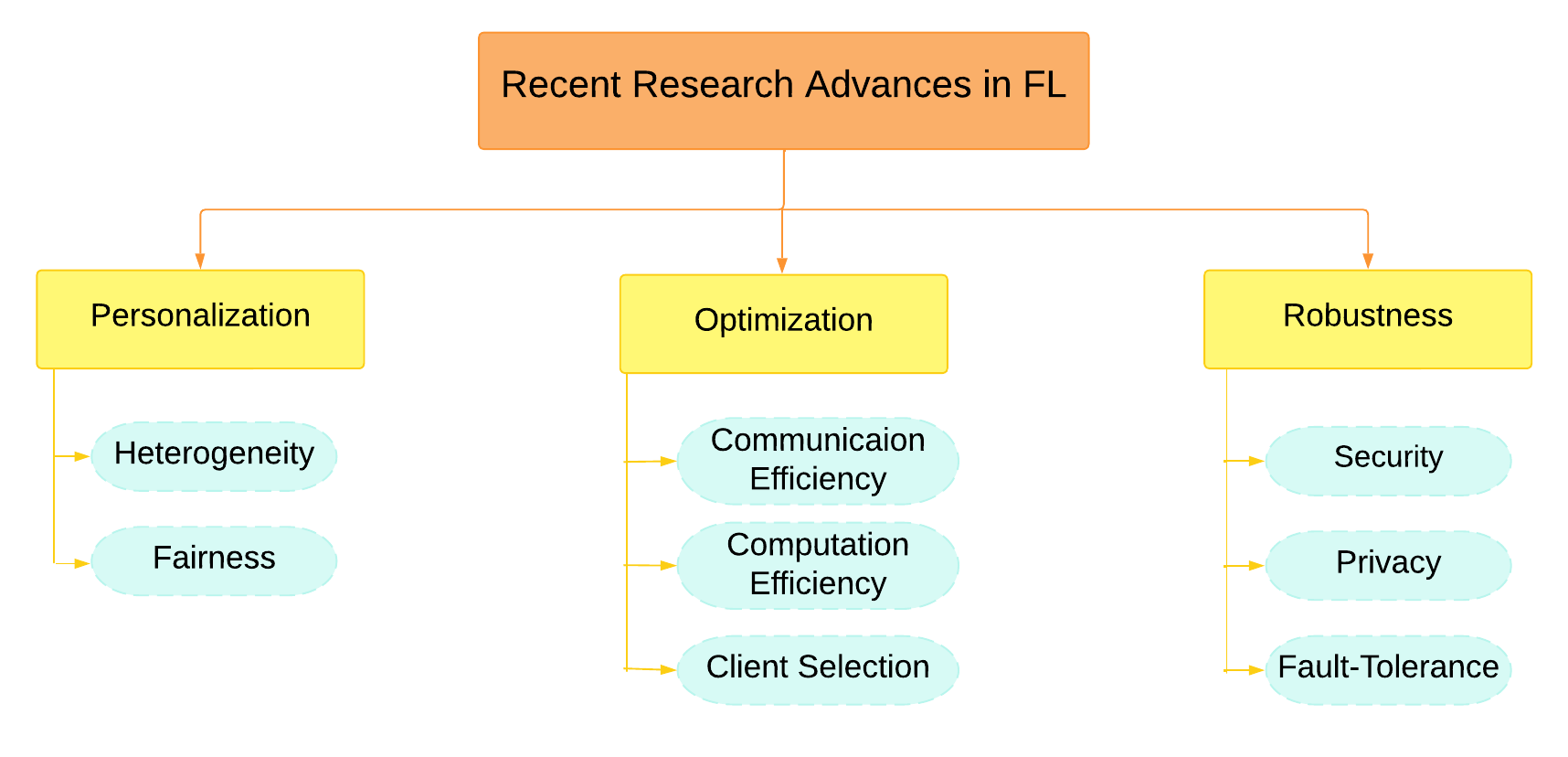}
    \caption{High-Level Classification of Recent Research Advances in Federated Learning.}
    \label{FL-recent-advances}
\end{figure}

As the FL landscape has witnessed a rapid evolution recently, active FL researchers are in constant pursuit of building the most efficient approaches to meet contemporary system requirements. By carefully assessing a range of evaluation metrics, they aim to ascertain the correctness and originality of their work. Usually, the new FL proposals effectively enhance one or more of the assessed metrics but lead to trade-offs that negatively impact the remaining ones. In our study, we intend to aid researchers in identifying the hot research topic in FL and ensure they are informed and up-to-date on the latest methods. For this aim, we present a high-level classification, as illustrated in Fig. \ref{FL-recent-advances}, that clusters these evaluation metrics into three research areas: \textit{optimization, personalization,} and \textit{robustness}. Additionally, our investigation will primarily focus on aggregation-based solutions (please refer to Section \ref{fl_recent_techniques_and_aggregation}) to elucidate recent advancements within each cluster. Simultaneously, we will explore relevant complementary techniques that may enhance the aggregation process.
The potential of aggregation as a solution area for addressing inherent challenges in Federated Learning is clear. Regardless, it's crucial to note that only a minority of survey articles have given significant attention to tackling challenges and showcasing contributions from the federated learning aggregation vision. 

\begin{itemize}

\item \textbf{Personalization.}

This category concerns the ability of the FL model to accommodate the varied characteristics and requirements of individual clients, regardless of their heterogeneity. The heterogeneity in the FL ecosystem may manifest in manifold ways. For instance, we briefly distinguish two forms of heterogeneity: \textit{statistical} and \textit{system} heterogeneity. The former deals with the different data quality, quantity, and distribution, while the latter is associated with the diversified hardware capacities, operating systems, and available resources across all clients. On the other hand, the FL framework must also consider a fair aggregation that does not discriminate based on the user's localization, sex, or other properties. Generally speaking, the ultimate goal in this context is to handle a good trade-off between personalization, bias mitigation, and privacy preservation. Thus, implementing a personalized federated learning solution can be an arduous endeavor. The motivated researchers towards tackling this class of challenges are required to successfully evaluate the confronted heterogeneity types and design an appropriate approach that accounts for these factors without compromising the private-preserving aspect of the FL mechanism. 
Section \ref{heteroheneity_in_FL} pinpoints the origins of heterogeneity. While in Section \ref{FL_agg_solution_hete}, we offer systematic categorization for the respective solutions in this context.

\item \textbf{Optimization.}

This category pertains to the various parameters to devise a practical FL model that converges rapidly. In other words, scientists interested in this class are willing to attain a gain in convergence rate. Accordingly, it is necessary to embrace the implementation of meticulous client selection, efficient communication, and an optimized scheme for resource allocation. However, several constraints may inhibit the achievement of this research target. For example, the FL systems encounter unfavorable environmental factors, including network latency, noisy communication channels, client mobility, and diverse types of heterogeneity among the participating entities. To expedite the convergence rate and adopt a rigorous approach that falls into this category, researchers need to ensure that the FL model is highly efficient, valid, and profitable under real-life conditions.
In section \ref{communication_efficiency_in_FL}, we will exhaustively elaborate on the communication constraints and client selection challenges, followed by their related solutions as found in the literature, in Section \ref{FL_agg_solution_communication}.

\item \textbf{Robustness.}

Since FL was initially introduced to ensure data privacy and facilitate more secure protocols, it becomes imperative to alleviate and remediate the security and privacy concerns that impede the realization of a resilient FL solution. The obstacles may either be inherent in the traditional ML model or have arisen due to the distributed nature of FL. The proposals belonging to this category aim to guarantee robustness against numerous threats. While developing their contributions, researchers must incorporate security measures to counter potential attacks, privacy-preserving mechanisms to prevent information disclosure, and fault-tolerance techniques to handle network failures and malicious actors. Introducing innovative solutions that pertain to this category has gained growing interest, especially in sectors handling sensitive data, such as healthcare and insurance.
we conduct in Section \ref{security_privacy_in_FL} an in-depth analysis of the security and privacy issues to foster a better understanding of the corresponding security breaches and privacy violations. Separately, the contemporary defense strategies proposed in the arts are extensively discussed in Section \ref{FL_agg_solution_sec_priv}.
 
\end{itemize}
\setlength\rotFPtop{0pt plus 1fil} 

\begin{sidewaystable}[htbp]
\tiny
\caption{Recent FL State-Of-The-Art Survey}
\label{tab_related_work_publishing_details}
\begin{tabularx}{\linewidth}{| >{\hsize=.05\hsize\linewidth=\hsize}X|  >{\hsize=.025\hsize\linewidth=\hsize}X| >{\hsize=.1\hsize\linewidth=\hsize}X| >{\hsize=.825\hsize\linewidth=\hsize}X |}

\hline
\textbf{Related survey} & \textbf{Year} & \textbf{Main Subject} & \textbf{Limitations}  \\
\hline

\cite{aledhari2020federated}
& 2020
& General FL Techniques
& - The list of the discussed challenges is limited to general concerns, such as \textit{Statistical Heterogeneity, System Heterogeneity, Network Connectivity, and Security}, with a very brief examination. 

- No classification is provided for the techniques found in the literature to address the discussed challenges. For example, in Section '\textit{Optimization Techniques To Federated Learning Models}', the authors linearly list a set of FL algorithms that are applied to enhance FL optimization without explicitly naming the solution classes.

- No clear separation is made in the content structure to distinguish between the discussed challenges and the techniques used for tackling these concerns.\\ \hline

\cite{abdulrahman2020survey}
& 2020
& Detailed FL Analysis 
& - This work classifies FL topics and research areas into four main categories, without providing clear criteria for their taxonomy, which may be confusing for readers. For example, the \textit{communication cost} and \textit{client selection} are mentioned in \textit{System Models and Design} class, while \textit{resource management} is a distinct class in their classification.

- The concept of aggregation algorithms is shortly presented under the title '\textit{Optimization and Aggregation Algorithms}', but in fact, the aggregation methods serve many goals other than FL system optimization, such as personalization, robustness, bias mitigation, fairness, etc.,\\ \hline
 
\cite{wahab2021federated}
& 2021
& Comprehensive FL Tutorial
& - This work provides a multi-level classification for FL challenges and associates them with state-of-the-art solutions, however, the list that encompasses  FL concerns and techniques is still limited and not updated compared to ours.

- No detailed discussion about FL aggregation or the proposal's evaluation.

- No experiments are performed to provide insightful performance comparisons between FL proposals.
\\ \hline

\cite{pfitzner2021federated} 
& 2021
& FL in Medical Context
& - This review is specifically dedicated to the healthcare informatics community, as they provide examples and contextualized discussion only in the medical context.

- The classification resulting from their literature analysis includes six classes: \textit{Characteristics of Federated Datasets, Learning Algorithms,  Communication Efficiency, Attacks, Defences, and Health}, which list the FL problems, solutions, system characteristics, and the application domain at one level. This simplistic presentation makes it hard for new researchers to comprehend the related topics to FL.  
\\ \hline
\cite{imteaj2021survey}
& 2021
& FL for Resource-Constrained IoT 
& - The study discussed FL from a restricted standpoint of resource-constrained IoT environments. 

- The taxonomy provided for the FL systems embraces categories related to the system design (e.g., \textit{Partitioning Samples, Federation Scale}) with other categories associated with FL techniques (\textit{Privacy Mechanisms, Encouragement Towards FL}) while ignoring key FL facets, such as FL aggregation and security concerns.

- No clear organization structure or comparison analysis is provided for the mentioned strategies.
\\ \hline

\cite{nguyen2021federated}
& 2021
& Comprehensive FL Survey
& - The proposed classification of related FL topics is mainly dedicated to the combination of FL and IoT (IoT services, IoT applications, and IoT challenges).

- No experiments are performed to provide insightful performance comparisons between FL proposals.

- No special emphasis is placed to discuss FL aggregation.
\\ \hline
\cite{nguyen2021federatedblockchain}
& 2021
& FL with Blockchain 
&- The mentioned challenges are briefly discussed, while the associated solutions are not classified.

- No sub-classification is provided for the studied challenges.

- No comparison analysis between the surveyed papers throughout the study.
\\ \hline

\cite{li2021survey} 
& 2021
& FL Systems
& - The taxonomy provided for the FL systems encompasses classes related to the system design (e.g., \textit{Data Partitioning, Communication Architecture, Scale}) with other classes associated with FL techniques (\textit{Privacy Mechanisms}), which may confuse the readers.

- The taxonomy presented for popular work in FL is not based on clear classification criteria (e.g., \textit{Effectiveness Improvement, Practicality Enhancement}). 
\\ \hline

\cite{mothukuri2021survey}
& 2021
& Security and Privacy in FL-IoHT
& - This survey focuses only on FL security and privacy issues without discussing other key FL aspects such as personalization, heterogeneity, and scalability.

- The proposed taxonomy includes \textit{aggregating and optimizing algorithms} at the same level as other classes of system design (e.g.,  \textit{data partition, data availability, network topology, and frameworks}), restricting the room for a thorough discussion about the FL aggregation and efficiency aspects.
\\ \hline

\cite{tan2022towards}
& 2022
& Personalized FL
& - This survey is devoted to FL personalization techniques without examining other major FL concerns, such as security and privacy.

- No comparison analyses between the surveyed papers throughout the study.

- The paper provides insights into Personalized FL Benchmarks without executing any experiments for practical FL analysis.
\\ \hline

\cite{criado2022non}
& 2022
& Heterogeneous FL 
& - The multi-level classification provided in this work concerns only the possible causes for statistical heterogeneity (non-iid data distribution) and the remarkable strategies to face it, while neglecting other main types of heterogeneity, including system, model, and resource heterogeneity.

- This work does not include other FL areas of investigation, such as scalability, privacy, and communication and computation efficiency.
\\ \hline

\cite{nguyen2022federatedhealth}
& 2022
& FL in Smart Healthcare
& - No hierarchical classification was provided to organize the paper's content, which focuses mainly on the healthcare application domain.

- No separation is made for distinguishing the FL encountered challenges and the devoted techniques in the literature.

- The different causes for FL issues (problems of privacy, security, efficiency, heterogeneity) are briefly discussed.
\\ \hline

\cite{lyu2022privacy}
& 2022
& Privacy and Robustness in FL 
& - The classification provided in this work concerns only the possible privacy attacks and the emerging technologies to fight against them while not discussing other related topics to FL, such as personalization, efficiency, and optimization.

- No experiments are performed to provide insightful performance comparisons between FL proposals.
\\ \hline

\cite{rodriguez2023survey}
& 2023
& Attacks and Defenses in FL
& - The multi-level classification presented in this work concerns only the security and privacy attacks and the related defense solutions, without discussing other key factors in the FL landscape. 

- No comparison analysis between the surveyed papers throughout the study.

- The experiments performed in this work are dedicated to only studying the effects of certain attacks, ignoring other influential elements, such as data heterogeneity, system heterogeneity, and client dropout. \\

\hline

\cite{almanifi2023communication}
& 2023
& Efficiency in FL 
& - The multi-level classification provided in this work concerns only the communication and computational efficiency, without examining other FL topics, such as personalization, robustness, and heterogeneity.

- No comparison analyses between the surveyed papers throughout the study.
\\ \hline

\cite{yin2024device}
& 2024
& On-Device Recommendation Systems
& - This study presents a comprehensive survey of the On-Device Recommendation Systems. Federated learning is examined as one of the possible approaches for training and securing such a system. Thereby, it does not consist of the main subject of investigation.\\ 
\hline

Our Survey
& 2023
& Comprehensive FL Tutorial 
&  \\ \hline

\end{tabularx}  
\end{sidewaystable}

\subsection{Related Work}    

Since its inception in 2016, Federated Learning has garnered significant attention, resulting in a notable surge in research publications over the past few years. Several detailed surveys have emerged to explore the FL area, each with a different focus. To categorize the evolving perspectives observed in FL surveys based on our in-depth lectures, we have discerned the primary optics characterizing the trajectory of FL survey papers over time as follows:

\begin{itemize}

    \item \textbf{Broad Description and General Concepts Examination [2016-2020]:}
    In the earlier years, up until 2020, researchers from diverse backgrounds, including distributed ML, databases, and edge intelligence, conducted fundamental efforts to provide a descriptive and broad overview of general concepts surrounding FL during its nascent stages, such as enabling technologies, protocols, architectures, frameworks, and application domains \cite{aledhari2020federated, abdulrahman2020survey}. However, the research contributions landscape has gradually shifted from general perspectives toward a more specialized analysis of FL. 
    
    \item \textbf{Specialized FL Examination: Projecting FL onto a Specific Domain Application [2020-2024]:}
    Reflecting the maturation of FL research, this line focuses on projecting FL onto a specific domain application, such as healthcare \cite{pfitzner2021federated, nguyen2022federatedhealth}, resource-constrained environment \cite{imteaj2021survey}, and recommendation systems \cite{yin2024device}, delivering specialized examinations of FL within distinct domain applications.

    \item \textbf{Specialized FL Examination: In-depth Exploration over Limited Aspects [2020-2024]:}
    The second direction in specialized FL surveys involves in-depth explorations of specific aspects (one or two) of the FL ecosystem, such as personalization \cite{tan2022towards} heterogeneity \cite{criado2022non}, privacy \cite{lyu2022privacy, rodriguez2023survey}, or resource efficiency \cite{almanifi2023communication}. Remarkably, some works amalgamate these two approaches, as exemplified in \cite{mothukuri2021survey}, wherein the authors conduct an intriguing FL study in the healthcare domain, delving into privacy and robustness.

    \item \textbf{Integration of Federated Learning with Emerging Technologies [2020-2024]:}
    An alternative and contemporary trend breaks down the integration of FL with emerging technologies, such as IoT \cite{nguyen2021federated}, IoMT, and blockchain \cite{nguyen2021federatedblockchain}, seeking to harness the synergistic potential of their fusion.
    
\end{itemize}

In order to position our work in the existing literature, we have carefully selected relevant surveys published in [2020-2023], which are encapsulated in Table \ref{tab_related_work_publishing_details}, to accentuate the primary topics and the main limitations in their work.

\subsection{Motivations and Contributions}

Notwithstanding the diversity of existing work on FL, the current literature lacks an all-encompassing survey paper that considers numerous perspectives we have observed during our state-of-the-art analysis. The motivation for this survey stems from the following observations:

The first observation is that \textit{recent efforts have primarily focused on fundamental knowledge and well-known challenges, such as statistical heterogeneity, security attacks, and energy efficiency}.  Nonetheless, there is a pressing need for up-to-date and more comprehensive research that delves deeper beyond these common aspects, examining the less-discussed FL considerations, such as client selection, model architecture, knowledge distillation, bias mitigation, and fairness.

The second finding is that \textit {existing surveys have not adequately met the need for a finely structured and multilevel classification scheme that effectively organizes work contributions and showcases recent advances in the field}. Instead, their classification schemes typically rely on a narrow perspective, focusing solely on FL challenges, architectures, or scales, often without providing clear and logical criteria for defining encapsulated categories. This lack of clarity and hierarchy results in an ambiguous content structure that makes it difficult for researchers to rapidly extract and comprehend relevant information aligned with their specific subject of interest. 

The third insight is that \textit{previous studies have largely neglected the critical aspect of FL aggregation. To the best of our knowledge, none of the prior studies have placed an exceptional emphasis on FL aggregation}. Although, the choice of aggregation algorithm and pipeline are pivotal features that affect the overall FL system performance significantly, especially in centralized FL. In this setup, a single server controls all the orchestration tasks. Therefore, any failure or breach of information on the server side can lead to flawed models and suboptimal outcomes.

The last notable point pertains to \textit{the lack of complete guidelines that outline systematic methodologies for conducting realistic experiments to quantify the contribution of novel FL proposals}. Researchers often spend significant time and effort identifying relevant parameters for their evaluation testbeds. By carefully selecting a testbed, they can effectively demonstrate the efficiency of their FL solutions. However, the absence of a notable reference source that provides substantial insights in this context presents a challenge. The evaluation configuration encompasses numerous FL components, including realistic or benchmark datasets, diverse types of heterogeneous data distribution at various levels, performance metrics tailored to different FL scenarios, DL model architecture and hyperparameters, and the number of participating clients.

To fill these gaps in the literature, we have been motivated to present an exhaustive FL survey that accounts for the following contributions:

\begin{itemize}
    \item We present a state-of-the-art survey that delves into the latest advances in federated learning. Our hybrid methodology for paper selection combines a bibliometric analysis with a systematic approach that offers a more vast and in-depth view of the FL paradigm.

    \item We investigated the holistic and contemporary techniques found in the current literature to address the inherent challenges of federated learning. In order to improve the organization of our paper, we first identified three prominent clusters of research contributions as the top-level view of our advanced FL taxonomy: \textit{personalization, optimization, and robustness}.

   \item To promote a deeper understanding, we go beyond the traditional path and organize the reviewed works by their respective cluster of FL advances. We introduce then a well-structured and multilevel classification scheme for each encountered challenge and its corresponding solutions, separately, resulting in six distinct schemes. 
   
   \item Moreover, the classification criteria of our FL taxonomy, on which we will elaborate later in the paper, are carefully defined to ensure clarity, hierarchical presentation, and comprehensiveness. This systematic FL map facilitates a straightforward analysis of the various facets of the FL domain, assisting researchers in effectively navigating the complexities and identifying emerging trends in the field.

  \item To our knowledge, this is the first study comprehensively examining FL aggregation. Our focus spans from fundamental considerations, such as the aggregation architectures and scales, to more sophisticated aspects, such as the underlying motivations and synchronization modes, resulting in a complete FL aggregation classification. 
  
  \item Based on this aggregation lens, we consistently explored the environment and goals of most surveyed papers throughout this study. This strategy allowed us to shed light on the context and the achieved purposes of researchers' efforts across broader lines of investigation.

 \item We conducted a series of experiments to guide researchers and provide practical insights into the process of evaluating FL proposals by simulating real-world settings. Specifically, we selected four algorithms from different classes of solutions and compared their performance in response to various and pertinent parameters. Through this experimental comparison, we discuss behavioral trends of algorithms incorporating various mechanisms to tackle FL challenges. 

 \item To aid fellow researchers in identifying future trends, we offer a captivating array of research paths that beckon further investigation, unlocking a wealth of opportunities for advancement. 
\end{itemize}

Table \ref{tab_related_work} presents the conducted analysis comparing our contributions with those of other survey papers, illuminating the distinctive value we bring to the research community. To facilitate the comparison, we have established a multi-level, multi-criteria framework to evaluate the similarities and differences between the existing literature and our work.

Firstly, we consider the main covered facets of federated learning, encompassing:
\begin{itemize}
    \item \textbf{FL Basics:} This criterion assesses whether the paper provides an overview of an FL system.

    \item \textbf{Heterogeneity:} Here, we evaluate if the paper covers common types of heterogeneity in FL systems.

   \item \textbf{Efficiency:} This criterion examines the extent to which the paper addresses efficiency within the FL domain.

   \item \textbf{Security and Privacy:} These criteria focus on the paper's treatment of security breaches and privacy violations observed within the context of FL.
\end{itemize}

Secondly, we delve into the organizational structure of the paper, prioritizing readability and clarity. To assess whether the survey paper is easily navigable and reader-friendly comprehension, we have defined the following metrics:

\begin{itemize}
    \item \textbf{Separation between System and Challenges:} An effectively organized survey would feature distinct sections dedicated to the operational aspects of the FL system and its inherent challenges. However, overlooking this detail by presenting them jointly in a monolith format impedes audience comprehension.

    \item \textbf{Separation between Challenges and Solutions:} Similarly, we advocate for separate sections discussing FL challenges and the contemporary techniques employed to address them, facilitating a clear understanding.

    \item \textbf{Multi-level Classification:} We scrutinized whether the survey paper employs a hierarchical framework for structuring FL aspects, challenges, and techniques, enabling readers to effectively identify relevant information.

    \item \textbf{Clear Criteria for Classification:} While some FL surveys provide a taxonomy for presenting their content, a luck of clarity regarding the criteria guiding their classification is prevalent.
\end{itemize}

Thirdly, we identify the distinctive research features that significantly enhance the survey's quality and set it apart from other similar works.

\begin{itemize}
    \item \textbf{Focus on Aggregation:} As elaborated earlier in the paper, aggregation profoundly impacts the FL system's performance. Nonetheless, existing literature often lacks thorough discussions on this aspect. Tackling this research gap is a primary motivation and distinguishes our survey.

    \item \textbf{Detailed Methodology:} This pertains to the methodology used by authors for selecting relevant references. A comprehensive survey carefully attends to selection strategies, typically employing standard and robust tools to filter vast research databases.

    \item \textbf{Evaluation of FL Proposals:} Survey papers outline accomplishments in distinct domains, but evaluating surveyed papers against pertinent and diverse criteria is crucial. The tabulated comparison presents a concise summary and highlights notable strengths and weaknesses of each proposal.

    \item \textbf{Experimentation:} Generally, simulations validate proposed algorithms and reveal their behavioral trends. However, a notable minority of FL survey papers integrate experimental aspects with theoretical examinations, leading to an incomplete analytical framework.
\end{itemize}

\addtocounter{table}{-1}

\setlength\rotFPtop{0pt plus 1fil} 

\begin{table}[htbp]

\tiny
\caption{Comparison of Our Work to Related Recent Surveys}
\label{tab_related_work}
\begin{tabular}{|m{0.7cm}|m{0.4cm}|m{0.5cm}|m{0.3cm}|m{0.3cm}|m{0.3cm}|m{0.95cm}|m{1.05cm}|m{0.6cm}|m{0.8cm}|m{0.75cm}|m{0.7cm}|m{0.9cm}|m{0.55cm}|}

\hline 
   \multirow{2}{*}{} & \multicolumn{5}{c|}{\textbf{Covered Aspect of FL}} &  \multicolumn{4}{c|}{\textbf{Content Structure}} & \multicolumn{4}{c|}{\textbf{Research Features}}  \\
\hline
\tiny {Related Surveys} & \tiny {FL Basics} & \tiny {Hetero-geneity} & \tiny {Effic-iency} & \tiny {Secu-rity} & \tiny {Priv-acy} & \tiny {Separation System Vs Challenges} & \tiny {Separation  Challenges Vs Solutions} & \tiny {Multi-level Classif} & \tiny {Clear Criteria for Classif} & \tiny {Focus on Aggregation} & \tiny {Detailed Methodology} & \tiny {Evaluation of FL proposals} & \tiny {Exper-iments} \\
\hline 
\cite{aledhari2020federated}

& \text{--}   &
 \checkmark  &
  \text{--}   &
  \checkmark &
 \text{--}   &
  \checkmark &
  \text{--}   &
  \text{--}   &
  \text{--}   &
 \text{--}   &
  \text{--}   &
  \text{--}   & 
  \text{--}  \\
  \hline

\cite{abdulrahman2020survey}
   & \checkmark
   & \text{--}  
   & \checkmark
   & \checkmark
   & \checkmark
   & \checkmark
   & \text{--}  
   & \text{--}  
   & \text{--}  
   & \text{--}  
   & \text{--}  
   & \text{--}   
   & \text{--}  \\
   \hline
  
  \cite{wahab2021federated}
   & \checkmark
   & \checkmark
   & \checkmark
   & \checkmark
   & \checkmark
   & \checkmark
   & \checkmark
   & \checkmark
   & \text{--}  
   & \text{--}  
   & \checkmark  
   & \text{--}  
   & \text{--}  
   \\
   \hline
  \cite{pfitzner2021federated} 
   & \checkmark
   & \checkmark
   & \checkmark
   & \text{--}   
   & \checkmark
   & \text{--}  
   & \text{--}  
   & \text{--}  
   & \text{--}  
   & \text{--}  
   & \checkmark
   & \text{--}  
   & \text{--}  
   \\
   \hline
   \cite{imteaj2021survey}
   & \checkmark
   & \checkmark
   & \checkmark
   & \text{--}  
   & \checkmark
   & \text{--}   
   & \checkmark
   & \text{--}  
   & \text{--}  
   & \text{--}  
   & \text{--}  
   & \text{--}  
   & \text{--}  
   \\
     \hline
 \cite{nguyen2021federated} 
   & \checkmark
   & \checkmark
   &  \checkmark
   &  \checkmark
   &  \checkmark
   &  \checkmark
   &   \text{--}  
   &   \checkmark
   &   \checkmark
   &   \text{--}  
   &   \text{--}  
   & \checkmark
   & \text{--}  
   \\
      \hline
 
\cite{nguyen2021federatedblockchain}
   & \checkmark
   & \text{--}  
   & \checkmark
   & \checkmark
   & \checkmark
   & \checkmark
   & \checkmark
   & \text{--}  
   & \text{--}  
   & \text{--}  
   & \text{--}  
   & \checkmark
   & \text{--}  
   \\

   \hline
 \cite{li2021survey}
   & \checkmark
   & \text{--}  
   & \checkmark 
   & \text{--}  
   & \checkmark
   & \text{--}  
   & \text{--}  
   & \checkmark
   & \text{--}  
   & \text{--}  
   & \text{--}  
   & \checkmark
   & \text{--}  \\ 
   \hline
  \cite{mothukuri2021survey}
   & \checkmark
   & \text{--}  
   & \text{--}  
   & \checkmark 
   & \checkmark
   & \checkmark
   & \text{--}  
   & \text{--}  
   & \text{--}  
   & \text{--}  
   & \checkmark
   & \checkmark
   & \text{--}  
   \\
   \hline
   \cite{tan2022towards}
   & \text{--}  
   & \checkmark
   & \checkmark
   & \text{--}  
   & \text{--}   
   & \checkmark
   & \text{--}  
   & \checkmark
   & \checkmark
   & \checkmark
   & \text{--}  
   & \text{--}  
   & \text{--}  
   \\
   \hline
 \cite{criado2022non}
   & \checkmark
   & \checkmark
   & \text{--}  
   & \text{--}  
   & \text{--}  
   & \text{--}  
   & \checkmark
   & \checkmark
   & \checkmark
   & \text{--}  
   & \text{--}  
   & \text{--}  
   & \text{--}  
   \\
   \hline

  \cite{nguyen2022federatedhealth}
   & \checkmark
   & \checkmark
   & \checkmark
   & \checkmark
   & \checkmark
   & \checkmark
   & \text{--}  
   & \text{--}  
   & \text{--}  
   & \text{--}  
   & \text{--}  
   & \checkmark
   & \text{--}  
   \\
   \hline
   \cite{lyu2022privacy}
   & \text{--}  
   & \text{--}  
   & \text{--}  
   & \checkmark
   & \checkmark
   & \text{--}  
   & \checkmark
   & \text{--}  
   & \text{--}  
   & \text{--}  
   & \text{--}  
   & \text{--}  
   & \text{--}  
   \\
   \hline
  \cite{rodriguez2023survey}
   & \checkmark
   & \text{--}  
   & \text{--}  
   & \checkmark
   & \checkmark
   & \checkmark
   & \checkmark
   & \checkmark
   & \checkmark
   & \text{--}  
   & \text{--}     
   & \text{--}  
   &\checkmark
   \\
   \hline
 \cite{almanifi2023communication}
   & \checkmark
   & \text{--}  
   & \checkmark
   & \text{--}  
   & \text{--}  
   & \checkmark
   & \checkmark
   & \checkmark
   & \text{--}  
   & \text{--}  
   & \text{--}  
   & \text{--}  
   & \text{--}  
   \\
   \hline

Our Survey 
   & \checkmark
   & \checkmark 
   & \checkmark
   & \checkmark
   & \checkmark
   & \checkmark
   & \checkmark
   & \checkmark
   & \checkmark
   & \checkmark 
   & \checkmark
   & \checkmark
   & \checkmark
   \\
\hline   
\end{tabular}
\end{table}

\subsection{Survey Methodology}
In order to deliver reliable and engaging state-of-the-art analysis, we have combined two robust methodologies, leveraging their notable strengths. We have started with an initial \textit{bibliometric study}, utilizing a software-assisted mechanism to explore vast scholarly databases and discern pivotal FL specifications. Building upon this foundation, we undertake a \textit{systematic literature review}, probing deeper into the influential works and emerging techniques. This hybrid strategy, rarely observed in previous FL research, ensures transparency, objectivity, and ample coverage of the FL realm. Hereafter, we will explain in detail each of these steps.  

\subsubsection{\textbf{Bibliometric Analysis with CiteSapce}}
Citespace is a Java application designed for analyzing and visualizing scholarly literature \cite{chen2006citespace}. It facilitates intricate patterns, trends, and connections uncovering within large-scale academic databases. Specifically, CiteSpace offers many research features such as co-citation analysis, co-keyword analysis, network visualization, and knowledge mapping. Additionally, it allows various filtering options, including time slicing, burst detection, and cluster identification, which empower scholars to track the essence of a research front and understand its dynamics as it evolves constantly. Thereby, scientists and analysts will be able to keep up with the rapid advances and remain informed about the latest trends in the body of the domain of interest.

\paragraph{Data Extraction}

We utilized Dimensions \cite{dimensions} as our primary data source, renowned for its daily updates with fresh articles from over 130 publishers, accessing an extensive collection of scholarly publications. Our research in the initial phase, conducted throughout the first half of 2023, involved a detailed analysis of titles and abstracts of research publications. To achieve this, we focused on articles, conference proceedings, and preprints published between 2020 and 2023, aiming to capture the latest advances in FL. As our study advanced, we broadened our data collection to include influential works from the latter half of 2023, ensuring a more comprehensive coverage. Moreover, It is important to note that in the subsequent phase of our methodology, we conducted a thorough traditional review, incorporating numerous papers predating 2020, acknowledging their foundational contributions to the field of FL.

To ensure a comprehensive investigation through our research queries, we compiled a set of keywords commonly associated with federated learning, including : \textit{("federated learning" OR "federated training" OR "federated machine learning" ... OR "federated implementation")}
By combining these keywords with various FL-specific-area terms, we sought to capture a clear view of the subject. For instance, to gain a high-level standpoint, we have combined the abovementioned words set with other survey-related keywords to form the following query: 

\textit{[federated-learning-keywords] AND ("survey" OR "overview" OR "review" OR "trends" OR "challenges")}

This query yielded 4040 relevant documents, which we used as our input for the bibliometric mapping by CiteSpace.

\paragraph{Network Mapping and Clustering}

In this section, we explore the network visualizations provided by CiteSpace tools. While we utilized diverse project configurations and application tools, we focus here on presenting an overview of the clustering results as one of the most powerful features. The clustering function organizes the nodes into clusters, each depicted by a unique color and title. The node corresponds to a specific type of data in each project case.

\begin{figure*}[htbp]
    \includegraphics[width=\textwidth]{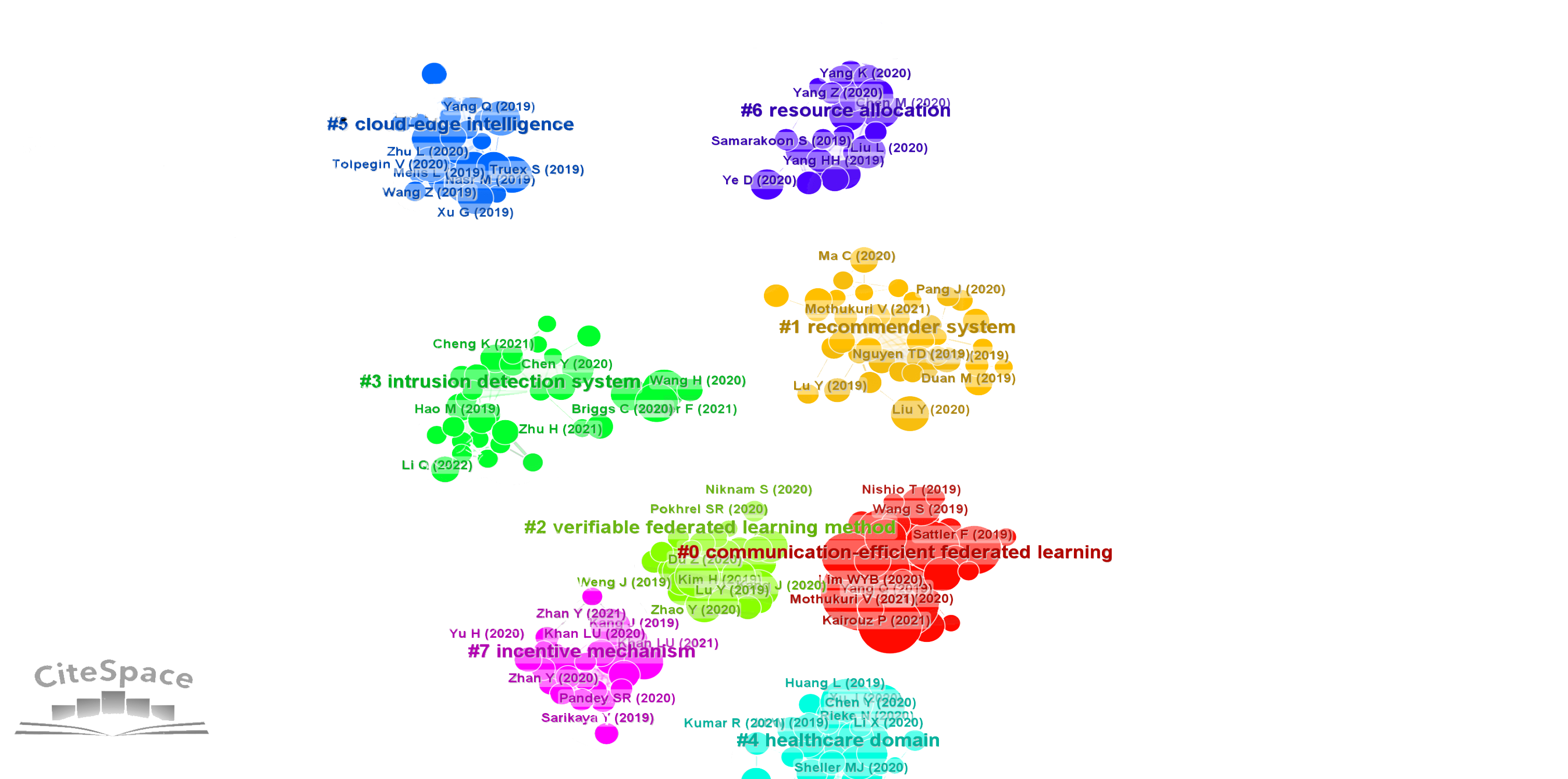}
    \caption{FL Research Area Clustering}
    \label{clusters-FL-general-co-reference}
\end{figure*}

One of the scenarios explored is co-reference-based clustering. Here, each node within a cluster represents the papers from our input dataset along with their cited references. As a result, a cluster will group together papers that examine similar or closely related topics. However, it is important to note that the automatically assigned titles are solely based on the paper titles and might not fully articulate the subject of interest within each class. Aiming for better insights, we utilize the cluster explorer tool, which provides more comprehensive details and statistics. Fig. \ref{clusters-FL-general-co-reference} displays the top height (8) ranked clusters.

\subsubsection{\textbf{Systematic Literature Review}}

We used the knowledge extracted from CiteSpace visualization to inform the second step of our survey methodology, which we started by asking the question:

\begin{itemize}
    \item \textit{ \textbf{Q1}: What are the prominent clusters of contributions under which fall most of the work in FL?}
\end{itemize}

Answering this question allowed us to provide a high-level view of the FL landscape by dividing it into three clusters: \textit{Personalization, Optimization, and Robustness}. Then, using these clusters as our compass, we embarked on a more detailed exploration of FL concerns, which guided us to our second question:

\begin{itemize}
    \item \textit{ \textbf{Q2}: What are the recent FL advances within each identified cluster?}
\end{itemize}

Addressing this question, we identified four system constraints: Heterogeneity, Efficiency, Security, and Privacy, which present the second level of our content organization (see Sections \ref{heteroheneity_in_FL}, \ref{communication_efficiency_in_FL}, \ref{security_privacy_in_FL}, respectively). Since we found impressively a significant number of publications that fall under these FL aspects, we conducted rigorous readings of FL research papers, selectively curated based on CiteSpace statistics, to answer the third question: 

\begin{itemize}
    \item \textit{ \textbf{Q3}: What are the corresponding encountered impediments and employed techniques within each FL aspect?}
\end{itemize}

We thoughtfully organized the results of our analysis regarding this question into classes and subclasses to separately highlight the challenges and recent solutions for each aspect.

\begin{figure}[htbp]
    \centering
    \includegraphics[width=0.7\textwidth]{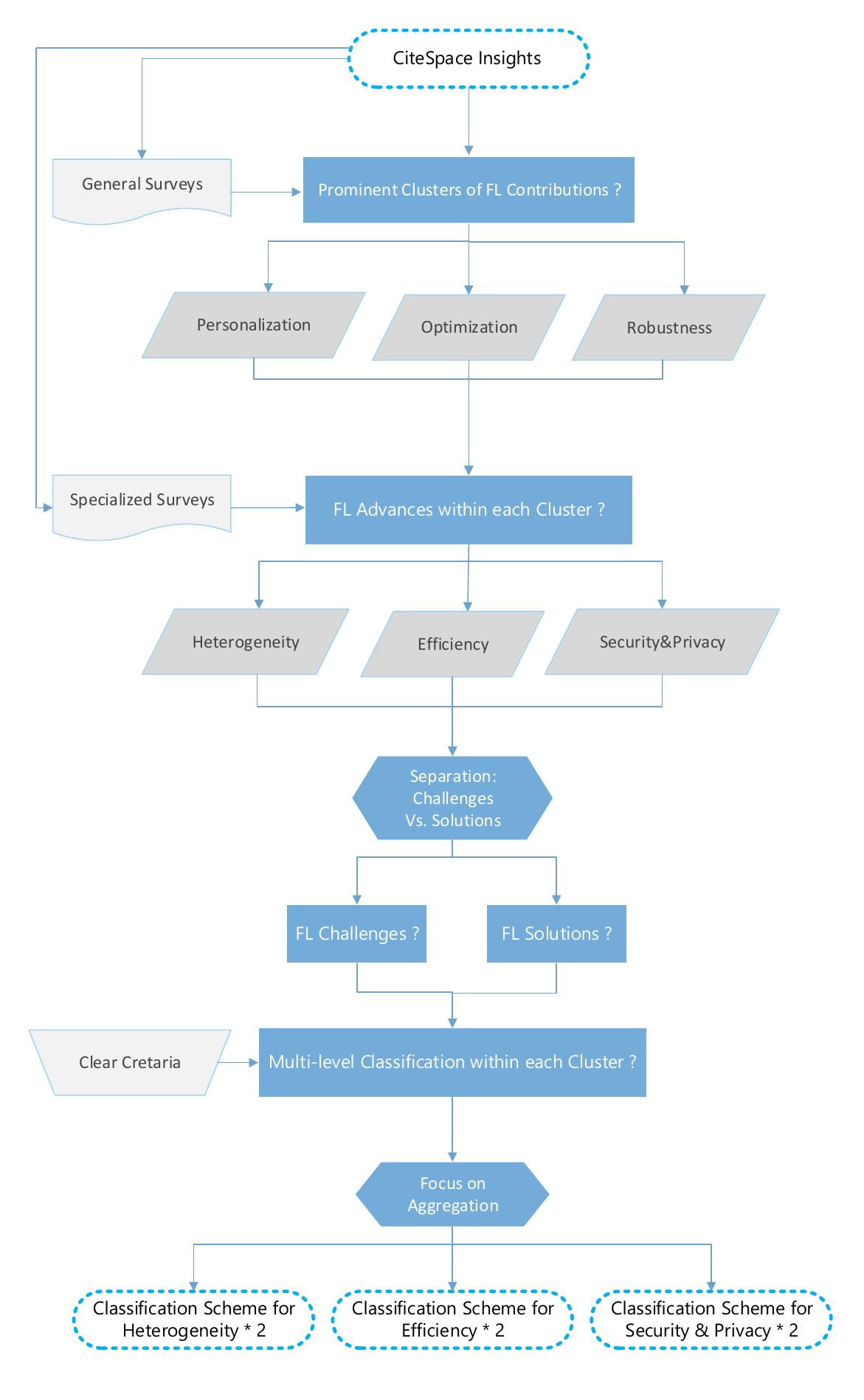}
    \caption{Flowchart For Our Systematic Survey Methodology.}
    \label{flowchart_systematic_method}
\end{figure}

In the final stage, we refined our content from the perspective of FL aggregation. This meticulous process led to the formulation of three unique classification schemes, outlining the impediments associated with efficient FL aggregation (See Section \ref{FL_ecosystem}). Also, it results in three other classification schemes showcasing the most recent and widely adopted techniques for FL advancements (See section \ref{fl_recent_techniques_and_aggregation}). Fig. \ref{flowchart_systematic_method}  presents a visualization of the entire strategy's flowchart diagram.  

\begin{figure*}[htbp]
\centering   

\subfloat[\scriptsize{Distribution of surveyed papers per publisher.}]
    {\includegraphics[scale=0.275]{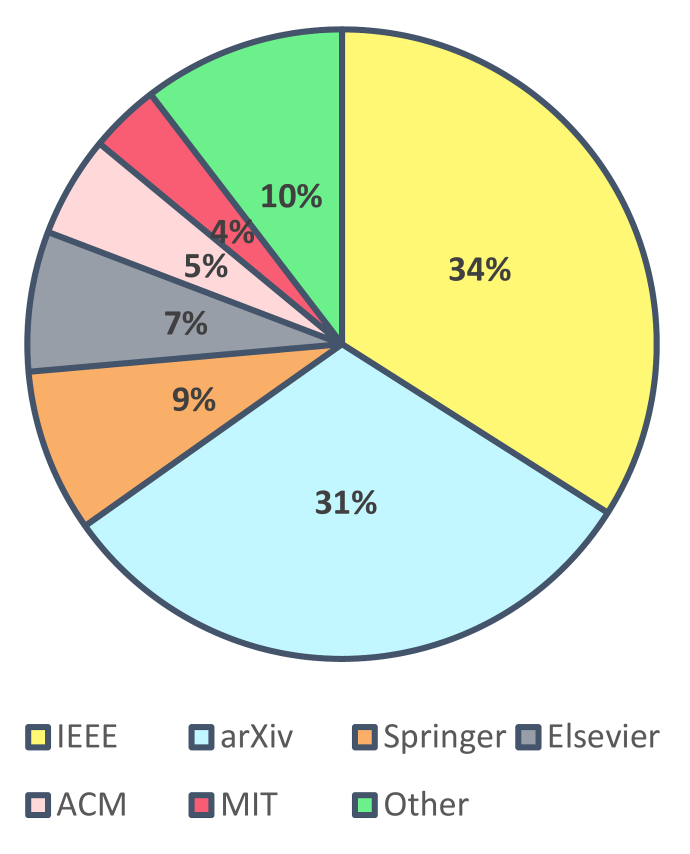}
    \label{survey-publishers-insights}}
    \hfill%
\subfloat[\scriptsize{Distribution of surveyed papers per year of publication.}]
    {\includegraphics[scale=0.3]{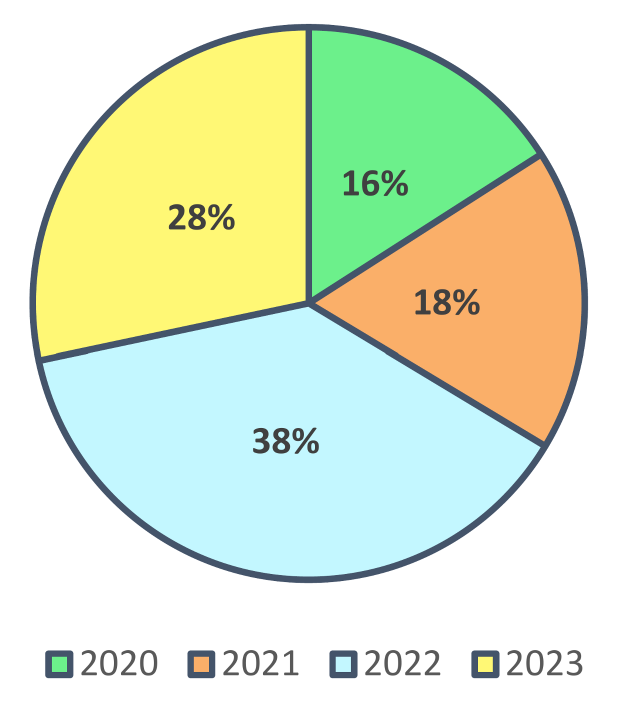}
    \label{survey-years-insights}}
    \hfill%
\subfloat[\scriptsize{Distribution of surveyed papers per covered FL aspects.}]
    {\includegraphics[scale=0.3]{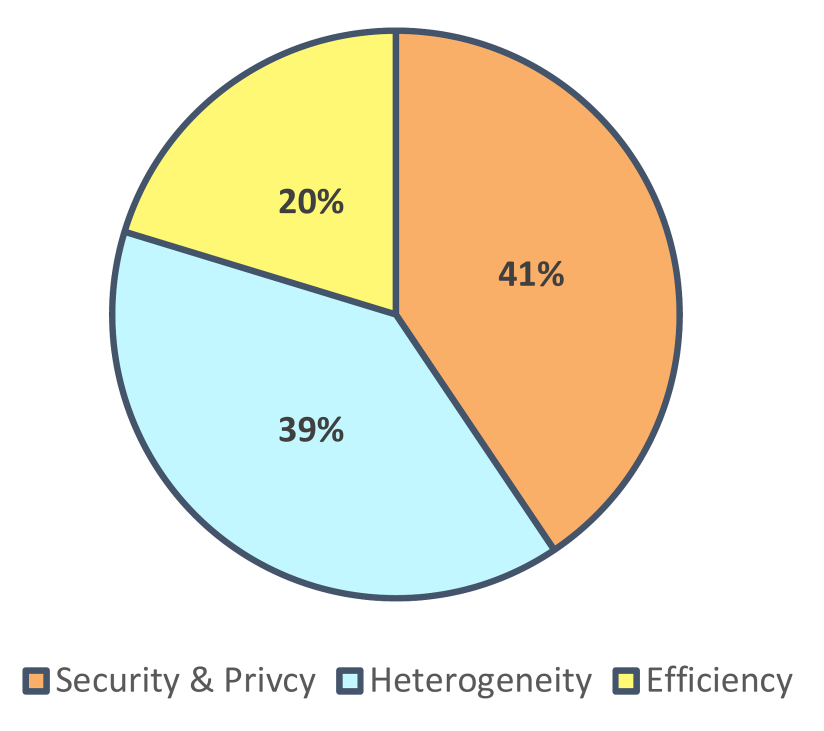}
    \label{survey-aspects-insights}}
    \hfill%
\subfloat[\scriptsize{Distribution of surveyed papers per publication type.}]
    {\includegraphics[scale=0.32]{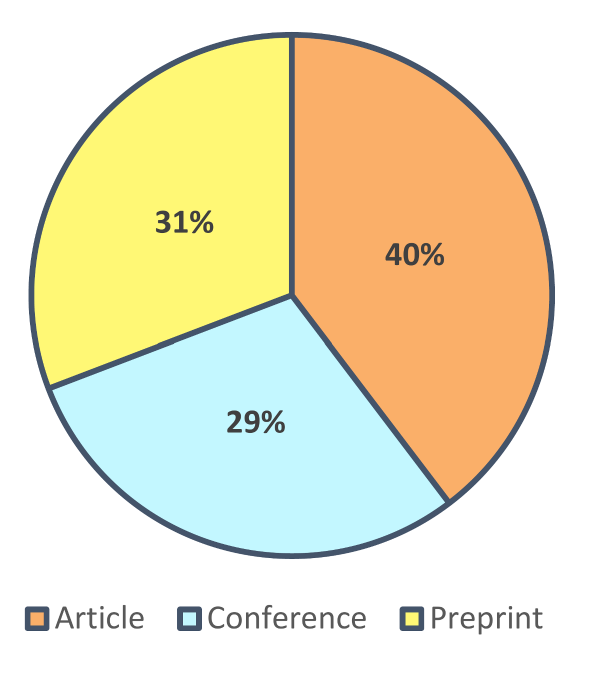 }
    \label{survey-paper-type}}

\caption{Quantitative Analysis of Surveyed Literature.}
\label{quantitative-survey-analysis}
\end{figure*}

\subsection{Survey Insights}

Dimensions remains our data source for selecting the surveyed paper. By leveraging the classification schemes obtained in the previous step, we formulated precise research queries relying on the third organizational level. In other words, we captured the terminology used to describe recent FL techniques as our research keywords. This methodology interestingly streamlined our research and limited the number of listed results. While at the same time, it enables us to emphasize the relevant studies that extensively dug into the underlying techniques.

The results of the numerical analysis conducted on the surveyed papers, which encompassed a substantial list of publications, are illustrated in Fig. \ref{quantitative-survey-analysis}. Fig. \ref{survey-publishers-insights} provides a percentage breakdown of the included literature on federated learning based on the respective publisher. Notably, more than two-thirds of the papers were found in reputable sources such as IEEE and arXiv, accounting for 34\% and 31\%, respectively. The remaining reviewed papers are distributed across other databases, including Springer, Elsevier, and ACM, with percentages of 9\%, 7\%, and 5\%, respectively. 

Fig. \ref{survey-years-insights} shows the distribution of publication years across the examined federated learning literature. Remarkably, 38\% of the papers were published in 2022, which aligns with our objective to provide an up-to-date study covering recent research advancements. Besides, 28\% of the works were published in 2023, as we continued to look for influential studies until the middle of this year. However, the proportion of papers published in 2021 and 2020 is relatively lower, accounting for 18\% and 16\%, respectively, as we excluded studies that were widely cited in earlier reviews. 
It is worth highlighting that while our primary emphasis on these statistics relies on the latest research findings, we also surveyed relevant papers predating 2020, which have significantly impacted the evolution of FL and garnered widespread adoption within the research community.

We present in Fig. \ref{survey-aspects-insights} the last percentage breakdown illustrating the top-level classification scheme clusters. From this figure, we can conclude that security and privacy have been investigated the most, comprising  41\% of the cited literature. The heterogeneity cluster follows closely with 39\%. The last line of our examination consists of communication efficiency with 20\% of the total content. 
Also, it is noteworthy that within our survey, 40\% of the integrated references are scholarly articles. While conference papers contribute to 29\% of the surveyed literature. Additionally, preprints constitute 31\% of the examined work, as illustrated in Figure \ref{survey-paper-type}.

\subsection{Survey Organization}
\begin{figure}[htbp]
    \centering
   \includegraphics[width=\textwidth]{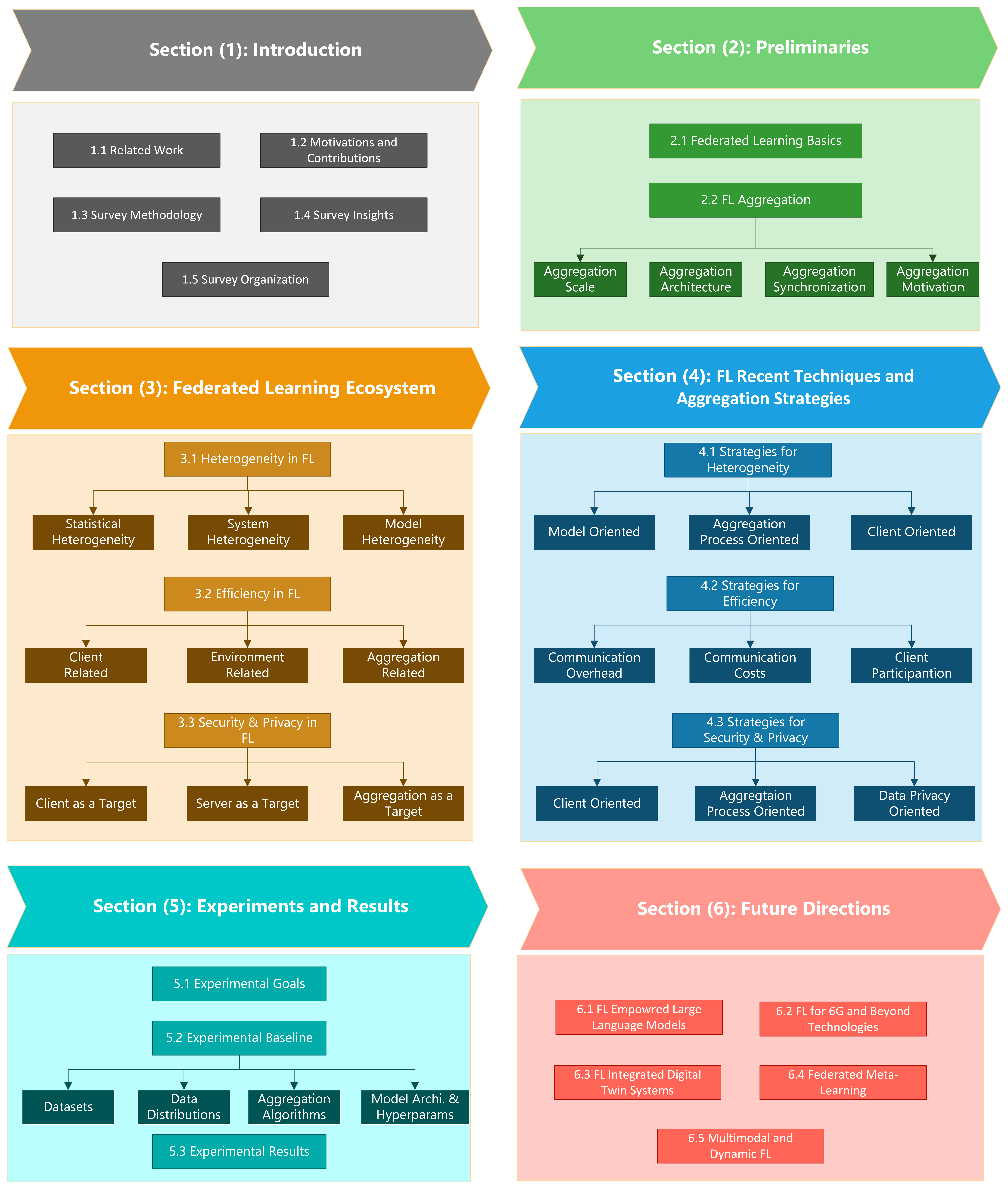}
    \caption{Our Survey Structure Organization.}
    \label{survey_organization}
\end{figure}

The remaining sections of this survey are structured as follows: In Section \ref{preliminaries}, we provide an overview of fundamental concepts in federated learning, covering its basic principles, formal definition, procedural steps, and various FL types based on data distribution. Additionally, we present a comprehensive classification of federated learning aggregation, emphasizing its pivotal role in overall system performance and reflecting our primary focus on studying the FL landscape. Section \ref{FL_ecosystem} delves into the characteristics of the federated learning ecosystem, exploring aspects such as heterogeneity, efficiency, security, and privacy. These discussions align with the three clusters identified in our high-level classification of the FL domain, spanning personalization, optimization, and robustness areas. Subsequently, Section \ref{fl_recent_techniques_and_aggregation} outlines the latest advancements in FL solutions, encompassing aggregation proposals as well as other relevant techniques that enhance the aggregation process. To enhance readability and navigability for readers, we offer a multi-level classification scheme for each FL cluster and summarize surveyed papers in dedicated tables, employing a multi-criteria framework for evaluation and comparison. Section \ref{FL_simulations} is dedicated to the experimental guidelines, simulation results, and interpretations. Specifically, we compare the performance of four FL aggregation algorithms and analyze their behavior across various real-world FL settings. Before concluding, Section \ref{FL_future_directions} highlights emerging trends and areas of interest, aiding researchers to pinpoint exciting directions for their future research endeavors. Finally, in Section \ref{FL_conclusions}, we discuss and recap our findings. Fig. \ref{survey_organization} provides an overview of the survey's organization.


\section{Preliminaries}
\label{preliminaries}

\subsection{Federated Learning Basics}
\begin{figure}[htbp]
    \centering
    \includegraphics[scale=0.5]{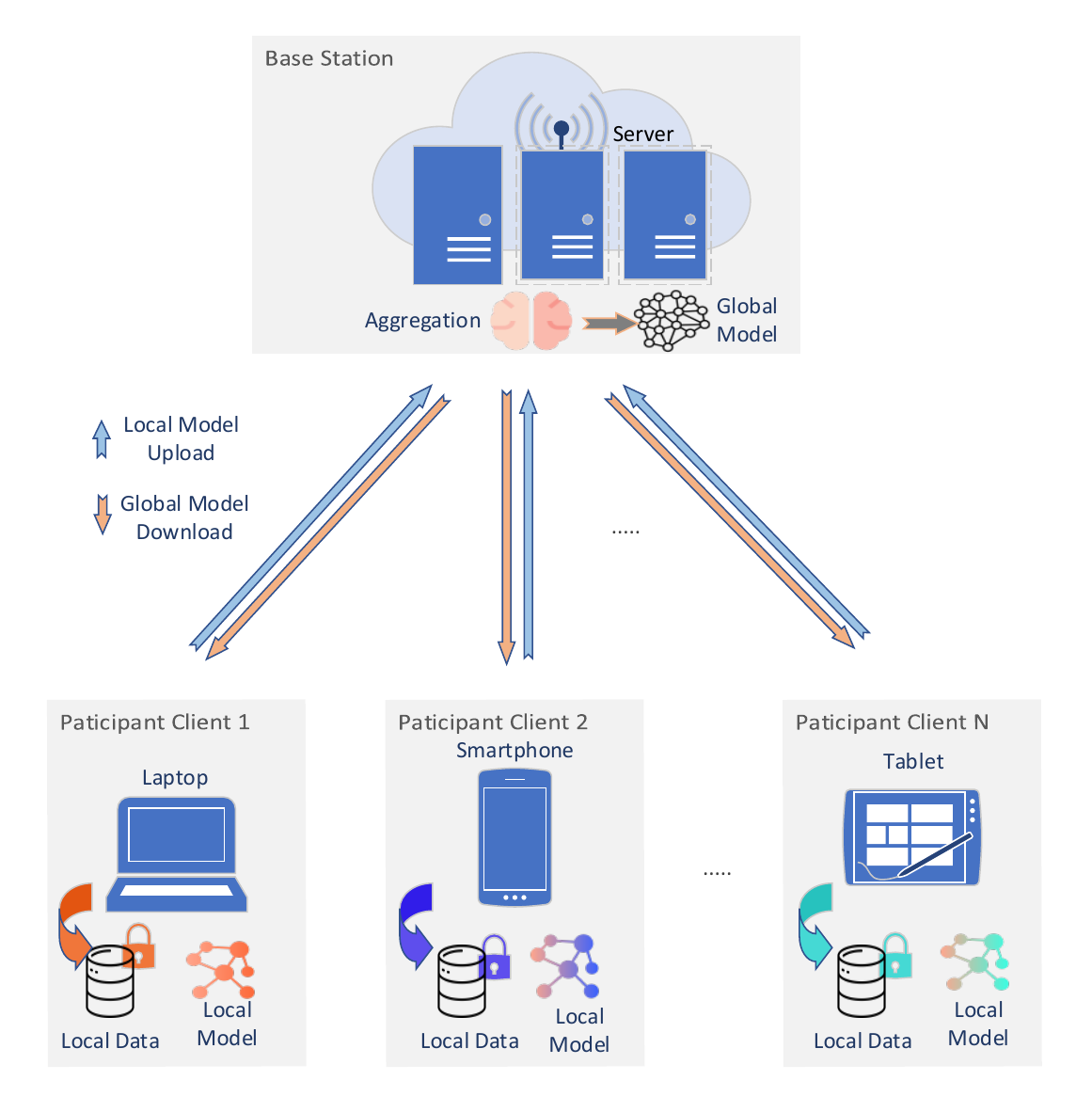}
    \caption{Generic Federated Learning Process.}
    \label{gene_FL_process}
\end{figure}

Federated Learning is a variant of Distributed Machine Learning paradigms where the data and the computational workload are distributed across multiple nodes connected to a network. This approach offers increased efficiency for training a robust and sophisticated ML model on large-scale datasets, which would be infeasible to process in one machine. Moreover, federated learning takes this concept further by focusing on data privacy. In the FL ecosystem, a set of participants called \textit{clients} collaboratively train high-quality AI models under the orchestration of a remote server called a \textit{parameter server} without the need to access their local data. The crucial aspect is that the private data never leaves the client site, but only the locally-built models are transmitted to the server. In response, the server aggregates these updates from all entities to form a global model and communicates it back to the active participants. The iterative process of local-global model exchanges, illustrated in Fig. \ref{gene_FL_process}, continues until achieving a desirable utility. 

\subsubsection{\textbf{FL Process}}
The generic FL procedure incorporates the following steps: 
\begin{enumerate}
    \item \textbf{System Initialization:} In the first phase, it is the responsibility of the server to decide the intended task application (e.g., disease prediction, system recommendation, activity detection) as well as the essential model parameters (model type and architecture, learning rates, number of clients per round). Furthermore, the server initiates the global model gradients and selects a group of clients to be involved in the next iteration.

    \item \textbf{Distributed Local Training:} The parameter server broadcasts the initial global model across the chosen clients to kickstart the distributed learning process. Upon receiving the global model, each node independently trains a local model leveraging its respective data to generate the updated gradients. The locally optimized models are then uploaded back to the server. 

    \item \textbf{Server Aggregation:} After receiving the updates from all participating clients, the server executes the aggregation algorithm to consolidate the gradients into a unified model (e.g., through averaging). Subsequently, the latest version of the global model is disseminated again to the involved entities. 
\end{enumerate}

Steps 2 and 3 are iterated until the desired performance is achieved.

In a formal context, we designate the selected client indices as \(c = 1, 2, 3, \ldots, C\), directed by the central server, where \(C \subseteq N\). Each node possesses its private dataset \(D_c = \{X_c, Y_c\}, \text{ where } X_c \in \mathbb{R}^{|D_c| \times d}\) represents the feature space vector, and \(Y_c \in \mathbb{R}^{|D_c| \times m}\) denotes the associated label matrix. During communication round $t$, client $c$ downloads the latest global model $w_t$ and uses its local data for training. The primary objective is to optimize a loss function that penalizes inaccuracies in the model's predictions for data points. Specifically, we denote \(l(W; x_i, y_i)\) as the loss function for the $i$-th data point, with W representing a matrix of model weights in a neuron network, then the mathematical equation for the local loss function of client c is given in Equation \ref{eq:client_loss}. 
\begin{equation}\label{eq:client_loss}
    L_c(W) = \frac{1}{|D_c|} \sum_{(x_i, y_i) \in D_c} l(W; x_i, y_i)
\end{equation}

Similarly, we present the global loss function encompassing all clients in Equation \ref{eq:global_loss}, wherein \(M = \sum_{c=1}^{C} |D_c|\) denotes the total number of data points across all C clients.

\begin{equation}\label{eq:global_loss}
    L(W) =  \sum_{c=1}^{C} \frac{|D_c|}{M} L_c(W)
\end{equation}

During the aggregation phase, the parameter server leverages an aggregation algorithm to generate the global model for round t+1 once all clients complete uploading their local models. The first widely-recognized method is FedAVG, which performs straightforward averaging of all the model weights as introduced in the pseudo-code \ref{fedavg_alg}. It is worth noting that the aggregation component plays a crucial role in FL since it facilitates the effective integration of knowledge learned by individual clients. Thus, it results in a more accurate and refined global model after each communication round. However, scholars have confirmed that this simplistic approach used in FedAVG may not fully address all the requirements of FL systems, which has spurred numerous research endeavors in this line of investigation.

\begin{algorithm}[htbp]
\scriptsize
\caption{FedAVG Algorithm}
\label{fedavg_alg}
\begin{algorithmic}[1]
\REQUIRE Set of clients $\{1, 2, \ldots, C\}$, learning rate $\alpha$
\STATE Initialize global model weights $W$ randomly
\FOR{$t=1$ to $T$}
    \FOR{each client $i$ in $\{1, 2, \ldots, C\}$}
        \STATE Receive global model weights $W_t$ from the server
        \STATE Update local model weights $W_i$ using client's data: $W_i \leftarrow \text{LocalUpdate}(W_t, D_{c_i}, \alpha)$
    \ENDFOR
    \STATE Calculate weighted average of local model weights: $W_{\text{avg}} \leftarrow \frac{1}{C} \sum_{i=1}^{C} W_i$
    \STATE Send $W_{\text{avg}}$ to the clients 
\ENDFOR
\end{algorithmic}
\end{algorithm}

\subsubsection{\textbf{FL Types}}

The key foundation of FL is the data matrix, which defines the distinct distribution patterns of data sample space and data feature space. As depicted in Fig. \ref{FL_types_data_partition}, FL can be classified into three categories depending on the data distribution of the participating clients, as follows:
\begin{figure*}[htbp]
    \centering
    \includegraphics[width=\textwidth]{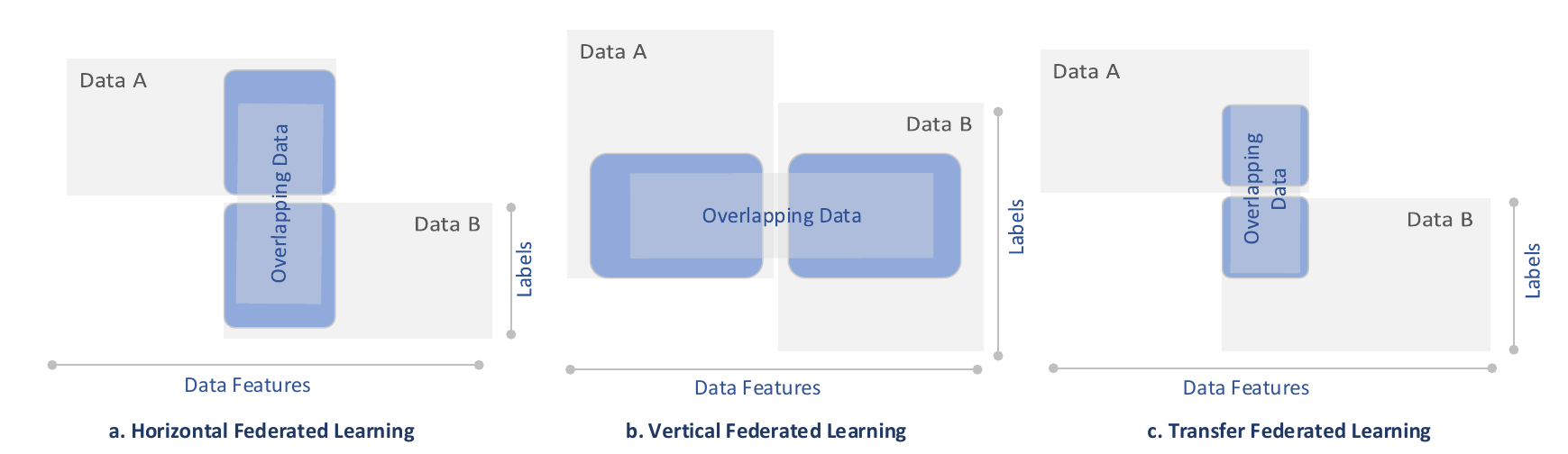}
    \caption{Different FL Types According to Client Data Partitions.}
    \label{FL_types_data_partition}
\end{figure*}

\begin{itemize}
    \item \textbf{Horizontal Federated Learning:} In Horizontal Federated Learning (HFL), the participating clients share the same type of data features while possessing different sets of data samples. To better understand this, let's consider a medical scenario where different hospitals collaborate to develop an AI model for predicting disease outcomes. In this case, all hospitals present the patients using their medical records, which form a common feature space. However, the patients associated with each hospital may differ, meaning they correspond to distinct populations with non-overlapping sets of patients. Despite this difference, HFL enables hospitals aimed at joining forces to collectively train a robust AI model using their shared feature space while respecting patients' privacy.

    \item \textbf{Vertical Federated Learning:} In Vertical Federated Learning (VFL), the datasets of the engaged nodes have a common simple space, but with different data features, which results in a combined dataset that is more diverse in terms of data types. For instance, consider a credit risk prediction task where the clients are banks within one country, each having a separate dataset. Some banks have stored the credit history data of their customers, while others have financial transaction records. In this case, there is substantial intersection at the level of clientele they serve. By pooling together the knowledge of their unique datasets, they can create a more comprehensive credit risk prediction model with improved accuracy.

    \item \textbf{Federated Transfer Learning:} Federated Transfer Learning (FTL) came to the fore when both HFL and VFL would not be effective. FTL can handle datasets in which the data features and samples are distinct across clients. When the intersection of the overlapping data samples and features is negligible considering all the participating clients, Transfer Learning methods are applied to map the various feature spaces into a new shared representation space to learn all the sample labels. 

\end{itemize}


\subsection{Federated Learning Aggregation}

\begin{figure}[htbp]
    \centering
    \includegraphics[scale=0.4]{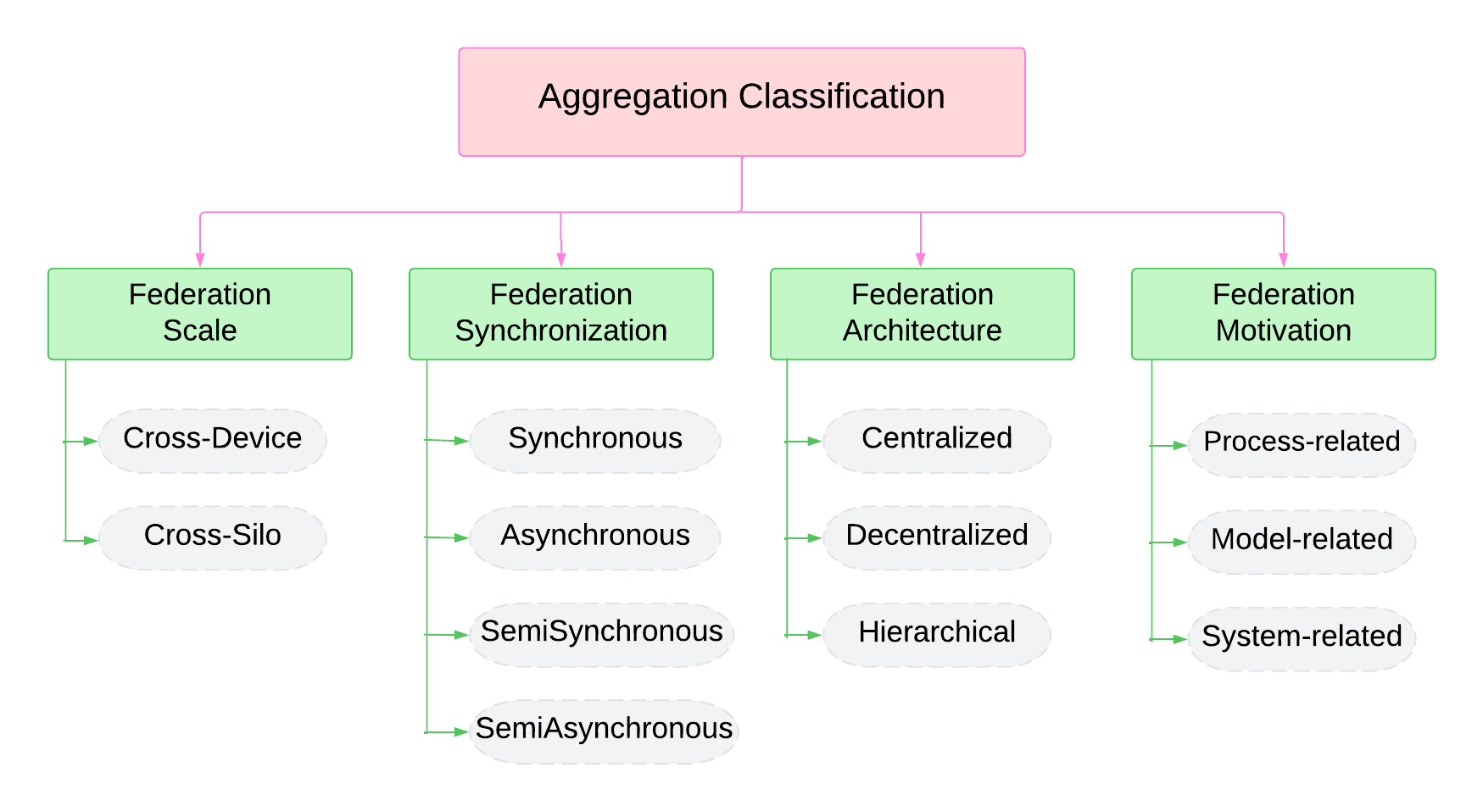}
    \caption{Broad Classification of Federated Learning Aggregation.}
    \label{classif-FL-aggreg}
\end{figure}

One of the crucial concerns of federated learning is how to combine the local models' updates from different clients into a global model that can generalize well to new data, regardless of the diversity of the participating parties. The aggregation methods are an integral part of the federated learning ecosystem as they provide a solution to this challenge. Furthermore, it has been determined that the aggregation techniques hold promise as a research direction to address the various challenges inherent to federated learning. Different algorithms have been proposed in the literature, ranging from simple averaging to more sophisticated techniques. Nonetheless, the choice of a specific one to adopt can significantly affect the performance of all the FL system evaluation metrics, from the convergence speed and accuracy to communication cost and privacy. Therefore, it is indispensable to understand the strengths and weaknesses of different aggregation strategies and select the proper option for the particular application and data distribution at hand. In this section, we will describe the aggregation process in federated learning, providing a comprehensive classification from various perspectives, according to the federated learning configuration, including the federation scale, the topology, the updates synchronization mode, and the motivation for joining the collaborative training. These perspectives are illustrated in Fig. \ref{classif-FL-aggreg}. Later in the paper, we will review the latest advancements in aggregation techniques put forth in scholarly publications and delve into their respective methodologies that aim to fulfill the FL requisites (see Section \ref{fl_recent_techniques_and_aggregation}). 

\subsubsection{\textbf{Federation Scale}}
\begin{figure*}[htbp]
    \centering
     \includegraphics[width=\textwidth]{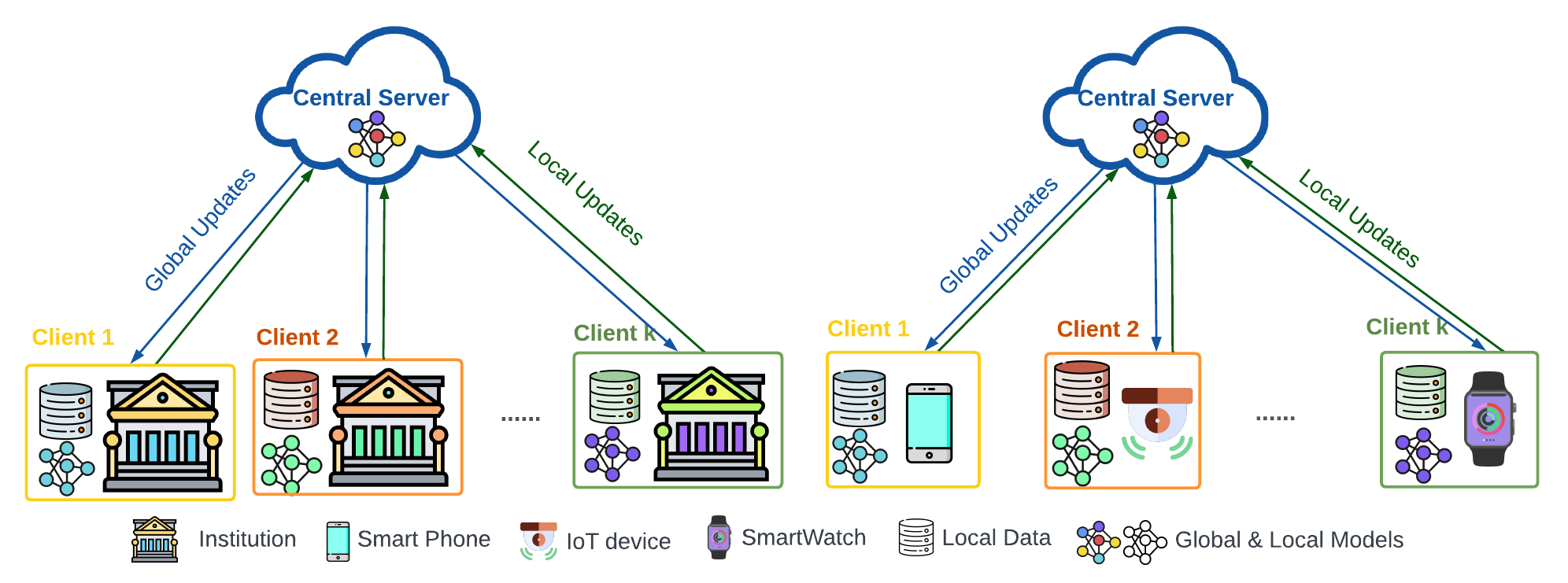}
     \caption{Cross-Silo and Cross-Device Aggregation in Federated Learning.}
     \label{FL_cross_silo_cross_device}
\end{figure*}

Aggregation methods in federated learning can be classified based on the number of active participants and the amount of data held by each one.  Two main categories of aggregation methods based on the scale of the federation and the data distribution are \textit{cross-device} aggregation and \textit{cross-silo} aggregation, as depicted in Fig. \ref{FL_cross_silo_cross_device}. 
Table \ref{tab-FL-scale} summarizes the unique features of each category, facilitating a clear understanding of their differences.

\begin{itemize}
    \item \textit{\textbf{Cross-Device Aggregation.}}
    The advent of federated learning stemmed primarily from the need to facilitate efficient machine learning on mobile and edge devices, which typically possess constrained computational capabilities and limited storage capacity. In this cross-device setting, local models are trained on data generated by individual users, making it well-suited for scenarios where FL users are mobile, and their number is quite large, practically in the range of \num{e10}. Notably, the first example of cross-device federated learning was Google's implementation of the Gboard mobile keyboard to build next-word prediction models.

    \item \textit{\textbf{Cross-Silo Aggregation.}}
    Given the promising outcomes of cross-device FL applications, a growing interest in extending FL usage into other applications has rapidly emerged. Cross-silos FL refers to the collective training between large institutions in various domains, including smart manufacturing, finance risk prediction for reinsurance, and medical data segmentation. The participants in cross-silo scenarios possess significant amounts of data and are relatively limited in number, typically less than 100 clients. In this context, it is reasonable to expect organizations such as hospitals and banks to be consistently involved in each round, owing to their substantial computing capabilities and relatively stable environments.
    
\end{itemize}

\begin{table} [htbp]
\scriptsize
\caption{Summary of Different Federation Scale Characteristics.}
\label{tab-FL-scale}

\begin{tabularx}{\textwidth}{| >{\hsize=.1\hsize\linewidth=\hsize}X|  >{\hsize=.1\hsize\linewidth=\hsize}X|  >{\hsize=.25\hsize\linewidth=\hsize}X | >{\hsize=.15\hsize\linewidth=\hsize}X|  >{\hsize=.1\hsize\linewidth=\hsize}X|  >{\hsize=.15\hsize\linewidth=\hsize}X|}
\hline  
 \textbf{Federation Scale} & \textbf{Client's Nature}  & \textbf{Available Resources} & \textbf{Local Data} & \textbf{Participant Number} & \textbf{Security \& Privacy} \\ 
\hline
\textbf{Cross-device} 
&  Mobile and IoT devices
&  Limited computational resources, storage, and battery lifespan 
&  Relatively small datasets
&  Up to millions
&  More vulnerable and challenging\\
\hline
\textbf{Cross-silo}
& Organizations
& Powerful computation and storage capabilities 
& Potentially large data
& No more than 100 clients
& Easier to ensure  \\
\hline
\end{tabularx}
\end{table}

\subsubsection{\textbf{Federation Architecture}}
\begin{figure}[htbp]
    \centering
    \includegraphics[width=\textwidth]{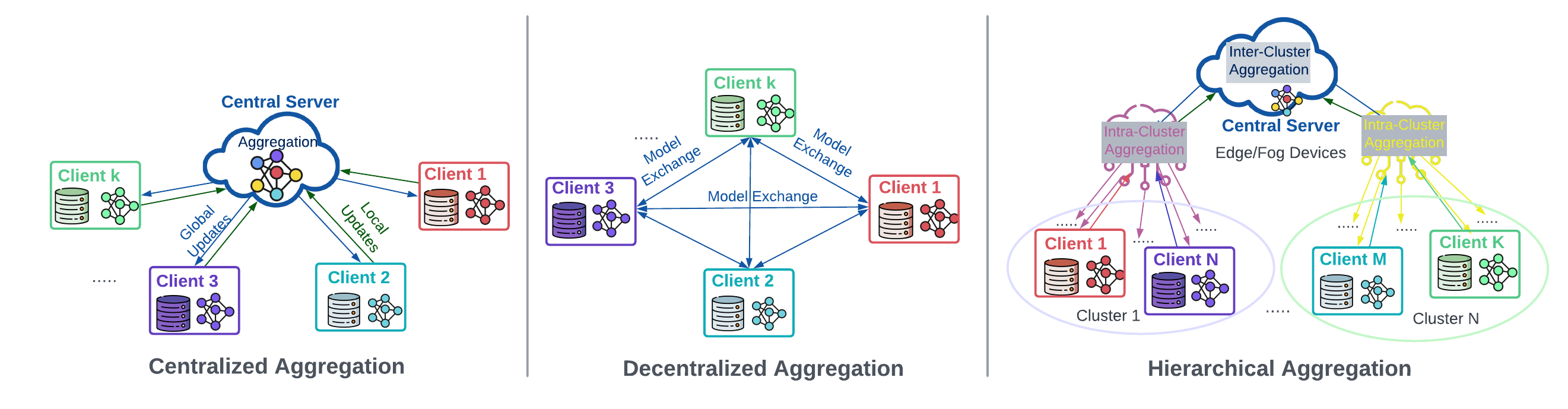}
    \caption{Types of Federated Learning Topology Architecture.}
    \label{FL_architectures}
\end{figure}

The architectural aspect of the federated learning process is also an essential factor for aggregation classification. The network topology that defines how the involved parties interact with each other or with a central server might be either centralized, decentralized, or hierarchical. Fig. \ref{FL_architectures} presents an illustration of the three architectures. While Table \ref{tab-FL-archi} sheds light on their characteristics, respectively.

\begin{itemize}
    \item \textit{\textbf{Centralized Federation.}}
    The centralized design is the most commonly used FL architecture, where a central server is responsible for learning coordination. This means that the client devices interact only with the server to send and receive model updates, in a synchronous or asynchronous communication regime. The single-server topology ensures that the whole training process is in the hand of one powerful actor, which helps avoid errors and facilitate a simple aggregation pipeline. However, it demands a robust and secure server that can handle extreme conditions (e.g., the high number of participants, the large model updates dimension) and alleviate the potential security and privacy attacks.

    \item \textit{\textbf{Decentralized Federation.}}
    In decentralized FL architecture, clients can communicate with each other in a peer-to-peer (P2P) fashion to build a global model without needing a central server. The main idea behind this design is to release the dependency on a central entity. Besides, it weakens the impact of the server becoming malicious or curious while ensuring the model's utility. On the flip side, it can become challenging to implement the Fl aggregation under a decentralized architecture that converges efficiently, especially in large-scale FL.

    \item \textit{\textbf{Hierarchical Federation.}}
    The novelty of hierarchical federation has come to the fore from the need to tackle the limitations of centralized and decentralized settings, specifically in scenarios where the participating clients are distributed across multiple levels or tiers of a hierarchy. This new approach allows for multi-level coordination by introducing an intermediate layer above the local clients for first-level aggregation, followed by a global one. Accordingly, the local workers are grouped in clusters considering different criteria, such as locations, data distributions, etc., and they only communicate with their associated parameter server, usually an edge or fog device, which is responsible for averaging the respected local updates. After a number of intra-cluster iterations,  an inter-cluster model aggregation is performed, usually in the cloud, to establish a global consensus. Hence, the tree-like structure introduced by the hierarchical structure can reduce communication overhead, enhance privacy protection, and improve participation flexibility and scalability.
    
\end{itemize}

\addtocounter{table}{-1}

\begin{table*} [htbp]
\scriptsize
\caption{Summary of Different Federation Architecture Characteristics.}
\label{tab-FL-archi}
\begin{tabularx}{\textwidth}{| >{\hsize=.1\hsize\linewidth=\hsize}X|  >{\hsize=.45\hsize\linewidth=\hsize}X|  >{\hsize=.45\hsize\linewidth=\hsize}X | }
\hline  
\textbf{Architecture} & \textbf{Pros}  & \textbf{Cons }\\
\hline
\textbf{Centralized} 
&  - Simple to implement 

- Easier to manage 

- Faster convergence 

- More accurate and consistent model 

&  - Risk of Privacy breaches 

- Risk of single point of failure 

- Less scalable  for large-scale FL \\
\hline

\textbf{Decentralized}
&  -  More scalable 

- More resilient to system failures 

- No single point of failure 

&  - Slower convergence 

- Difficult to coordinate, manage, and implement  

- No stored global model (distributed across all devices) 

- Difficult to ensure models accuracy  \\
\hline

\textbf{Hierarchical} 
& - Highly scalable 

- Flexible client participation 

- Reduced communication overhead by careful design 

- Balance between privacy and accuracy
&  - More complex implementation 

- Requires additional infrastructure  - Potential for hierarchical biases \\ 
\hline
\end{tabularx}

\end{table*}

\subsubsection{\textbf{Federation Synchronization}}
The synchronization mode in federated learning refers to how the involved clients synchronize their local model updates with the aggregator node. We can find four distinct ways for synchronizing the aggregation process, including synchronous, asynchronous, semi-synchronous, and semi-asynchronous aggregation. We will elucidate in the following the advantages and disadvantages of each mode and summarize the provided description in Table \ref{tab-FL-synchro}.

\addtocounter{table}{-1}

\begin{table*} [htbp]
\scriptsize
\caption{Summary of Pros and Cons of Different Federation Synchronization Modes. }
\label{tab-FL-synchro}
\begin{tabularx}{\textwidth}{| >{\hsize=.19\hsize\linewidth=\hsize}X|  >{\hsize=.43\hsize\linewidth=\hsize}X|  >{\hsize=.38\hsize\linewidth=\hsize}X |}
\hline  
\textbf{Synchronization Type} & \textbf{Pros}  & \textbf{cons} \\
\hline
\textbf{Synchronized} 
&  - Easier to implement and manage 

- Guaranteed convergence 

- Suitable for small datasets 

&   - Sensitive to stragglers 

- Less scalable to large and heterogeneous clients

- Potent devices are under-used \\
\hline

\textbf{Asynchronized}
&  
- Flexibility in scheduling client updates 

- More tolerant to device heterogeneity (stragglers) 

- Less communication overhead

&  - Stale updates issue 

- Convergence not guaranteed  

- May suffer from lower convergence rate 

- Difficult to ensure models accuracy  \\
\hline

\textbf{Semi-Synchronized} 
&  - Reduces communication overhead (vs. synch. methods) 

- More resilient to stragglers (v.s synch. methods)

- Stronger convergence guarantee (vs. asynch. methods) 

&  - Requires careful tuning for better trade-offs 

- More difficult to implement (vs. asynch. methods) \\
\hline

\textbf{Semi-Asynchronized} 
& - Reduce communication overhead (vs. synch. methods) 

- Improved convergence  (vs. asynch. methods) 

- Can handle more heterogeneous client

&  - Requires careful hyperparameters tuning 

- Stale gradients issue 

- May suffer from slower convergence \\
\hline

\end{tabularx}
\end{table*}

\begin{itemize}
    \item \textit{\textbf{Synchronous Aggregation}}.
    Federated learning is widely employed using the synchronous policy, where the participating devices upload their locally trained models simultaneously. By adopting this approach, the server performs the aggregation operation only after receiving updates from all clients, ensuring consistency in model updates and yielding improved accuracy. However, this synchronous aggregation method entails a significant drawback in the form of high communication overhead. This overhead can adversely impact the convergence speed, particularly in heterogeneous and expansive environments (e.g., when handling massive edge IoT devices).

    \item \textit{\textbf{Asynchronous Aggregation}}.
     The primary objective behind introducing asynchronous FL aggregation is to handle the stragglers' issue in cross-device FL settings and to mitigate scalability concerns. In this approach, clients autonomously update their local models and communicate with the aggregator (e.g., the server) whenever available. Similarly, the server independently aggregates the received updates. As a result, asynchronous aggregation substantially diminishes the communication overhead compared to its synchronous counterpart, as clients do not need to await others' updates. Yet, it is worth noting that this approach may lead to slower convergence, decreased accuracy, and a less stable model. The potential flaws arise from the fact that the updates might be outdated or conflicting, introducing uncertainties and complexities into the aggregation process.

    \item \textit{\textbf{Semi-synchronous Aggregation}}.
    It represents a hybrid approach that integrates synchronous and asynchronous aggregation techniques, offering a middle ground between their benefits. In semi-synchronous aggregation mode, the server waits for a subset of clients to complete their computations before aggregating their parameters. 
    This approach strives to strike a delicate balance between convergence speed and communication overhead.

    \item \textit{\textbf{Semi-asynchronous Aggregation}}.
    This approach is akin to the process employed in semi-synchronous aggregation in how it combines synchronous and asynchronous perspectives. Regardless, what sets the semi-asynchronous method apart lies in which part of the aggregation pipeline is synchronous and which part is asynchronous. In a typical implementation of this approach,  each client device conducts a specific number of local iterations, updating its model parameters asynchronously. Then, the clients upload their updates to the central server, which performs synchronous aggregation at fixed intervals. The challenge, however, resides in the optimal synchronization interval choice that requires careful calibration. 
\end{itemize}

\subsubsection{\textbf{Federation Motivation}}
\label{FL-motivations}

As a distributed machine learning paradigm, federated learning has been the subject of extensive research aimed at addressing not only privacy-preserving concerns but also attaining numerous other remarkable advancements and overcoming traditional limitations in machine learning. Proposing an aggregation method to serve these ambitious goals has drawn significant attention recently since it directly and potentially impacts the overall system performance. Moreover, the unique characteristics of each aggregation approach make it more suitable for some applications and environments, and less for others. This subsequent section highlights the motivations behind proposing various aggregation methods in the literature. Furthermore, to effectively categorize the diverse research visions, we have structured them into three essential families: process-related, model-related, and system-related objectives, as tabulated in Table \ref{tab-FL-motivation}. Later in the paper (see Section \ref{fl_recent_techniques_and_aggregation}), we will present an extensive collection of the existing aggregation algorithms and evaluate each solution based on the associated configuration and the confirmed goals. \\

\addtocounter{table}{-1}

\begin{table*} [htbp]
\scriptsize
\caption{Summary of Different Federation Motivations and Their Impact on The FL Aggregation.}
\label{tab-FL-motivation}
\begin{tabularx}{\linewidth}{| >{\hsize=.15\hsize\linewidth=\hsize}X|  >{\hsize=.425\hsize\linewidth=\hsize}X|  >{\hsize=.425\hsize\linewidth=\hsize}X |}
\hline  
\textbf{Motivation} & \textbf{Description}  & \textbf{How it helps FL aggregation}\\
\hline
\textbf{Process-Related} & \multicolumn{2}{c|}{} \\ \hline
\textbf{Communication Efficiency}
& - Minimize the communication load 

- Optimize the communication costs

& - Speed up the convergence rate

- Improve FL scalability \\ \hline

\textbf{Computation Efficiency}
& - Efficient utilization of computational and storage capabilities 

& - Improve FL feasibility \& scalability 

- Optimize the energy consumption  \\ \hline

\textbf{Convergence Rate} 
&  - Speed up the training process to achieve the desirable model utility 

&   - Optimize the consumption of available resources \\ \hline
\textbf{Model-Related} & \multicolumn{2}{c|}{} \\ \hline

\textbf{Personalization} 
& - Tailor the global model to the diversified client requirements and data distributions 

&  -  Enable learning customized models for distinct clients \\  \hline

\textbf{Generalization} 
& - Ensure that the global model can generalize well to unseen data 

& - Maintain good performance on a wide range of devices and application settings \\ \hline

\textbf{Regularization} 
&  - Penalize clients those updates fall too far from the global model 

&  - Enhance the generalization performance - Prevent "client drift" issue  \\ \hline

\textbf{Fairness} 
& - Ensure that the global model does not discriminate against some groups of people based on their localization, sex, etc.

& - Bias mitigation - Create models that are equally effective for all clients \\  \hline

\textbf{Heterogeneity} 
&  - Consider the statistical and system heterogeneity

& - Build more robust and consistent models

-  Improve FL feasibility under real-life settings  \\ \hline
\textbf{System-Related} & \multicolumn{2}{c|}{} \\ \hline
\textbf{Security} 

&  - Implement security measurements accounting for all the potential attacks during the different training stages 

&  - Guarantee the FL resilience against adversarial attacks \\ \hline

\textbf{Privacy} 
&  - Preserve the users' data, local updates, and final deployed model from information disclosure

&  - Encourage stakeholders to join the collaborative learning  - Enable high-quality data exploitation  \\ \hline

\textbf{Scalability} 
&  - Ensure optimal performance under challenging circumstances

&  - Ensure the FL system's extensibility to large model and millions of users \\  \hline

\end{tabularx}
\end{table*}

\section{Federated Learning Ecosystem}
\label{FL_ecosystem}

\subsection{Heterogeneity In Federated Learning Ecosystem}
\label{heteroheneity_in_FL}

\begin{figure}[htbp]
    \centering
    \includegraphics[scale=0.4]{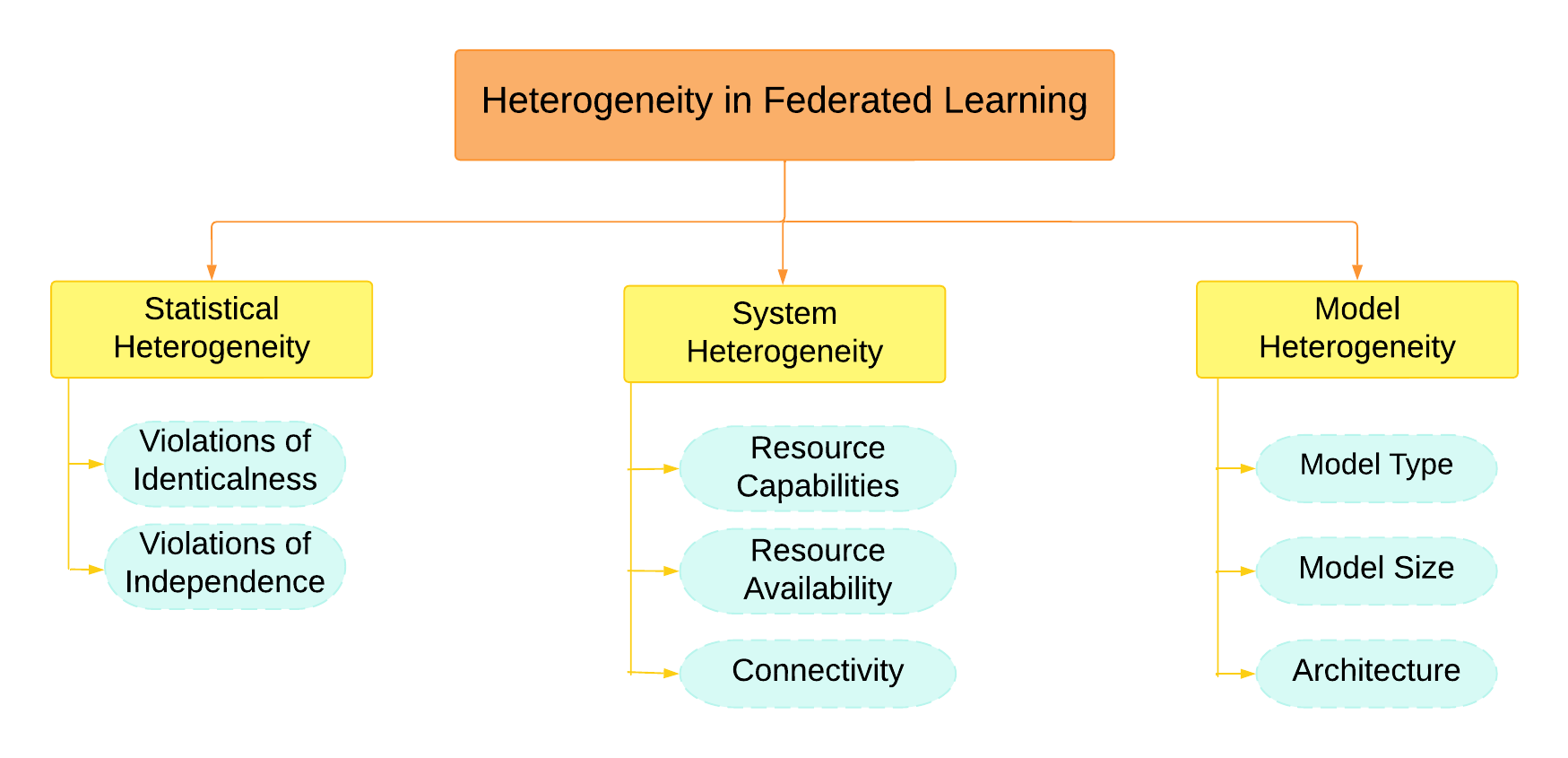}
    \caption{Classification of Heterogeneity Types in Federated Learning.}
    \label{heterogeneity-classes}
\end{figure}

The problem of heterogeneity in the FL settings has been extensively studied over the past few years \cite{gao2022survey, xu2021asynchronous}. It arises from the fact that the clients involved in the training process often have different systems and data characteristics, despite their shared goal of building a robust and powerful intelligent model.
To offer a broader perspective, we classify the encompassed types of heterogeneity in FL into three distinct categories: \textit{(i) statistical heterogeneity}, \textit{(ii) system heterogeneity}, and \textit{(iii) model heterogeneity}. In the following,  we will delve into the sources of heterogeneity as one of the most demanding challenges. The provided classification is summarized in Fig. \ref{heterogeneity-classes}. 

\subsubsection{\textbf{Statistical Heterogeneity}}
Statistical heterogeneity refers to the non-independent and not identically distributed (non-IID) nature of data \cite{li2020federated}. The issue of non-IID data emerges as a consequence of the data generation paradigm in the context of FL systems. The traditional ML models assume the data to be in IID distribution, which stands in contrast to the real-life situation of collaborative training, where each device collects data differently depending on the user's needs, preferences, location, and available resources \cite{criado2022non}. As a result, the assembled datasets will contain dissimilar data features, target class distributions, and even unbalanced sizes. Technically, it means that the data held locally in a single client can not be representative of the overall data distribution across all the active participants. Hence, the development of an effective federated learning framework will be compounded by an additional layer of complexity. The challenge is to propose a global model that can satisfy both a high level of generalization and an optimal personalization that accounts for the unique characteristics of each client \cite{tan2022towards}.

To dig deeper into the possible causes of statistical heterogeneity we divide the non-IID data distribution into two classes: \textit{violations of identicalness} and \textit{violations of independence}. 
\paragraph{Violations of Identicalness}
    As mentioned in \cite{kairouz2021advances}, the non-Identical data can pertain to five distinct subclasses: 
    \begin{itemize}
        \item \textit{Feature distribution skew}.
        This scenario occurs when the input feature space varies across clients even if their knowledge is the same.
        \item \textit{Label distribution skew}.
        It arises due to regional differences, which can impact the label distribution.
        \item \textit{Same label, different features}.
        For the same label, disparate vectors of input features may be associated.
        \item \textit{Same features, different labels}.
        Since users have personal preferences and geo-regions, one vector of input feature may lead to distinct labels.
        \item \textit{Quantity skew or unbalancedness}.
        Clients may have vastly varying amounts of collected data, depending on their device capabilities.
    \end{itemize}
\paragraph{Violations of Independence}
    Besides the effects of non-identical data distribution, there are common patterns of data deviation from being independently drawn from an overall distribution. For example, if the data is in an insufficiently-random order (e.g. ordered by time or/and by localization), it will cause a cycling situation in which a regime of data sample permutation is established. 

\subsubsection{\textbf{System Heterogeneity}}

The heterogeneity of the federated learning ecosystem may also stem from the varying nature of the device systems employed by each client. When a group of devices trains over different datasets, they often use varying hardware. The heterogeneity of device systems can be observed through differences in hardware capabilities (CPU, GPU, memory), network connectivity (4G, 5G, WiFi), and resource availability (battery power and lifespan) \cite{jiang2022towards, fedlbs24iwcmc}. This diversity leads to disparate computation and storage capacities, communication efficiency, and resource availability across a vast range. To illustrate the major issues resulting from this dispersion of client systems, we consider the following situations \cite{xu2021asynchronous}:
\paragraph{The stragglers problem}
    
    The issue of stragglers is a common problem in FL settings. It arises in collaborative training as devices often suffer from slow network connectivity and constrained resource availability. Consequently, they may become offline unexpectedly due to unreliable connections. These conditions hamper their ability to be constantly active in the training process. The described phenomenon is referred to, as the "stragglers issue" \cite{mcmahan2017communication} or "clients dropout" \cite{aledhari2020federated} where the server is compelled to wait inefficiently for those unreliable clients to upload their local updates.
    
\paragraph{Low round time efficiency}
    
    Without adopting an effective mechanism for client selection, faster clients who have completed their local training are held up by slower or unreliable clients who have fallen behind \cite{wang2021field}. Thus, the aggregation phase will witness significant delays leading to an undesirable impact on the convergence rate of the global model, or even worse, impeding it from convergence at all.
    
\paragraph{Low resource utilization}
    
    A poor client selection strategy can also result in infrequently picking competent devices to join a communication round. It means that the exceptional computational abilities of some devices will not be thoroughly exploited through the classical FL protocol.

\subsubsection{\textbf{Model Heterogeneity}}
Whilst federated learning assures the data privacy held separately on clients' sites, the potential privacy threats exposed by sharing only model parameters have hindered certain business advancements from seeing the light \cite{wahab2021federated}. In fact, given the inherent heterogeneity of data and device resources in FL environments, the objective and motivation behind joining such collaborative training may differ across various parties, depending on their specific requirements \cite{li2021survey}. Model heterogeneity refers to the client's willingness to design their separate local model independently from each other. However, the traditional averaging-based aggregation method lacks flexibility when each client trains its model in a black-box manner to the other clients. To offer a more comprehensive understanding of this type of heterogeneity, we organize it into three subclasses that often exhibit simultaneously in real-world scenarios.

\begin{itemize}
    \item \textit{Type-based model heterogeneity}.
    
    Various machine learning models can be appropriate for the same task at hand. The selection of the best-fitted model depends on the type of data features and characteristics of each client. Some may opt for a \textit{Linear} model (e.g. Linear Regression, Ridge Regression), While others may choose to adopt a \textit{Deep Learning} model such as a Neural Network (e.g. CNN, DNN), or \textit{Tree} model (e.g. Decision Tree, Random Forest) \cite{zhang2021survey}. Consider, for instance, two hospitals with different data types: one may hold a dataset of images such as CT scans and MRIs. At the same time, the other entity may possess medical records in tabular form. In such cases, regardless of sharing the same goal of prediction, the former may prefer to adopt a Convolutional Neural Network (CNN), while the latter may find a Decision Tree model more appropriate for their data type.

    \item \textit{Size-based model heterogeneity}.
    
    Due to system heterogeneity, adjustments to the trained model size are desirable to match the available resource restrictions \cite{gao2022survey}. For example, we can build a human activity recognition model using data collected from smartphones and wearable devices. Although smartphones have powerful computation resources, wearable devices can capture more precise movement data with limited computational resources. Consequently, a larger model can be trained for smartphones, whereas a lightweight model is more appropriate for wearable devices.
    
    \item \textit{Architecture-based model heterogeneity}.
    
    When aiming for an FL solution, even if the participating entities have reached a consensus on the choice of model type and dedicated resources, they may still be unwilling to divulge the details of their model architecture.  Instead, they may prefer to execute the FL process in a black box fashion without communicating any information about the neural network's depth, the loss function, the optimizer, or any other pertinent properties.
\end{itemize}

To summarize, handling the diverse heterogeneity types in FL is a growing consideration. Nonetheless, implementing a tailored aggregation method that can uphold this diversity is still a significant hurdle to overcome \cite{gao2022survey}.

\subsection{Communication Efficiency in Federated Learning Ecosystem}
\label{communication_efficiency_in_FL}

Enhancing model performance and dealing with substandard data is not exclusive to federated learning, rather, these challenges are present in various machine learning techniques. Nevertheless, the peculiarity of federated learning's distributed nature brings a unique challenge to communication efficiency \cite{almanifi2023communication}. Adopting a federated learning mechanism, as opposed to the traditional centralized ML settings, mandates scrutiny of the updates' transmission efficiency. On the one hand, the federated learning architecture enables the synchronization of training models across multiple sites, resulting in reduced computation time. However, on the other hand, this can lead to an onerous increase in communication costs as the convergence rate slows down due to the large exchanged model size or the restricted qualification of the active devices \cite{sun2021decentralized}. 

One communication round in federated learning consists of two types of transmission:  \textit{upstream communication} and \textit{downstream communication}. Upstream communication describes the process wherein clients transmit the updated local model to the server. Conversely, downstream communication pertains to the procedure by which clients acquire the current global model from the server. These operations are reiterated until a powerful global model is achieved. Several survey papers have approached the optimization challenge of achieving efficient communication from various standpoints \cite{shahid2021communication, jiang2022towards, zhao2023towards, almanifi2023communication}. According to \cite{almanifi2023communication}, communication is deemed efficient by the ability to transmit data through a reliable channel while expending minimal energy. More technically, efficiency is attained by ensuring that data exchanges in both downstream and upstream channels incur no overhead and consume the minimum resources. 

\begin{figure}[htbp]
    \centering
    \includegraphics[scale=0.4]{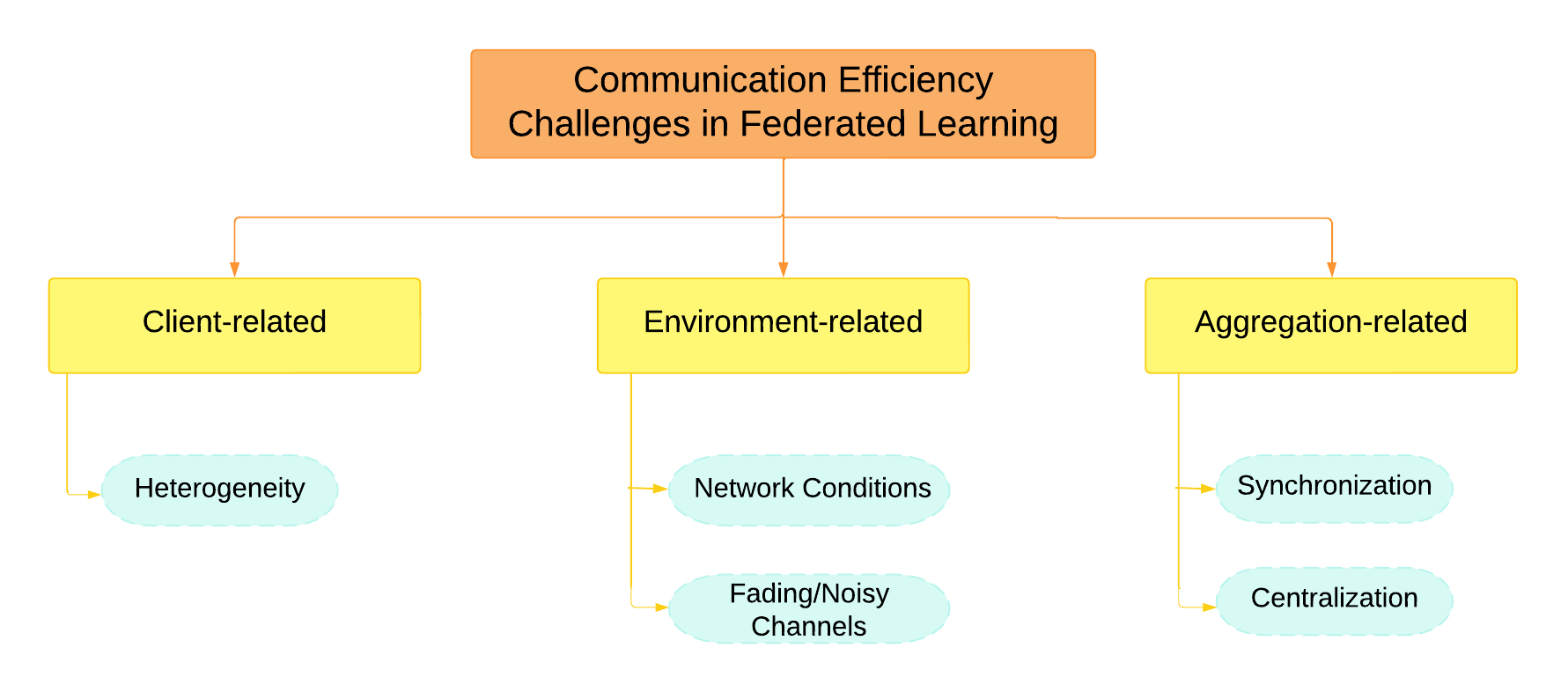}
    \caption{Sources of Communication Efficiency Issue in Federated Learning.}
    \label{Comm-Effi-causes}
\end{figure}

The present section will examine the impediments that stunt efficient communication between clients and servers. Fig. \ref{Comm-Effi-causes} depicts these obstacles in three distinct classes. 


\subsubsection{\textbf{Client-related challenges}}
As stated in section \ref{heteroheneity_in_FL}, the coexistence of various types of heterogeneity represents a formidable challenge to the successful implementation of federated learning. More specifically, it is well known that machine learning techniques are intrinsically power-intensive, and the concomitant presence of heterogeneous data, models, and device resources exacerbates the issue of attaining a desirable communication efficiency. For instance, in situations where millions of devices communicate with a remote server with no selection scheme, some devices may experience battery drain and drop out of training, while others may remain idle despite possessing superior competencies. This leads to a significant deterioration in communication efficiency. 
    
\subsubsection{\textbf{Environment-related challenges}}
The network conditions, including bandwidth, reliability, and connectivity, can be a source of disruption in the federated learning process \cite{10183138}. This is because some participating devices may suffer from insufficient bandwidth, which results in unreliable communication with the server. Additionally, differences in upload and download speeds, or disparities in the overall network reliability across selected clients, can give rise to potential bottlenecks that impede the learning process \cite{shahid2021communication}. Furthermore, the communication channels are often noisy and fading, which slows down the convergence rate and reduces the global model performance \cite{zhao2023towards}.
    
\subsubsection{\textbf{Aggregation process-related challenges}}
Traditional federated learning, characterized by synchronized and centralized aggregation, presents significant communication hurdles. These challenges can manifest in several forms, such as the stragglers' occurrence, inefficient energy utilization, fluctuating network conditions, and disparities in resource availability among devices, leading faster devices to remain idle for extended periods. Therefore, it becomes imperative to explore alternative aggregation mechanisms as viable solutions to facilitate efficient communication \cite{xu2021asynchronous}.

\subsection{Security and Privacy in Federated Learning Ecosystem}
\label{security_privacy_in_FL}

Disseminating model parameters in a distributed environment introduces novel risks and vulnerabilities to the overall system's security and privacy \cite{bouacida2021vulnerabilities}. Given that the federated learning paradigm is preferred when security and privacy concerns are of great importance, it is crucial to draw attention to these challenges and explore the available defenses and privacy-preserving techniques. 
Generally, the security and privacy attacks can be attributed to three possible adversaries: a malicious server, an insider adversary, or an outsider adversary. 

\begin{figure}[htbp]
    \centering
    \includegraphics[scale=0.4]{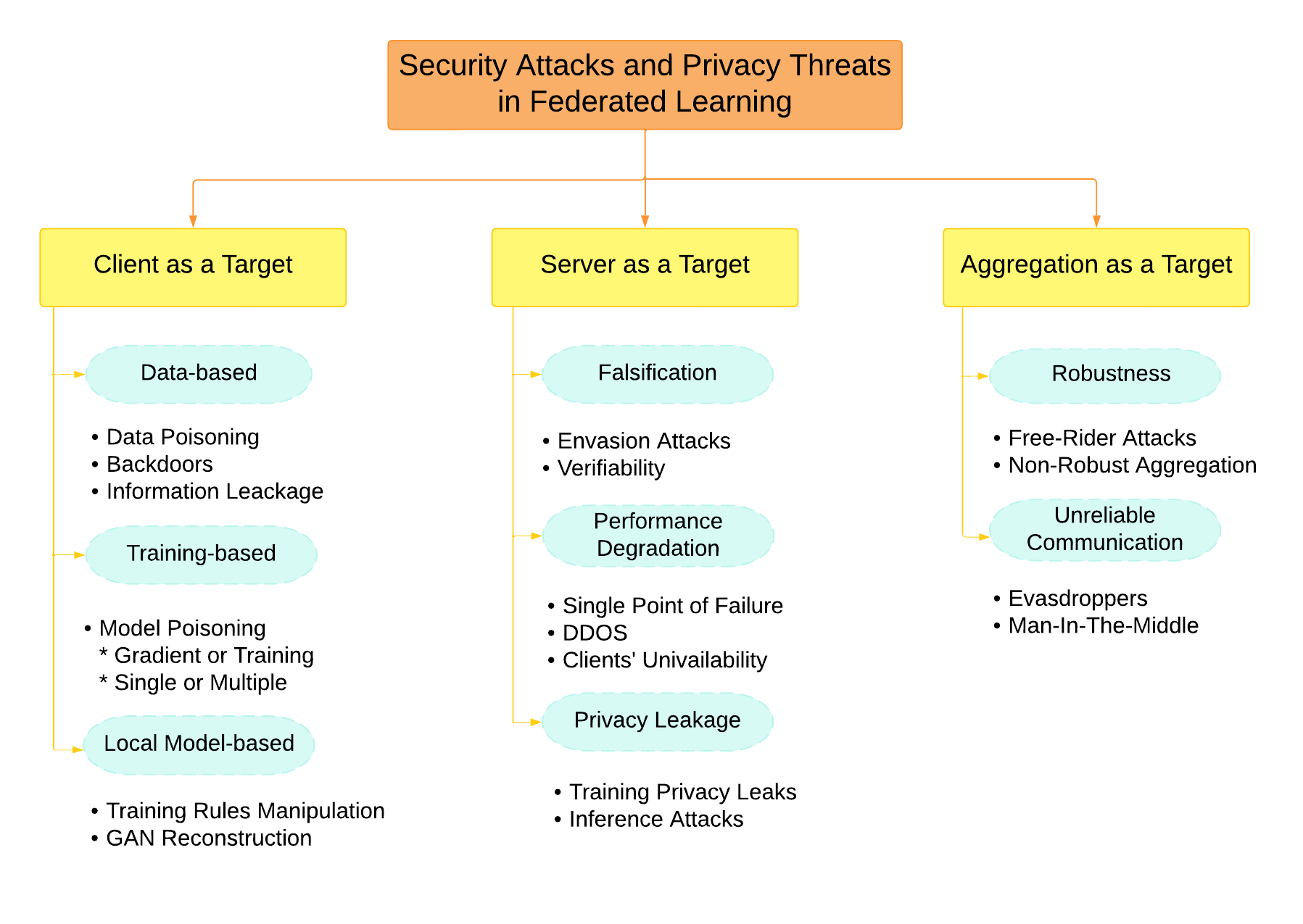}
    \caption{Classification of Security and Privacy Attacks in Federated Learning.}
    \label{classif-FL-security-privacy-attacks}
\end{figure}

Numerous studies have classified the existing attacks and vulnerabilities in the FL environment from various perspectives \cite{rodriguez2023survey}, from general targeted and untargeted attacks \cite{wahab2021federated} In our work, we account for the principal components of the FL pipeline as probable attack targets. We have thus organized the security and privacy attack surfaces into three primary classes (Client as a Target, Server as a Target, and Aggregation Process as a Target), with each class further divided into several subclasses. Fig. \ref{classif-FL-security-privacy-attacks} illustrates this categorization. We believe this hierarchical classification facilitates a deep comprehension and expedites the knowledge identification for researchers interested in securing FL systems.

\subsubsection{\textbf{Client as a Target}}
In decentralized learning, where the server is not aware of the reliability of the orchestrated clients, setting defensive countermeasures in case of client deviation becomes challenging. This limitation is known as the \textit{client unreliability} problem, referring to the inability of the server to detect whether a client's behavior is normal or deviated from the expected protocol \cite{ma2021federated}. This vulnerability opens the door for potential attacks exploiting various elements on the client side. To systematically categorize the threats in this class, we have classified them as follows:

\begin{itemize}
    \item \textit{Data-based Attacks.}
    During data preprocessing, cleaning operations offer adversaries potential openings \cite{ma2021federated}. For instance, the \textit{data poisoning attack} is a common threat to the integrity of the training dataset in FL \cite{hallaji2022federated}. An attacker with access to the local data can harm the accuracy of the learned model by tampering with or adding malicious data to the training set, resulting in a biased or impaired global model. Two common types of data poisoning attacks are clean-label and dirty-label attacks \cite{qammar2022federated}. Clean-label attacks \cite{shafahi2018poison} manipulate local input data without altering the labels. Conversely, dirty-label attacks \cite{lim2020federated} modify data sample labels, often using label-flipping or toxic sample generation, making them difficult to mitigate as the adversary can change labels without altering data features. For example, in a handwritten digit prediction, a label-flipping attack could involve flipping all the digits 3s into 5s and vice versa, causing the model to predict incorrectly, as illustrated in Fig. \ref{label_flipping_attack}.

\begin{figure}[htbp]
    \centering
    \includegraphics[width=0.4\textwidth]{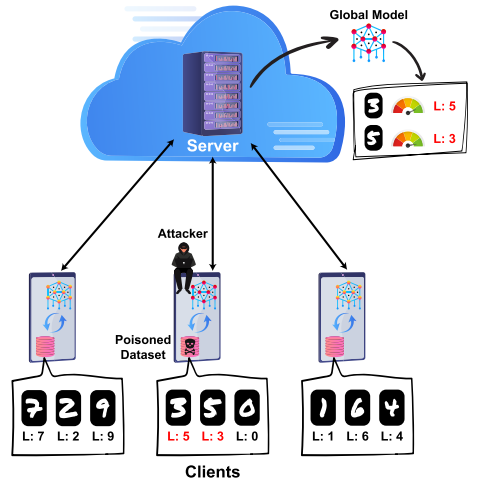}
    \caption{An example of data poisoning attack through label-flipping \cite{bouacida2021vulnerabilities}.}
    \label{label_flipping_attack}
\end{figure}
    \textit{Backdoors} are also data-based attacks \cite{gu2017badnets}, aiming to deteriorate the performance of a specific subtask while preserving the performance of the overall model task. By altering data characteristics, an adversary can induce the model to respond according to its intentions when the input contains the backdoor features \cite{zhang2023survey}. Furthermore, even without direct exchange of raw data, transferring model updates does not ensure complete privacy protection \cite{yin2021comprehensive}. Hence, there remains a risk of \textit{information leakage}, as demonstrated in prior studies \cite{hitaj2017deep, melis2019exploiting}.

    \item \textit{Local Model-based Attacks.}
     During training, \textit{model poisoning} attacks, often more efficient than data poisoning attacks, target the integrity of the FL process. Malicious parties, whether participating clients or external adversaries, can alter local updates before submitting them to the server. In the FL environment, where ensuring the trustworthiness of all active devices is not always feasible, preventing such attacks becomes challenging \cite{wang2022defense}.

    \item \textit{Training-based Attacks.}
    The \textit{manipulation of training rules} is a technique for compromising the computation and the global model's availability in federated learning. If an attacker gains control over one client, they could adjust the training hyperparameters, such as learning rate, local epochs, and batch size, to delay the convergence rate or even halt the global model's learning process \cite{bouacida2021vulnerabilities}. 
    A different approach for disrupting the training is by using \textit{Generative Adversarial Networks} (GANs). A GAN-trained model enables the attacker to reconstruct and control benign-like users' data, resulting in sensitive information disclosure and compromising the global model by injecting poisoned updates \cite{hitaj2017deep, zhang2019poisoning}.  
\end{itemize} 

\subsubsection{\textbf{Server as a Target}}

In federated learning, the server plays a crucial role by holding essential information about the model's architecture and receiving the local user's weight values in each round. Therefore, it becomes an attractive target for attacks. 
This section categorizes server-side attacks into three main categories as follows:
\begin{figure}[htbp]
    \centering
    \includegraphics[width=0.5\textwidth]{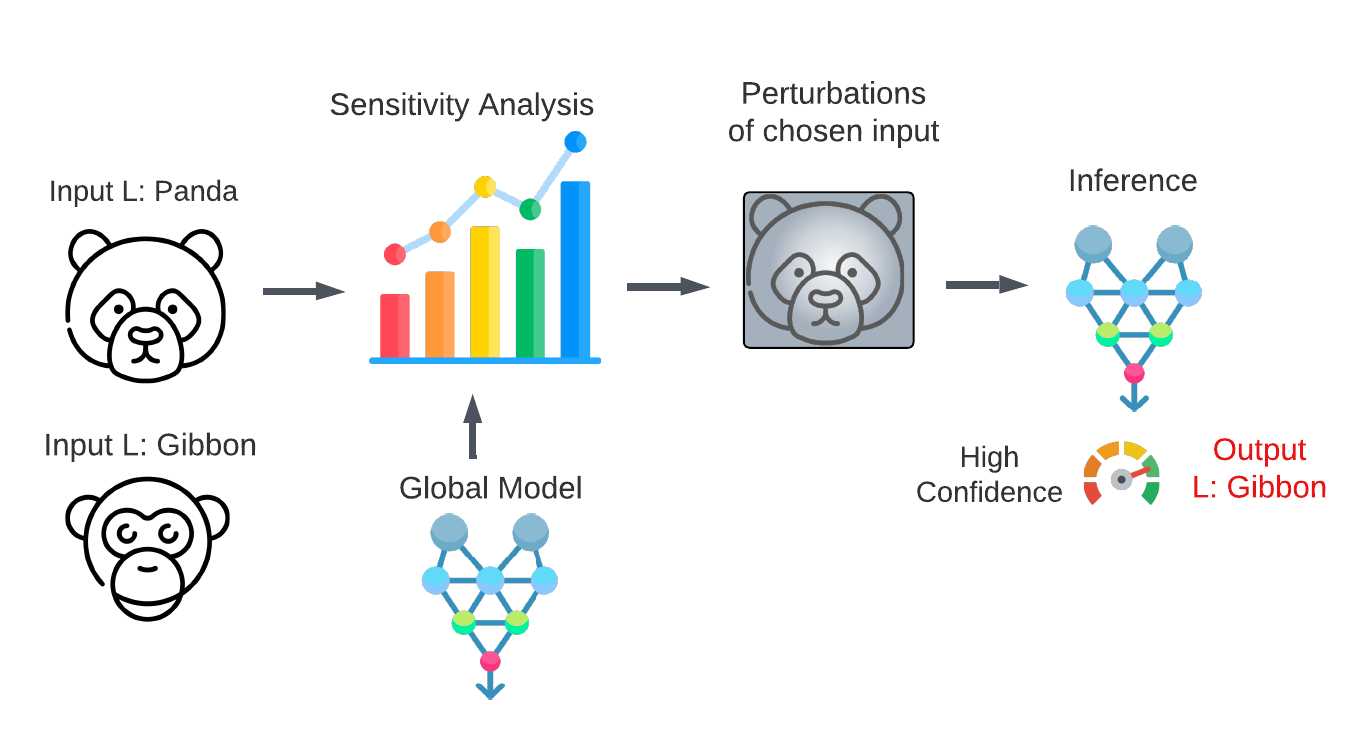}
    \caption{An example of adversarial samples crafting process for evasion attacks.}
    \label{evasion_attack}
\end{figure}

\begin{itemize}
    \item \textit{Falsification.}
    During the inference phase, \textit{evasion attacks} aim to fool the target model by introducing subtle variations to the input data \cite{mothukuri2021survey}.
    One widely-used technique for evading ML models involves adversarial samples, wherein an attacker alters test samples to closely resemble the original data. Yet, they are carefully determined according to the model's sensitivity to yield a class change \cite{wang2023potent}. For example, an attacker might add a small amount of noise to an image of a panda, causing the model to identify it as a gibbon mistakenly \cite{liu2022threats}. See Fig. \ref{evasion_attack}. 
    Additionally, we bring into this category the problem of \textit{verifiability} \cite{bouacida2021vulnerabilities}, which concerns the user's and server's ability to ascertain that all participants are working honestly without implicitly conducting any curious behavior that may reveal private information or impair the federation utility.
    
    \item \textit{Performance Degradation.}
     In traditional FL, a malicious or compromised server can steal private information or easily manipulate the global model to exploit the shared computation in building malicious tasks. Besides, the QoS in FL applications mainly relies on the server robustness to prevent adversary intrusions. Specifically, an unstable server exposes the entire system to the vulnerability of the \textit{Single Point of Failure attack} \cite{qammar2022securing}. In other scenarios, a compromised participant may have an alternative goal beyond sabotaging the model. Instead, they may seek to disrupt the process by submitting fake updates until it crushes. Since the server is typically unaware of such conduct, this operation, commonly known as a \textit{Distributed Denial of Service} (DDoS) attack, will persist and paralyze the entire federated learning system \cite{qammar2022securing}. \textit{Clients' unavailability} is another issue that can slow down the convergence of the global model \cite{mothukuri2021survey}. As outlined in Section \ref{heteroheneity_in_FL}, clients may dropout of the training due to limited device resources or network connectivity issues, leaving the server in an unproductive waiting state. 

     \item \textit{Privacy Data Leaks.}
     The privacy leakage issue has attracted significant attention \cite{geiping2020inverting, yin2021comprehensive}. Although the row data never leaves the client's device, exchanging the gradients between the active devices and the server engenders a serious privacy leakage. Moreover, after publishing the final model, inner adversaries or outsiders may attempt to infer sensitive information regarding a subject or a dataset. 
     \textit{Inference attacks} \cite{nasr2019comprehensive} are a common threat to privacy during training and model deployment. Two common types of inference attacks are membership inference and properties inference. A membership inference \cite{hu2022membership} intends to determine whether a particular sample belongs to the training dataset. For instance, an attacker might try to identify if a patient's medical record was included in a disease prediction model, this could reveal that the patient has that condition. A property inference \cite{melis2019exploiting}, on the other hand, attempts to infer characteristics of the training data. In particular, these properties might be unrelated to the primary learning task. For example, in an \textit{age prediction} model, this could involve inferring if \textit{glasses-wearers} tend to be younger or older.

\end{itemize}

\subsubsection{\textbf{Aggregation Process as a Target}}
The aggregation process is a vital component of distributed learning, where convergence over this environment occurs after hundreds to thousands of communication rounds. However, an insecure communication channel increases vulnerability to privacy thefts and security menaces, particularly when subjected to external attacks. We discuss and categorize, in this section, prevalent attacks targeting the aggregation process.

\begin{itemize}
    \item \textit{Unreliable Communication.}
    Many types of potential attacks might come from external actors. For instance, \textit{Eavesdroppers} can probe the intermediate training updates (e.g., weights or gradients) or the final model (e.g., weights or the query results provided by a published API) by intercepting the communication between FL actors \cite{wang2019eavesdrop}. By doing so, they can either gain access to confidential information or substitute the original updates with crafted ones \cite{yin2021comprehensive}. 
    In addition, if an outsider executes a \textit{man-in-the-middle attack}, the updates may be stolen, adjusted, or deviated from their desired destination. 

    \item \textit{Robustness.}
    From the training phase standpoint, the \textit{free-rider attack} \cite{lin2019free, fraboni2021free} poses a distinct threat to the FL aggregation. In this attack, a participant mimics benign client behavior to acquire the global model without contributing to learning. The attacker stays passive, submitting meaningless updates or not updating at all, conserving local resources and avoiding data sharing while benefiting from the improved model and shared computing power \cite{qammar2022federated}. Hence, various risks outlined above render a \textit{non-robust aggregation} algorithm ineffective in protecting FL systems from potential harm. Consequently, rather than bolstering participants with a robust model, the FL approach could yield an impaired model, leading to erroneous decision-making and severe privacy breaches. Therefore, a robust aggregation method is paramount for upholding the FL integrity \cite{10062923}.

\end{itemize}

\section{Federated Learning Recent Techniques and Aggregation Strategies}
\label{fl_recent_techniques_and_aggregation}

\subsection{\textbf{Strategies for Heterogeneity and Personalization Concerns}}
\label{FL_agg_solution_hete}
\begin{figure}[htbp]
    \centering
    \includegraphics[scale=0.4]{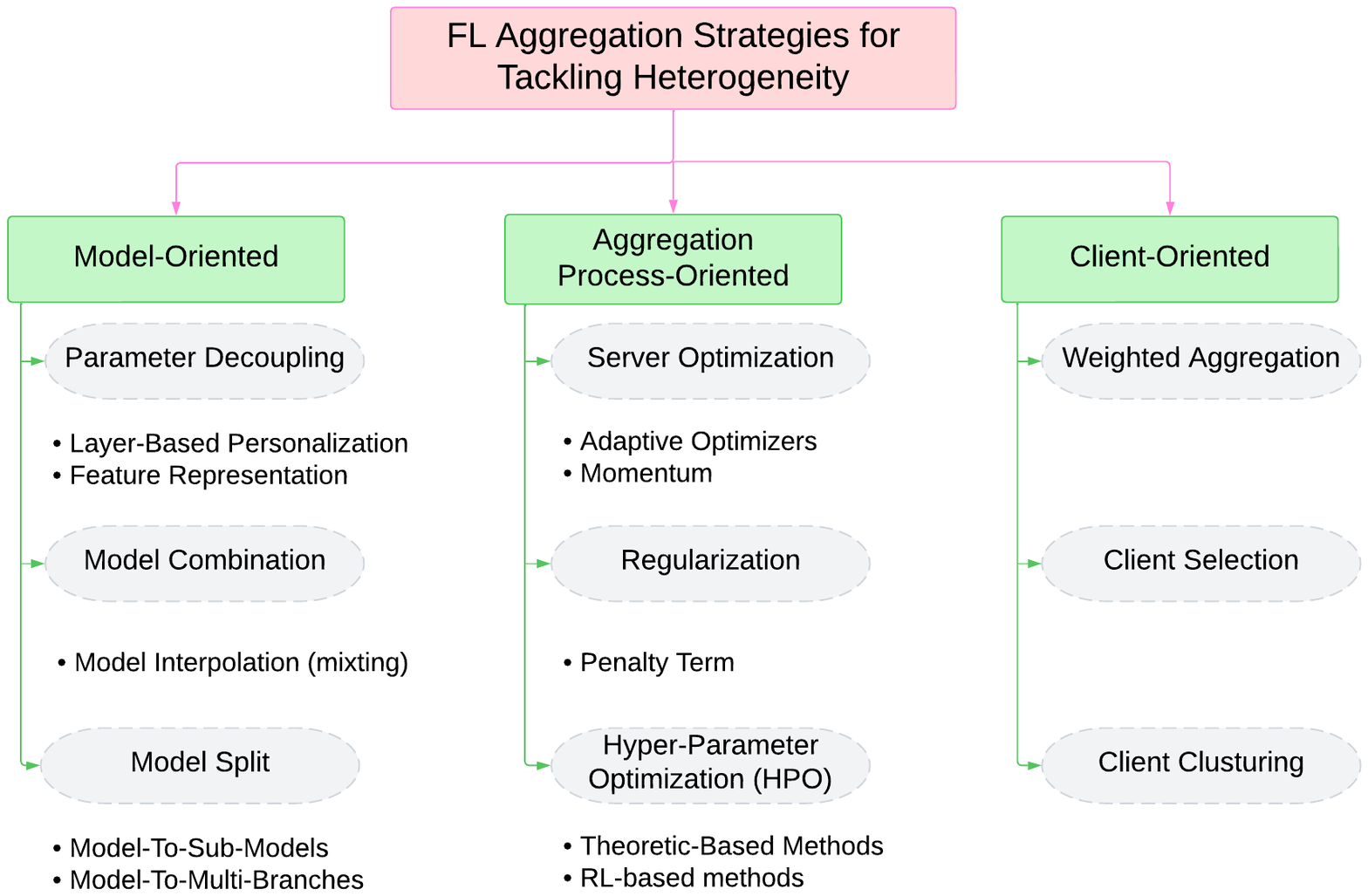}
    \caption{Classification of Federated Learning Aggregation Strategies for Tackling Heterogeneity.}
    \label{FL_strategies_heterogeneity}
\end{figure}

As broadly discussed earlier, the issue of heterogeneity is often a stumbling block when attempting to apply successful federated learning in real-world scenarios. This section explores the crucial role of implementing an aggregation method that addresses different types of heterogeneity while ensuring the FL model learns informative patterns from all available resources.
We classified the recent strategies in the literature into three distinct classes, as depicted in Fig. \ref{FL_strategies_heterogeneity}.

    \subsubsection{Model-Oriented} 
    Model-oriented strategies focus on improving personalization through the manipulation of the global and local model architecture. An important decision in this regard is whether to limit the collaborative training to only the upper/lower layers of the model, enable users to modify a shared global model, personalize the entire model locally, or train different parts of the final model separately on each device. Please refer to Table \ref{tab-summary-FL-aggreg-hete-model} for a thoughtfully curated collection of scholarly publications that have made significant contributions to this area of research.

\begin{itemize}
        \item \textit{Parameter Decoupling}

        This paradigm aims to achieve personalized models and assuage the heterogeneity impact. It involves dividing the model parameters into two or more sets and optimizing each set separately. 
        One configuration of parameter decoupling entails a \textit{layer-based split} of the neural network model \cite{arivazhagan2019federated}, wherein the layers are segmented into two sets: base layers and personalized layers. This structure helps ensure the privacy of the personalized layers and results in a high degree of personalization.
        
        For example, EPSL \cite{lin2023efficient} and ModFL \cite{liang2022modular} target personalization in resource-constrained environments. Alternatively,  propose an optimal layer selection strategy for energy, time, and privacy trade-offs. Leveraging pre-trained BERT encoders for NLP, FedSplitBERT \cite{lit2022federated} tackles both heterogeneity and communication challenges.
        \textit{Feature representation} is another decoupling approach. For instance, the authors in \cite{bui2019federated} have proposed the FURL method, which enables existing personalization techniques within FL by splitting model parameters. User-specific features remain private, while shared features are learned collaboratively. Another study in \cite{rakotomamonjy2023personalisedx} has addressed the issue of heterogeneous raw data representation among FL clients. Their framework called FLIC employs local embedding functions to map the data into a common space. Similar approaches utilize feature anchor vectors \cite{zhou2022fedfa}, low-dimensional classifier \cite{collins2021exploiting}, and more \cite{sun2021partialfed}.
        In essence, according to \cite{li2021aggregate}, parameter decoupling for privacy and personalization can be categorized into single-branch and multi-branch approaches. The former directly privatizes specific layers while aggregating the remaining ones via the server. The latter keeps the entire model shared but privatizes certain components.
        
        \item \textit{Global-Local Models Combination}
        
        A slightly different approach involves \textit{combining global and local models} \cite{criado2022non, hanzely2020federated}. This technique is used to personalize the final model employed by each client instead of adopting one globally deployed model for all clients. Unlike standard FL, each client has two models: a global model trained collaboratively and a private local model for fine-tuning the FL outcome. This benefits scenarios with low correlation between local and global data distribution. The authors in \cite{deng2020adaptive} advocate an APFL method that combines global and local models using an adaptively learned weight for improved personalization and generalization. Similarly, the framework in \cite{zec2020specialized} leverages federated averaging and mixtures of experts to achieve personalized models via \textit{model interpolation}.
        
        \item \textit{Model Split}
        
        To improve communication efficiency and tackle device heterogeneity, model split is a promising technique. This approach divides the model, often a neural network, into sub-models or branches. Each device then trains a specific portion, reducing the communication burden and computational cost per device.
        Within this particular branch, the researchers in \cite{dun2023efficient} offered AsyncDrop, as a new asynchronous solution to handle device heterogeneity in large-scale FL. This approach leverages dropout regularization, randomly masking a subset of neurons in each layer during training. This effectively creates sub-models where all layers are present but only a portion of neurons are active. Devices are then assigned sub-models for training based on their computational capabilities. In \cite{cui2022fedbranch}, the suggested FedBranch framwork adopts the strategy of model-splitting into a multi-branch neural network. Moreover, FedBranch employs a layer-wise aggregation to combine branch outputs and integrates a task offloading algorithm for efficient distribution of training tasks across branches. Building on the same foundation laid by \cite{dun2023efficient}, FedBranch approach assigns a suitable branch model to each participating client based on their computational resources. Finally, In \cite{mori2022personalized}, the authors have investigated another personalized approach using multi-branch architecture to establish pFedMB, enabling similar clients to automatically share knowledge without directly calculating the similarities, as with FedAMP \cite{huang2021personalized} and FedFomo \cite{zhang2020personalized}.

    \end{itemize}

\addtocounter{table}{-1}

\setlength\rotFPtop{0pt plus 1fil} 
\begin{table} [htbp]
\tiny
\caption{Summary of Federated Learning Aggregation Methods for Tackling Heterogeneity - Model-Oriented.}
\label{tab-summary-FL-aggreg-hete-model}
\begin{threeparttable}
\begin{tabularx}{\linewidth}{|X|XXX|XXXXXXXXXXX|}
\hline
\multicolumn{1}{|c|}{\multirow{3}{*}{\textbf{Related Work}}} &
  \multicolumn{3}{c|}{\multirow{2}{*}{\textbf{Environment}}} &
  \multicolumn{11}{c|}{\textbf{\textbf{Verified Goals}}} \\ \cline{5-15} 
 &
  \multicolumn{3}{c|}{} &
  \multicolumn{3}{c|}{\textbf{Process}} &
  \multicolumn{5}{c|}{\textbf{Model}} &
  \multicolumn{3}{c|}{\textbf{System}} \\ \cline{2-15} 
 &
  \multicolumn{1}{c|}{\textbf{Synch}} & 
  \multicolumn{1}{c|}{\textbf{Users}} & 
  \multicolumn{1}{c|}{\textbf{Archi}} & 
  
  \multicolumn{1}{c|}{\textbf{Cm. E.}} & 
  \multicolumn{1}{c|}{\textbf{Cp. E.}} & 
  \multicolumn{1}{c|}{\textbf{Conv}} & 
  \multicolumn{1}{c|}{\textbf{Pers}} & 
  \multicolumn{1}{c|}{\textbf{Gen}} & 
  \multicolumn{1}{c|}{\textbf{Reg}} & 
  \multicolumn{1}{c|}{\textbf{Fair}} & 
  \multicolumn{1}{c|}{\textbf{Heter}} & 
  \multicolumn{1}{c|}{\textbf{Sec}} & 
  \multicolumn{1}{c|}{\textbf{Priv}} & 
  \multicolumn{1}{c|}{\textbf{Scal}} \\  
  \hline

  \multicolumn{1}{|c|}{} &
  \multicolumn{14}{c|}{\textbf{Parameter Decoupling}}  \\ \hline
  
  \multicolumn{1}{|c|}{\textbf{Layer-based}} &
  \multicolumn{14}{c|}{}  \\ \hline

  \multicolumn{1}{|c|}{ EPSL \cite{lin2023efficient}} &
  \multicolumn{1}{c|}{Synch} &
  \multicolumn{1}{c|}{20} &
  \multicolumn{1}{c|}{Cent} &
  \multicolumn{1}{c|}{ \checkmark} &
  \multicolumn{1}{c|}{ \checkmark} &
  \multicolumn{1}{c|}{ \checkmark} &
  \multicolumn{1}{c|}{ \checkmark} &
  \multicolumn{1}{c|}{--} &
  \multicolumn{1}{c|}{--} &
  \multicolumn{1}{c|}{--} &
  \multicolumn{1}{c|}{\checkmark} &
  \multicolumn{1}{c|}{--} &
  \multicolumn{1}{c|}{--} &
  \multicolumn{1}{c|}{--}  \\ 
  \hline

  \multicolumn{1}{|c|}{FedSplitBERT \cite{lit2022federated}} &
  \multicolumn{1}{c|}{Synch} &
  \multicolumn{1}{c|}{10} &
  \multicolumn{1}{c|}{Cent} &
  \multicolumn{1}{c|}{\checkmark} &
  \multicolumn{1}{c|}{--} &
  \multicolumn{1}{c|}{--} &
  \multicolumn{1}{c|}{\checkmark} &
  \multicolumn{1}{c|}{--} &
  \multicolumn{1}{c|}{--} &
  \multicolumn{1}{c|}{--} &
  \multicolumn{1}{c|}{\checkmark} &
  \multicolumn{1}{c|}{--} &
  \multicolumn{1}{c|}{--} &
  \multicolumn{1}{c|}{--} \\ 
  \hline

 \multicolumn{1}{|c|}{\textbf{Feature Rep}} &
  \multicolumn{14}{c|}{}  \\ 
  \hline

  
  \multicolumn{1}{|c|}{FLIC \cite{rakotomamonjy2023personalisedx}} & 
  \multicolumn{1}{c|}{Synch} &
  \multicolumn{1}{c|}{200} &
  \multicolumn{1}{c|}{Cent} &
  \multicolumn{1}{c|}{\checkmark} &
  \multicolumn{1}{c|}{--} &
  \multicolumn{1}{c|}{\checkmark} &
  \multicolumn{1}{c|}{\checkmark} &
  \multicolumn{1}{c|}{--} &
  \multicolumn{1}{c|}{\checkmark} &
  \multicolumn{1}{c|}{--} &
  \multicolumn{1}{c|}{\checkmark} &
  \multicolumn{1}{c|}{--} &
  \multicolumn{1}{c|}{--} &
  \multicolumn{1}{c|}{\checkmark} \\ 
  \hline

  \multicolumn{1}{|c|}{FedFA \cite{zhou2022fedfa}} &
  \multicolumn{1}{c|}{Synch} &
  \multicolumn{1}{c|}{10} &
  \multicolumn{1}{c|}{Cent} &
  \multicolumn{1}{c|}{--} &
  \multicolumn{1}{c|}{--} &
  \multicolumn{1}{c|}{--} &
  \multicolumn{1}{c|}{\checkmark} &
  \multicolumn{1}{c|}{--} &
  \multicolumn{1}{c|}{\checkmark} &
  \multicolumn{1}{c|}{--} &
  \multicolumn{1}{c|}{\checkmark} &
  \multicolumn{1}{c|}{--} &
  \multicolumn{1}{c|}{--} &
  \multicolumn{1}{c|}{--} \\ 
  \hline

  \multicolumn{1}{|c|}{FedRep \cite{collins2021exploiting}} &
  \multicolumn{1}{c|}{Synch} &
  \multicolumn{1}{c|}{1000} &
  \multicolumn{1}{c|}{Cent} &
  \multicolumn{1}{c|}{--} &
  \multicolumn{1}{c|}{--} &
  \multicolumn{1}{c|}{\checkmark} &
  \multicolumn{1}{c|}{\checkmark} &
  \multicolumn{1}{c|}{--} &
  \multicolumn{1}{c|}{--} &
  \multicolumn{1}{c|}{--} &
  \multicolumn{1}{c|}{\checkmark} &
  \multicolumn{1}{c|}{--} &
  \multicolumn{1}{c|}{--} &
  \multicolumn{1}{c|}{\checkmark} \\ 
  \hline

  \multicolumn{1}{|c|}{} &
  \multicolumn{14}{c|}{\textbf{Model Combination}}  \\ 
  \hline
  \multicolumn{1}{|c|}{\textbf{Model Mixting}} &
  \multicolumn{14}{c|}{}  \\ 
  \hline
   
  \multicolumn{1}{|c|}{APFL \cite{deng2020adaptive}} & 
  \multicolumn{1}{c|}{Synch} &
  \multicolumn{1}{c|}{100} &
  \multicolumn{1}{c|}{Cent} &
  \multicolumn{1}{c|}{\checkmark} &
  \multicolumn{1}{c|}{--} &
  \multicolumn{1}{c|}{\checkmark} &
  \multicolumn{1}{c|}{\checkmark} &
  \multicolumn{1}{c|}{\checkmark} &
  \multicolumn{1}{c|}{--} &
  \multicolumn{1}{c|}{--} &
  \multicolumn{1}{c|}{\checkmark} &
  \multicolumn{1}{c|}{--} &
  \multicolumn{1}{c|}{--} &
  \multicolumn{1}{c|}{--} \\ 
  \hline

  \multicolumn{1}{|c|}{ \cite{zec2020specialized}} &
  \multicolumn{1}{c|}{Synch} &
  \multicolumn{1}{c|}{20} &
  \multicolumn{1}{c|}{Cent} &
  \multicolumn{1}{c|}{--} &
  \multicolumn{1}{c|}{--} &
  \multicolumn{1}{c|}{--} &
  \multicolumn{1}{c|}{\checkmark} &
  \multicolumn{1}{c|}{--} &
  \multicolumn{1}{c|}{--} &
  \multicolumn{1}{c|}{--} &
  \multicolumn{1}{c|}{\checkmark} &
  \multicolumn{1}{c|}{--} &
  \multicolumn{1}{c|}{\checkmark} &
  \multicolumn{1}{c|}{--} \\ 
  \hline

  \multicolumn{1}{|c|}{ \cite{hanzely2020federated}} &  
  \multicolumn{1}{c|}{Synch} &
  \multicolumn{1}{c|}{100} &
  \multicolumn{1}{c|}{Cent} &
  \multicolumn{1}{c|}{\checkmark} &
  \multicolumn{1}{c|}{--} &
  \multicolumn{1}{c|}{\checkmark} &
  \multicolumn{1}{c|}{\checkmark} &
  \multicolumn{1}{c|}{--} &
  \multicolumn{1}{c|}{\checkmark} &
  \multicolumn{1}{c|}{--} &
  \multicolumn{1}{c|}{\checkmark} &
  \multicolumn{1}{c|}{--} &
  \multicolumn{1}{c|}{--} &
  \multicolumn{1}{c|}{--} \\ 
  \hline


  \multicolumn{1}{|c|}{} &
  \multicolumn{14}{c|}{\textbf{Model Split}}  \\ 
  \hline

  \multicolumn{1}{|c|}{\textbf{Multi-Brunches}} &
  \multicolumn{14}{c|}{}  \\ 
 \hline
  
  \multicolumn{1}{|c|}{FedBranch \cite{cui2022fedbranch}} &   
  \multicolumn{1}{c|}{Synch} &
  \multicolumn{1}{c|}{100} &
  \multicolumn{1}{c|}{Cent} &
  \multicolumn{1}{c|}{\checkmark} &
  \multicolumn{1}{c|}{--} &
  \multicolumn{1}{c|}{--} &
  \multicolumn{1}{c|}{--} &
  \multicolumn{1}{c|}{--} &
  \multicolumn{1}{c|}{\checkmark} &
  \multicolumn{1}{c|}{--} &
  \multicolumn{1}{c|}{\checkmark} &
  \multicolumn{1}{c|}{--} &
  \multicolumn{1}{c|}{--} &
  \multicolumn{1}{c|}{--} \\ 
  \hline

  \multicolumn{1}{|c|}{pFedMB \cite{mori2022personalized}} &
  \multicolumn{1}{c|}{Synch} &
  \multicolumn{1}{c|}{50} &
  \multicolumn{1}{c|}{Cent} &
  \multicolumn{1}{c|}{--} &
  \multicolumn{1}{c|}{--} &
  \multicolumn{1}{c|}{--} &
  \multicolumn{1}{c|}{\checkmark} &
  \multicolumn{1}{c|}{--} &
  \multicolumn{1}{c|}{--} &
  \multicolumn{1}{c|}{--} &
  \multicolumn{1}{c|}{\checkmark} &
  \multicolumn{1}{c|}{--} &
  \multicolumn{1}{c|}{--} &
  \multicolumn{1}{c|}{--}\\ 
  \hline

  \multicolumn{1}{|c|}{\textbf{Sub-models}} &
  \multicolumn{14}{c|}{}  \\ 
  \hline
  
  \multicolumn{1}{|c|}{AsyncDrop \cite{dun2023efficient}} & 
  \multicolumn{1}{c|}{Asynch} &
  \multicolumn{1}{c|}{100} &
  \multicolumn{1}{c|}{Cent} &
  \multicolumn{1}{c|}{--} &
  \multicolumn{1}{c|}{\checkmark} &
  \multicolumn{1}{c|}{\checkmark} &
  \multicolumn{1}{c|}{\checkmark} &
  \multicolumn{1}{c|}{--} &
  \multicolumn{1}{c|}{--} &
  \multicolumn{1}{c|}{--} &
  \multicolumn{1}{c|}{\checkmark} &
  \multicolumn{1}{c|}{--} &
  \multicolumn{1}{c|}{--} &
  \multicolumn{1}{c|}{--} \\ 
  \hline

\end{tabularx}
\begin{tablenotes}
      \footnotesize \item \textit{Synch}: Synchronization mode \{Synchronous, Asynchronous\}. \textit{Users}: Maximum active users used in experiments. \textit{Archi}: Architecture  \{Centralized, Decentralized(P2P), Hierarchical\}. \textit{Cm. E.}: Communication Efficiency. \textit{Cp. E.}: Computation Efficiency.\textit{ Conv}: Convergence Analysis. \textit{Pers}: Personalization. \textit{Gen}: Generalization. \textit{Fair}: Fairness. \textit{Heter}: Heterogeneity. \textit{Sec}: Security. \textit{Priv}: Privacy. \textit{Scal}: Scalability.
\end{tablenotes}
\end{threeparttable}

\end{table}

    \subsubsection{Aggregation Process-Oriented.}
    Aggregation-oriented strategies involve optimizing various aspects of the aggregation process, including training hyperparameters, loss functions, gradient variances, convergence rates, and learning direction. The primary objective is to identify and implement the optimal aggregation environment that accelerates FL convergence while accommodating the unique characteristics of individual users. Table \ref{tab-summary-FL-aggreg-hete-agg} delineates the sub-categories and furnishes instances of literature papers that pertain to this particular research area.

\begin{figure}[htbp]
    \centering
    \includegraphics[width=0.5\textwidth]{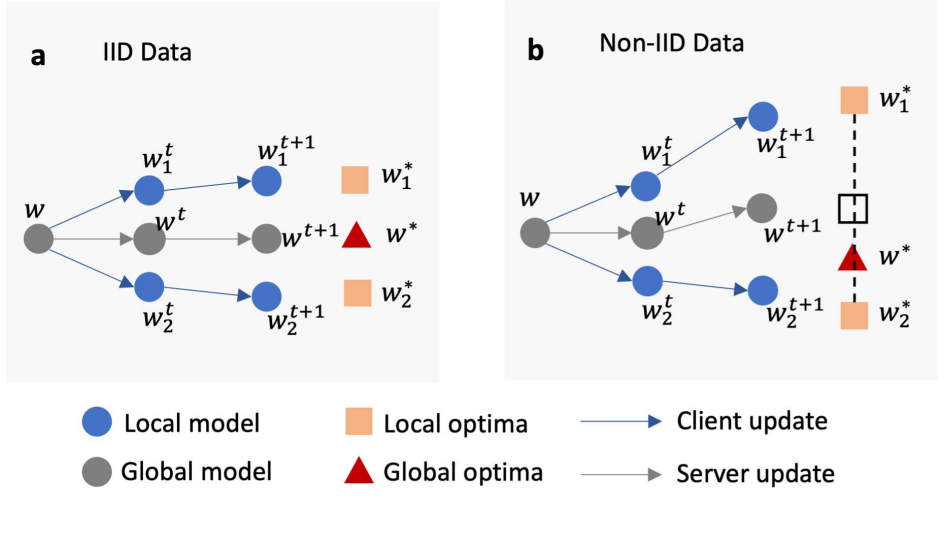}
    \caption{Illustration of Client Drift Phenomena in Federated Learning for Two Clients \cite{tan2022towards}.}
    \label{client-drift}
    \end{figure}

\begin{itemize}

    \item \textit{Server Optimization}

        Apart from the architectural decision choice, another solution is to modify the server-side optimizers. \textit{Adaptive gradient} methods are extensively employed in traditional ML. Unlike fixed learning rates in vanilla gradient descent, which rely on either a constant rate or a preset schedule, adaptive methods adjust the rate on the fly based on the gradient magnitude. This flexibility has shown theoretical and practical advantages in FL \cite{reddi2020adaptive}, unlocking superior generalization performance ober non-adaptive methods including Adagrad \cite{duchi2011adaptive}, Adam \cite{kingma2014adam}, and Yogi \cite{zaheer2018adaptive}.
        Meanwhile, the authors in \cite{li2023fedda} argue that server-side adaptive gradients alone do not fully leverage adaptive information. To address this, they propose FedDA a framework enabling each client to adjust its learning rate based on local gradients and past updates.
        Another server-side optimization technique is \textit{momentum}. In federated learning, SGD with momentum averages gradients from clients while considering past updates, guiding the global model toward the right learning direction. In other words, it bolsters faster convergence and escapes from poor local minima.
        For instance, the STEM algorithm \cite{khanduri2021stem} utilizes momentum for both client and server updates, achieving an optimal balance between updates' frequency and minibatch sizes. Similarly, FedGLOMO \cite{das2022faster} leverages global and local momentum terms to reduce variance and accelerate convergence. These approaches, along with FedMom and FedNAG \cite{huo2020faster, yang2022federated} (variants integrating Nesterov's Accelerated Gradient), showcase the effectiveness of momentum in heterogeneous FL scenarios.

        \item \textit{Regularization}

        Regularization, a technique to prevent overfitting in ML, can also combat client drift in FL settings. Client drift occurs when local models diverge from the global optimum due to heterogeneous data (Figure \ref{client-drift}). Regularization adds a penalty term to the loss function to penalize models that deviate too far, encouraging them to stay closer to the global model.
        Recent work incorporates dynamic regularization \cite{acar2021federated}, triplet term regularization \cite{li2023fedtrip}, but also integrates it with other techniques, such as knowledge distillation \cite{luo2023improving} and stratified sampling \cite{lu2023federated}.

        \item \textit{Hyper-Parameters Optimization (HPO)}

        FL \textit{hyperparameter tuning} prioritizes communication and computation efficiency over accuracy, in construct ML HPO focused on accuracy. This optimization accelerates convergence by adjusting factors like client selection, local training steps, and aggregation frequency – all crucial for balancing performance and efficiency \cite{hefel24iwcmc}.
        
        In fact, FL grapples with a distinct challenge: optimizing hyperparameters for a distributed system. Here, two main approaches emerge: \textit{theoretic-based} \cite{luo2021cost, wu2021fast, shi2020device} and \textit{Reinforcement Learning (RL)-based} methods \cite{guo2022auto, zhang2021deep, lu2020blockchain}. Theoretical methods offer efficient solutions by simplifying the problem with environmental assumptions, which may not hold true  in the face of data's dynamism and lack of clear patterns. This is where RL-based methods shine. They treat the tuning task as a dynamic decision-making process, offering greater adaptability. A prime example is Dap-FL \cite{chen2023dap}. It implements a Deep Deterministic Policy Gradient (DDPG) algorithm to adjust clients' learning rates and training epochs based on their progress and the global model. Auto-FedRL \cite{guo2022auto} pushes the HPO boundaries even further. It doesn't stop at a limited set of hyperparameters. Instead, it employs an online RL agent to dynamically adjust a larger spectrum of hyperparameters for both clients and the server.
    
    \end{itemize}

\addtocounter{table}{-1}
  
\setlength\rotFPtop{0pt plus 1fil} 

\begin{table}[htbp]
\tiny
\caption{Summary of Federated Learning Aggregation Methods for Tackling Heterogeneity - Aggregation Process-Oriented.}
\label{tab-summary-FL-aggreg-hete-agg}

\begin{threeparttable}
\begin{tabularx}{\linewidth}{|X|XXX|XXXXXXXXXXX|}
\hline
\multicolumn{1}{|c|}{\multirow{3}{*}{\textbf{Related Work}}} &
  \multicolumn{3}{c|}{\multirow{2}{*}{\textbf{Environment}}} &
  \multicolumn{11}{c|}{\textbf{\textbf{Verified Goals}}} \\ \cline{5-15} 
 &
  \multicolumn{3}{c|}{} &
  \multicolumn{3}{c|}{\textbf{Process}} &
  \multicolumn{5}{c|}{\textbf{Model}} &
  \multicolumn{3}{c|}{\textbf{System}} \\ \cline{2-15} 
 &
  \multicolumn{1}{c|}{\textbf{Synch}} & 
  \multicolumn{1}{c|}{\textbf{Users}} & 
  \multicolumn{1}{c|}{\textbf{Archi}} & 
  
  \multicolumn{1}{c|}{\textbf{Cm. E.}} & 
  \multicolumn{1}{c|}{\textbf{Cp. E.}} & 
  \multicolumn{1}{c|}{\textbf{Conv}} & 
  \multicolumn{1}{c|}{\textbf{Pers}} & 
  \multicolumn{1}{c|}{\textbf{Gen}} & 
  \multicolumn{1}{c|}{\textbf{Reg}} & 
  \multicolumn{1}{c|}{\textbf{Fair}} & 
  \multicolumn{1}{c|}{\textbf{Heter}} & 
  \multicolumn{1}{c|}{\textbf{Sec}} & 
  \multicolumn{1}{c|}{\textbf{Priv}} & 
  \multicolumn{1}{c|}{\textbf{Scal}} \\  
  \hline

  \multicolumn{1}{|c|}{ \textbf{Used Technique}} &
  \multicolumn{14}{c|}{\textbf{Server Optimization}} \\ \hline
  
  \multicolumn{1}{|c|}{\textbf{Adaptive Optimizers}} &
  \multicolumn{14}{c|}{} \\ \hline

  \multicolumn{1}{|c|}{FedDA \cite{li2023fedda}} &
  \multicolumn{1}{c|}{Synch} &
  \multicolumn{1}{c|}{10} &
  \multicolumn{1}{c|}{Cent} &
  \multicolumn{1}{c|}{\checkmark} &
  \multicolumn{1}{c|}{--} &
  \multicolumn{1}{c|}{\checkmark} &
  \multicolumn{1}{c|}{--} &
  \multicolumn{1}{c|}{--} &
  \multicolumn{1}{c|}{--} &
  \multicolumn{1}{c|}{--} &
  \multicolumn{1}{c|}{\checkmark} &
  \multicolumn{1}{c|}{--} &
  \multicolumn{1}{c|}{--} &
  \multicolumn{1}{c|}{--} \\ 
  \hline

  \multicolumn{1}{|c|}{ \cite{reddi2020adaptive}} &
  \multicolumn{1}{c|}{Synch} &
  \multicolumn{1}{c|}{300k} &
  \multicolumn{1}{c|}{Cent} &
  \multicolumn{1}{c|}{\checkmark} &
  \multicolumn{1}{c|}{--} &
  \multicolumn{1}{c|}{\checkmark} &
  \multicolumn{1}{c|}{--} &
  \multicolumn{1}{c|}{--} &
  \multicolumn{1}{c|}{--} &
  \multicolumn{1}{c|}{--} &
  \multicolumn{1}{c|}{\checkmark} &
  \multicolumn{1}{c|}{--} &
  \multicolumn{1}{c|}{--} &
  \multicolumn{1}{c|}{--} \\ 
  \hline

  \multicolumn{1}{|c|}{\textbf{Momentum}} &
  \multicolumn{14}{c|}{} \\ \hline
  
  \multicolumn{1}{|c|}{FedGLOMO \cite{das2022faster}} & 
  \multicolumn{1}{c|}{Synch} &
  \multicolumn{1}{c|}{50} &
  \multicolumn{1}{c|}{Cent} &
  \multicolumn{1}{c|}{\checkmark} &
  \multicolumn{1}{c|}{--} &
  \multicolumn{1}{c|}{\checkmark} &
  \multicolumn{1}{c|}{--} &
  \multicolumn{1}{c|}{--} &
  \multicolumn{1}{c|}{\checkmark} &
  \multicolumn{1}{c|}{--} &
  \multicolumn{1}{c|}{\checkmark} &
  \multicolumn{1}{c|}{--} &
  \multicolumn{1}{c|}{--} &
  \multicolumn{1}{c|}{--} \\ 
  \hline

  \multicolumn{1}{|c|}{MIME \cite{karimireddy2020mime}} &
  \multicolumn{1}{c|}{Synch} &
  \multicolumn{1}{c|}{20} &
  \multicolumn{1}{c|}{Cent} & 
  \multicolumn{1}{c|}{\checkmark} &
  \multicolumn{1}{c|}{--} &
  \multicolumn{1}{c|}{\checkmark} &
  \multicolumn{1}{c|}{--} &
  \multicolumn{1}{c|}{--} &
  \multicolumn{1}{c|}{\checkmark} &
  \multicolumn{1}{c|}{--} &
  \multicolumn{1}{c|}{\checkmark} &
  \multicolumn{1}{c|}{--} &
  \multicolumn{1}{c|}{--} &
  \multicolumn{1}{c|}{--} \\ 
  \hline

  \multicolumn{1}{|c|}{\textbf{}} &
  \multicolumn{14}{c|}{\textbf{Regularization}} \\ \hline

   \multicolumn{1}{|c|}{\textbf{Penalty Term}} &
  \multicolumn{14}{c|}{} \\ \hline
  
  \multicolumn{1}{|c|}{FedDyn \cite{acar2021federated}} &
  \multicolumn{1}{c|}{Synch} &
  \multicolumn{1}{c|}{100} &
  \multicolumn{1}{c|}{Cent} &
  \multicolumn{1}{c|}{\checkmark} &
  \multicolumn{1}{c|}{\checkmark} & 
  \multicolumn{1}{c|}{\checkmark} &
  \multicolumn{1}{c|}{--} &
  \multicolumn{1}{c|}{--} &
  \multicolumn{1}{c|}{\checkmark} &
  \multicolumn{1}{c|}{--} &
  \multicolumn{1}{c|}{\checkmark} &
  \multicolumn{1}{c|}{--} &
  \multicolumn{1}{c|}{--} &
  \multicolumn{1}{c|}{\checkmark} \\ 
  \hline

  \multicolumn{1}{|c|}{FedTrip \cite{li2023fedtrip}} & 
  \multicolumn{1}{c|}{Synch} &
  \multicolumn{1}{c|}{4} &
  \multicolumn{1}{c|}{Hier} &
  \multicolumn{1}{c|}{\checkmark} &
  \multicolumn{1}{c|}{\checkmark} & 
  \multicolumn{1}{c|}{\checkmark} &
  \multicolumn{1}{c|}{--} &
  \multicolumn{1}{c|}{--} &
  \multicolumn{1}{c|}{\checkmark} &
  \multicolumn{1}{c|}{--} &
  \multicolumn{1}{c|}{\checkmark} &
  \multicolumn{1}{c|}{--} &
  \multicolumn{1}{c|}{--} &
  \multicolumn{1}{c|}{--} \\ 
  \hline

  \multicolumn{1}{|c|}{} &
  \multicolumn{14}{c|}{\textbf{\textbf{Hyper-Parameter Optimization HPO (fine-tuning)}}} \\ \hline

 \multicolumn{1}{|c|}{\textbf{Theoretic-based}} &
  \multicolumn{14}{c|}{} \\ \hline
  
  \multicolumn{1}{|c|}{ \cite{luo2021cost}} & 
  \multicolumn{1}{c|}{Synch} &
  \multicolumn{1}{c|}{20} &
  \multicolumn{1}{c|}{Cent} &
  \multicolumn{1}{c|}{\checkmark} &
  \multicolumn{1}{c|}{\checkmark} &
  \multicolumn{1}{c|}{\checkmark} &
  \multicolumn{1}{c|}{--} &
  \multicolumn{1}{c|}{--} &
  \multicolumn{1}{c|}{--} &
  \multicolumn{1}{c|}{--} &
  \multicolumn{1}{c|}{\checkmark} &
  \multicolumn{1}{c|}{--} &
  \multicolumn{1}{c|}{--} &
  \multicolumn{1}{c|}{--} \\ 
  \hline

  \multicolumn{1}{|c|}{Flora \cite{zhou2021flora}} & 
  \multicolumn{1}{c|}{-} & 
  \multicolumn{1}{c|}{10} &
  \multicolumn{1}{c|}{Cent} &
  \multicolumn{1}{c|}{\checkmark} &
  \multicolumn{1}{c|}{\checkmark} &
  \multicolumn{1}{c|}{--} &
  \multicolumn{1}{c|}{--} &
  \multicolumn{1}{c|}{--} &
  \multicolumn{1}{c|}{--} &
  \multicolumn{1}{c|}{--} &
  \multicolumn{1}{c|}{\checkmark} &
  \multicolumn{1}{c|}{--} &
  \multicolumn{1}{c|}{--} &
  \multicolumn{1}{c|}{--} \\ 
  \hline
  
  \multicolumn{1}{|c|}{\textbf{RL-based}} &
  \multicolumn{14}{c|}{} \\ \hline
  
  \multicolumn{1}{|c|}{Auto-FedRL\cite{guo2022auto}} &
  \multicolumn{1}{c|}{Synch} &
  \multicolumn{1}{c|}{3} &
  \multicolumn{1}{c|}{Cent} &
  \multicolumn{1}{c|}{--} &
  \multicolumn{1}{c|}{\checkmark} &
  \multicolumn{1}{c|}{\checkmark} &
  \multicolumn{1}{c|}{--} &
  \multicolumn{1}{c|}{--} &
  \multicolumn{1}{c|}{--} &
  \multicolumn{1}{c|}{--} &
  \multicolumn{1}{c|}{\checkmark} &
  \multicolumn{1}{c|}{--} &
  \multicolumn{1}{c|}{--} &
  \multicolumn{1}{c|}{--} \\ 
  \hline
  
  \multicolumn{1}{|c|}{Dap-FL \cite{chen2023dap}} & 
  \multicolumn{1}{c|}{Synch} &
  \multicolumn{1}{c|}{33} &
  \multicolumn{1}{c|}{Cent} &
  \multicolumn{1}{c|}{\checkmark} &
  \multicolumn{1}{c|}{\checkmark} &
  \multicolumn{1}{c|}{\checkmark} &
  \multicolumn{1}{c|}{--} &
  \multicolumn{1}{c|}{--} &
  \multicolumn{1}{c|}{--} &
  \multicolumn{1}{c|}{--} &
  \multicolumn{1}{c|}{\checkmark} &
  \multicolumn{1}{c|}{\checkmark} &
  \multicolumn{1}{c|}{\checkmark} &
  \multicolumn{1}{c|}{--} \\ 
  \hline

\end{tabularx}

\begin{tablenotes}
      \footnotesize \item \textit{Synch}: Synchronization mode \{Synchronous, Asynchronous\}. \textit{Users}: Maximum active users used in experiments. \textit{Archi}: Architecture  \{Centralized, Decentralized(P2P), Hierarchical\}. \textit{Cm. E.}: Communication Efficiency. \textit{Cp. E.}: Computation Efficiency.\textit{ Conv}: Convergence Analysis. \textit{Pers}: Personalization. \textit{Gen}: Generalization. \textit{Fair}: Fairness. \textit{Heter}: Heterogeneity. \textit{Sec}: Security. \textit{Priv}: Privacy. \textit{Scal}: Scalability.
\end{tablenotes}
\end{threeparttable}

\end{table}

\subsubsection{Client-Oriented}
    Client-oriented strategies aim to increase the involvement of the most reliable nodes with high-quality data and favorable learning capabilities. The main idea is to mitigate the negative impact of struggling nodes on the overall aggregation performance by carefully choosing the participating clients, evaluating the quality of their updates, or organizing them hierarchically to improve the averaging process. see Table \ref{tab-summary-FL-aggreg-hete-client} for a summary.
\begin{itemize}
        \item \textit{Weighted Aggregation}
        
        The classical aggregation methods (e.g., FedAvg), struggle with real-world non-IID data, leading to suboptimal convergence and utility \cite{khaled2020tighter, hsu2019measuring}. To navigate this drawback, \textit{weighted aggregation} involves assigning weights to local models to discriminate the importance of contributing users. This strategy acknowledges that updates from users with high-quality, relevant data are more valuable for the global model. To tackle this, the FAIR framework rests on three-component solution: 1) estimate each device's contribution quality using historical records, 2) reward high-quality participation through a quality-aware incentive mechanism, and 3) automatically weight local models, ensuring best contributors have a stronger impact on the final model. Other weighting techniques include measuring loss variation \cite{talukder2022computationally} and adopting a hierarchical multi-parameter weighting scheme \cite{herabad2023communication}.
        
        \item \textit{Client Selection}
        
        The objective of carefully choosing a subset of active clients to participate in a communication round is to optimize the performance of federated learning while considering the diverse nature of nodes. The \textit{client selection} methods aim to make the most optimal use of the limited and heterogeneous clients' resources, including data and computing capabilities. Various techniques beyond random selection have been proposed to accelerate the global convergence \cite{fu2022client}. These strategies involve sophisticated perspectives such as employing a scoring system based on past performance \cite{cho2022flame} and prioritizing nodes with superior computing power and accuracy \cite{lu2023towards}.

        \item \textit{Client Clusturing.}

        Client clustering avenue is inspired by the unsupervised clustering ML paradigm. Indeed, it groups clients with similar data distributions, addressing the challenge of data non-IIDness. As an alternative perspective deviating from adopting a single global model in traditional FL, client clustering allows for hierarchical knowledge transfer. Specifically, this involves training task-specific models for each cluster. Subsequently, the server aggregates the models from various clusters to enable high-level knowledge sharing.
        In edge computing environments, both studies \cite{wolfrath2022haccs} and \cite{lu2023auction} leverage clustering to group edge devices based on their similar data distributions. The former work relies on update metadata (e.g., mean, variance), while the later directly analyzes local gradients for more accurate information. In recommendation systems, \cite{imran2023refrs} leverages neural-based clustering to capture user preferences and group semantically similar user models. Another secure solution can be found in in \cite{firdaus2023personalized}.
\end{itemize}

\addtocounter{table}{-1}
\setlength\rotFPtop{0pt plus 1fil} 

\begin{table} [htbp]
\tiny
\caption{Summary of Federated Learning Aggregation Methods for Tackling Heterogeneity - Client-Oriented.}
\label{tab-summary-FL-aggreg-hete-client}
\begin{threeparttable}

\begin{tabularx}{\linewidth}{|X|XXX|XXXXXXXXXXX|}
\hline
\multicolumn{1}{|c|}{\multirow{3}{*}{\textbf{Related Work}}} &
  \multicolumn{3}{c|}{\multirow{2}{*}{\textbf{Environment}}} &
  \multicolumn{11}{c|}{\textbf{\textbf{Verified Goals}}} \\ \cline{5-15} 
 &
  \multicolumn{3}{c|}{} &
  \multicolumn{3}{c|}{\textbf{Process}} &
  \multicolumn{5}{c|}{\textbf{Model}} &
  \multicolumn{3}{c|}{\textbf{System}} \\ \cline{2-15} 
 &
  \multicolumn{1}{c|}{\textbf{Synch}} & 
  \multicolumn{1}{c|}{\textbf{Users}} & 
  \multicolumn{1}{c|}{\textbf{Archi}} & 
  
  \multicolumn{1}{c|}{\textbf{Cm. E.}} & 
  \multicolumn{1}{c|}{\textbf{Cp. E.}} & 
  \multicolumn{1}{c|}{\textbf{Conv}} & 
  \multicolumn{1}{c|}{\textbf{Pers}} & 
  \multicolumn{1}{c|}{\textbf{Gen}} & 
  \multicolumn{1}{c|}{\textbf{Reg}} & 
  \multicolumn{1}{c|}{\textbf{Fair}} & 
  \multicolumn{1}{c|}{\textbf{Heter}} & 
  \multicolumn{1}{c|}{\textbf{Sec}} & 
  \multicolumn{1}{c|}{\textbf{Priv}} & 
  \multicolumn{1}{c|}{\textbf{Scal}} \\  
  \hline

 
  \multicolumn{1}{|c|}{} &
  \multicolumn{14}{c|}{\textbf{Weighted Aggregation}} \\ \hline

  \multicolumn{1}{|c|}{FAIR \cite{deng2022improving}} & 
  \multicolumn{1}{c|}{Synch} &
  \multicolumn{1}{c|}{30} &
  \multicolumn{1}{c|}{Cent} &
  \multicolumn{1}{c|}{\checkmark} &
  \multicolumn{1}{c|}{\checkmark} &
  \multicolumn{1}{c|}{\checkmark} &
  \multicolumn{1}{c|}{--} &
  \multicolumn{1}{c|}{--} &
  \multicolumn{1}{c|}{--} &
  \multicolumn{1}{c|}{--} &
  \multicolumn{1}{c|}{\checkmark} &
  \multicolumn{1}{c|}{--} &
  \multicolumn{1}{c|}{--} &
  \multicolumn{1}{c|}{--} \\ 
  \hline

  \multicolumn{1}{|c|}{RoLePRO \cite{rawat2022robust}} & 
  \multicolumn{1}{c|}{Synch} &
  \multicolumn{1}{c|}{--} &
  \multicolumn{1}{c|}{Cent} &
  \multicolumn{1}{c|}{\checkmark} &
  \multicolumn{1}{c|}{--} &
  \multicolumn{1}{c|}{\checkmark} &
  \multicolumn{1}{c|}{--} &
  \multicolumn{1}{c|}{--} &
  \multicolumn{1}{c|}{--} &
  \multicolumn{1}{c|}{--} &
  \multicolumn{1}{c|}{\checkmark} &
  \multicolumn{1}{c|}{--} &
  \multicolumn{1}{c|}{--} &
  \multicolumn{1}{c|}{--} \\ 
  \hline

  \multicolumn{1}{|c|}{} &
  \multicolumn{14}{c|}{\textbf{Client Selection}} \\ \hline

  \multicolumn{1}{|c|}{FLAME \cite{cho2022flame}} &
  \multicolumn{1}{c|}{Synch} &
  \multicolumn{1}{c|}{149*7devices} &
  \multicolumn{1}{c|}{Cent} &
  \multicolumn{1}{c|}{\checkmark} &
  \multicolumn{1}{c|}{\checkmark} &
  \multicolumn{1}{c|}{\checkmark} &
  \multicolumn{1}{c|}{\checkmark} &
  \multicolumn{1}{c|}{\checkmark} &
  \multicolumn{1}{c|}{--} &
  \multicolumn{1}{c|}{--} &
  \multicolumn{1}{c|}{\checkmark} &
  \multicolumn{1}{c|}{--} &
  \multicolumn{1}{c|}{--} &
  \multicolumn{1}{c|}{--} \\ 
  \hline

  \multicolumn{1}{|c|}{ETTA \cite{lu2023towards}} & 
  \multicolumn{1}{c|}{Synch} &
  \multicolumn{1}{c|}{20} &
  \multicolumn{1}{c|}{Cent} &
  \multicolumn{1}{c|}{\checkmark} &
  \multicolumn{1}{c|}{\checkmark} &
  \multicolumn{1}{c|}{\checkmark} &
  \multicolumn{1}{c|}{--} &
  \multicolumn{1}{c|}{--} &
  \multicolumn{1}{c|}{--} &
  \multicolumn{1}{c|}{--} &
  \multicolumn{1}{c|}{\checkmark} &
  \multicolumn{1}{c|}{\checkmark} &
  \multicolumn{1}{c|}{--} &
  \multicolumn{1}{c|}{--} \\ 
  \hline

  \multicolumn{1}{|c|}{} &
  \multicolumn{14}{c|}{\textbf{Client Clustering}} \\ \hline

  \multicolumn{1}{|c|}{BPFL \cite{firdaus2023personalized}} & 
  \multicolumn{1}{c|}{Cent} &
  \multicolumn{1}{c|}{3} &
  \multicolumn{1}{c|}{Hier} &
  \multicolumn{1}{c|}{\checkmark} &
  \multicolumn{1}{c|}{\checkmark} &
  \multicolumn{1}{c|}{--} &
  \multicolumn{1}{c|}{--} &
  \multicolumn{1}{c|}{--} &
  \multicolumn{1}{c|}{--} &
  \multicolumn{1}{c|}{--} &
  \multicolumn{1}{c|}{\checkmark} &
  \multicolumn{1}{c|}{\checkmark} &
  \multicolumn{1}{c|}{\checkmark} &
  \multicolumn{1}{c|}{--} \\ 
  \hline

  \multicolumn{1}{|c|}{ \cite{lu2023auction}} &
  \multicolumn{1}{c|}{Synch} &
  \multicolumn{1}{c|}{100} &
  \multicolumn{1}{c|}{Hier} &
  \multicolumn{1}{c|}{\checkmark} &
  \multicolumn{1}{c|}{\checkmark} &
  \multicolumn{1}{c|}{\checkmark} &
  \multicolumn{1}{c|}{--} &
  \multicolumn{1}{c|}{\checkmark} &
  \multicolumn{1}{c|}{--} &
  \multicolumn{1}{c|}{--} &
  \multicolumn{1}{c|}{\checkmark} &
  \multicolumn{1}{c|}{--} &
  \multicolumn{1}{c|}{--} &
  \multicolumn{1}{c|}{--} \\ 
  \hline

\end{tabularx}

\begin{tablenotes}
      \footnotesize \item \textit{Synch}: Synchronization mode \{Synchronous, Asynchronous\}. \textit{Users}: Maximum active users used in experiments. \textit{Archi}: Architecture  \{Centralized, Decentralized(P2P), Hierarchical\}. \textit{Cm. E.}: Communication Efficiency. \textit{Cp. E.}: Computation Efficiency.\textit{ Conv}: Convergence Analysis. \textit{Pers}: Personalization. \textit{Gen}: Generalization. \textit{Fair}: Fairness. \textit{Heter}: Heterogeneity. \textit{Sec}: Security. \textit{Priv}: Privacy. \textit{Scal}: Scalability.
\end{tablenotes}
\end{threeparttable}
\end{table}

\subsection{\textbf{Strategies for Communication Efficiency and Optimization}} 
\label{FL_agg_solution_communication}

\begin{figure}[htbp]
    \centering
    \includegraphics[scale=0.4]{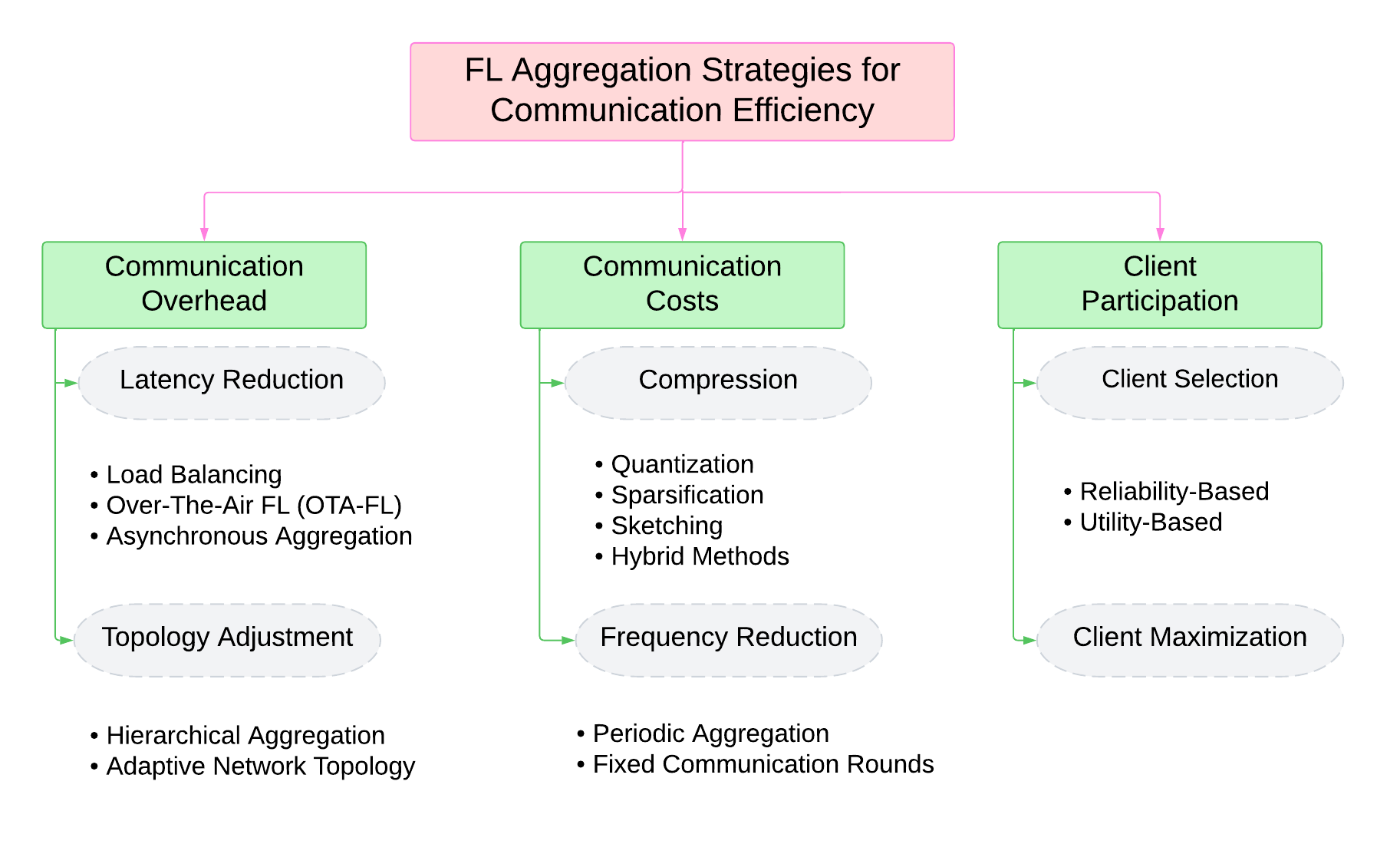}
    \caption{Classification of Federated Learning Aggregation Strategies for Communication Efficiency.}
    \label{FL_strategies_Communication}
\end{figure}

Communication efficiency is one essential aspect of FL, yet it often poses a critical bottleneck. As outlined in Section \ref{communication_efficiency_in_FL}, factors like limited bandwidth, latency, and large models hinder efficient communication. Various optimization strategies have emerged in the literature, aiming to surmount these hurdles and expedite convergence. Fig. \ref{FL_strategies_Communication} provides a visual taxonomy of these practices, which we will delve into in this section.

\subsubsection{Communication Overhead}

    Typically, communication overhead arises when multiple transmitters attempt to send a significant amount of data to a central hub, as is the case with the client-server model exchange in federated training. This overhead is further exacerbated due to the limited resources and network bandwidth. To position the available solutions, we categorize them into two main approaches: \textit{reducing training latency} and \textit{adapting the network topology}.
    Table \ref{tab-summary-FL-aggreg-comm-overhead} enumerates a selection of recent endeavors falling within the purview of these two methodologies.
    
\begin{itemize}
    \item \textit{Training Latency Reduction.}
    
    On the client side, the training latency is controlled by the local capabilities and workload. Despite the immutability of the former, the latter leaves room for incorporating inventive adjustments. We introduce in this category the subsequent solutions, including load balancing, Over-The-Air (OTA)-FL, and asynchronous aggregation. 

    \textit{\textbf{Load Balancing}.} 
    In order to mitigate the severe effect of FL clients' disparities, many researchers suggest balancing the amount of data retained by each client. One such approach consists of turning RL techniques to determine the optimal amount of data that must be used in a training round for each client \cite{shin2023federated}. For instance, in \cite{tu2020network}, researchers have derived a convex optimization accounting for local resources and model accuracy to determine the most suitable clients for processing specific data points. In another investigation, the CATA-Fed system \cite{jeong2022cluster} considers data diversity and data load when selecting clients. The goal of this scheduling algorithm is twofold: global bias alleviation with fair workload distribution. More strategies can be found in \cite{trindade2022resource, morell2022optimising}.

    \textit{\textbf{Over-The-Air Computation.}}
    Over-the-air computation is a powerful strategy that strikes a balance between data privacy and effective knowledge sharing. Technically speaking, over-the-air computation stands for computing a nomographic function (e.g., mean) of distributed data from multiple transmitters. In contrast to the traditional computation-communication separation methods in FL, this design carries out parts of the aggregation process \textit{over-the-air} by local devices to reduce the required bandwidth, thereby, speeding up the federated training process. In other words, the summation of the models' gradient is carried out over the air in a distributed manner, and only the sums (not all the gradients) are transmitted to the aggregator to finally compute the sums' average.
    Amidst numerous research endeavors \cite{guo2021over, sifaou2022over}, the researchers in \cite{yang2022over} have explored the impact of noisy communication channels on OTA-FL. The developed ACPC-OTA-FL algorithm allows each client to flexibly ascertain its transmission power level and the number of local update steps, maximizing the utilization of available resources. In contrast, the CHARLES framework \cite{mao2022charles} leverages the estimation of Channel State Information (CSI) to assess the effects of fading wireless channels within heterogeneous OTA-FL scenarios.

    \textit{\textbf{Asynchronous aggregation.}}
     Observing the numerous limitations caused by the synchronicity assumption of the standard FL ecosystem, the research intention has been directed to release the federated learning environments from this strict requirement by leveraging the asynchronous aggregation scheme \cite{yang2022efficient, taha2022unbounded}.

     \item \textit{Network Topology Adjustment.}
     
     Network topology refers to an architectural design that dedicates the connection and devices within a network. However, the conventional star topology commonly used in FL may not always be an efficient choice. Therefore, some researchers advise modifying the network design to enhance system efficiency. Intuitively, different topologies inherently incur varying communication costs, and then hopefully, selecting an optimized network topology will help achieve the desired efficiency level. To present the existing solutions in this regard, we bring two extensively adopted FL structures: \textit{hierarchical aggregation} and \textit{adaptive network topology}. 
     
     \textit{\textbf{Hierarchical Aggregation.}}
     The hierarchical FL embraces a tree-like structure, allowing for partial aggregation \cite{su2021secure}. Proximate devices upload their updates to a small base station (SBS) for initial aggregation. Subsequently, a macro base station (MBS) conducts the final aggregation \cite{10062841}. In this setup, leveraging edge devices as SBSs within the modern FL systems has shown great potential in reducing communication bottlenecks \cite{yi2022hfedmtl, singh2022dew}. 

     \textit{\textbf{Adaptive Network Topology.}}
     The interplay of communication topology and training duration dilemma has attracted significant focus. From one side, a highly connected topology expedites the FL convergence. On the flip side, a more connected topology can also prolong the duration of a communication round. One contribution in this domain \cite{marfoq2020throughput} relies on the max-plus algebra and leverages quantifiable network characteristics, including computation times, link latency, and transmission times to optimize system throughput, measured by the number of completed rounds per time unit. A different study \cite{zhou2022communication} proposed a novel client-server interaction design. This approach empowers each client to decide whether or not to share its model updates based on two thresholds: a probability threshold that limits update frequency to prevent server overload and an informative model determination threshold that ensures only informative updates are transmitted.

  \end{itemize}


\addtocounter{table}{-1}

\setlength\rotFPtop{0pt plus 1fil} 
\begin{table} [htbp]
\tiny
\caption{Summary of Federated Learning Aggregation Methods for Tackling Communication Efficiency - Communication Overhead.}
\label{tab-summary-FL-aggreg-comm-overhead}
\begin{threeparttable}

\begin{tabularx}{\linewidth}{|X|XXX|XXXXXXXXXXX|}
\hline
\multicolumn{1}{|c|}{\multirow{3}{*}{\textbf{Related Work}}} &
  \multicolumn{3}{c|}{\multirow{2}{*}{\textbf{Environment}}} &
  \multicolumn{11}{c|}{\textbf{\textbf{Verified Goals}}} \\ \cline{5-15} 
 &
  \multicolumn{3}{c|}{} &
  \multicolumn{3}{c|}{\textbf{Process}} &
  \multicolumn{5}{c|}{\textbf{Model}} &
  \multicolumn{3}{c|}{\textbf{System}} \\ \cline{2-15} 
 &
  \multicolumn{1}{c|}{\textbf{Synch}} & 
  \multicolumn{1}{c|}{\textbf{Users}} & 
  \multicolumn{1}{c|}{\textbf{Archi}} & 
  
  \multicolumn{1}{c|}{\textbf{Cm. E.}} & 
  \multicolumn{1}{c|}{\textbf{Cp. E.}} & 
  \multicolumn{1}{c|}{\textbf{Conv}} & 
  \multicolumn{1}{c|}{\textbf{Pers}} & 
  \multicolumn{1}{c|}{\textbf{Gen}} & 
  \multicolumn{1}{c|}{\textbf{Reg}} & 
  \multicolumn{1}{c|}{\textbf{Fair}} & 
  \multicolumn{1}{c|}{\textbf{Heter}} & 
  \multicolumn{1}{c|}{\textbf{Sec}} & 
  \multicolumn{1}{c|}{\textbf{Priv}} & 
  \multicolumn{1}{c|}{\textbf{Scal}} \\  
  \hline

%
  \multicolumn{1}{|c|}{\textbf{Used Techniques}} &
  \multicolumn{14}{c|}{\textbf{Training Latency Reduction}}  \\ \hline
  \multicolumn{1}{|c|}{\textbf{Load Balancing}} &
  \multicolumn{14}{c|}{}  \\ \hline

  \multicolumn{1}{|c|}{\cite{tu2020network}} &  
  \multicolumn{1}{c|}{Synch} &
  \multicolumn{1}{c|}{10} &
  \multicolumn{1}{c|}{Hier} &
  \multicolumn{1}{c|}{\checkmark} &
  \multicolumn{1}{c|}{\checkmark} &
  \multicolumn{1}{c|}{\checkmark} &
  \multicolumn{1}{c|}{--} &
  \multicolumn{1}{c|}{--} &
  \multicolumn{1}{c|}{--} &
  \multicolumn{1}{c|}{--} &
  \multicolumn{1}{c|}{\checkmark} &
  \multicolumn{1}{c|}{--} &
  \multicolumn{1}{c|}{--} &
  \multicolumn{1}{c|}{--}  \\ 
  \hline

  \multicolumn{1}{|c|}{CATA-Fed \cite{jeong2022cluster}} & 
  \multicolumn{1}{c|}{Synch} &
  \multicolumn{1}{c|}{40} &
  \multicolumn{1}{c|}{Hier} &
  \multicolumn{1}{c|}{\checkmark} &
  \multicolumn{1}{c|}{\checkmark} &
  \multicolumn{1}{c|}{\checkmark} &
  \multicolumn{1}{c|}{\checkmark} &
  \multicolumn{1}{c|}{\checkmark} &
  \multicolumn{1}{c|}{--} &
  \multicolumn{1}{c|}{\checkmark} &
  \multicolumn{1}{c|}{\checkmark} &
  \multicolumn{1}{c|}{--} &
  \multicolumn{1}{c|}{--} &
  \multicolumn{1}{c|}{\checkmark}  \\ 
  \hline

  \multicolumn{1}{|c|}{\textbf{OTA-FL}} &
  \multicolumn{14}{c|}{}  \\ \hline

  \multicolumn{1}{|c|}{ACPC-OTA-FL\cite{yang2022over}} & 
  \multicolumn{1}{c|}{Synch} &
  \multicolumn{1}{c|}{10} &
  \multicolumn{1}{c|}{Cent} &
  \multicolumn{1}{c|}{\checkmark} &
  \multicolumn{1}{c|}{\checkmark} &
  \multicolumn{1}{c|}{\checkmark} &
  \multicolumn{1}{c|}{\checkmark} &
  \multicolumn{1}{c|}{--} &
  \multicolumn{1}{c|}{--} &
  \multicolumn{1}{c|}{--} &
  \multicolumn{1}{c|}{\checkmark} &
  \multicolumn{1}{c|}{--} &
  \multicolumn{1}{c|}{--} &
  \multicolumn{1}{c|}{--}  \\ 
  \hline

  \multicolumn{1}{|c|}{CHARLES \cite{mao2022charles}} & 
  \multicolumn{1}{c|}{Synch} &
  \multicolumn{1}{c|}{10} &
  \multicolumn{1}{c|}{Cent} &
  \multicolumn{1}{c|}{\checkmark} &
  \multicolumn{1}{c|}{\checkmark} &
  \multicolumn{1}{c|}{\checkmark} &
  \multicolumn{1}{c|}{\checkmark} &
  \multicolumn{1}{c|}{--} &
  \multicolumn{1}{c|}{--} &
  \multicolumn{1}{c|}{--} &
  \multicolumn{1}{c|}{\checkmark} &
  \multicolumn{1}{c|}{--} &
  \multicolumn{1}{c|}{--} &
  \multicolumn{1}{c|}{--}  \\ 
  \hline

  \multicolumn{1}{|c|}{\textbf{Asynch. Agg.}} &
  \multicolumn{14}{c|}{}  \\ \hline

  \multicolumn{1}{|c|}{EHAFL\cite{yang2022efficient}} & 
  \multicolumn{1}{c|}{Asynch} &
  \multicolumn{1}{c|}{100} &
  \multicolumn{1}{c|}{Hier} &
  \multicolumn{1}{c|}{\checkmark} &
  \multicolumn{1}{c|}{\checkmark} &
  \multicolumn{1}{c|}{--} &
  \multicolumn{1}{c|}{--} &
  \multicolumn{1}{c|}{\checkmark} &
  \multicolumn{1}{c|}{--} &
  \multicolumn{1}{c|}{--} &
  \multicolumn{1}{c|}{\checkmark} &
  \multicolumn{1}{c|}{\checkmark} &
  \multicolumn{1}{c|}{--} &
  \multicolumn{1}{c|}{--}  \\ 
  \hline

  \multicolumn{1}{|c|}{} &
  \multicolumn{14}{c|}{\textbf{Network Topology Adjustment}}  \\ \hline
  \multicolumn{1}{|c|}{\textbf{Hier. Agg.}} &
  \multicolumn{14}{c|}{}  \\ \hline

  \multicolumn{1}{|c|}{\cite{su2021secure}} & 
  \multicolumn{1}{c|}{Synch } &
  \multicolumn{1}{c|}{50} &
  \multicolumn{1}{c|}{Hier} &
  \multicolumn{1}{c|}{\checkmark} &
  \multicolumn{1}{c|}{\checkmark} &
  \multicolumn{1}{c|}{\checkmark} &
  \multicolumn{1}{c|}{\checkmark} &
  \multicolumn{1}{c|}{\checkmark} &
  \multicolumn{1}{c|}{--} &
  \multicolumn{1}{c|}{--} &
  \multicolumn{1}{c|}{\checkmark} &
  \multicolumn{1}{c|}{\checkmark} &
  \multicolumn{1}{c|}{\checkmark} &
  \multicolumn{1}{c|}{--}  \\ 
  \hline

  \multicolumn{1}{|c|}{HFL-HLSTM \cite{singh2022dew}} & 
  \multicolumn{1}{c|}{Cent} &
  \multicolumn{1}{c|}{120} &
  \multicolumn{1}{c|}{Hier} &
  \multicolumn{1}{c|}{\checkmark} &
  \multicolumn{1}{c|}{\checkmark} &
  \multicolumn{1}{c|}{--} &
  \multicolumn{1}{c|}{--} &
  \multicolumn{1}{c|}{--} &
  \multicolumn{1}{c|}{--} &
  \multicolumn{1}{c|}{--} &
  \multicolumn{1}{c|}{--} &
  \multicolumn{1}{c|}{\checkmark} &
  \multicolumn{1}{c|}{\checkmark} &
  \multicolumn{1}{c|}{--}  \\ 
  \hline

  \multicolumn{1}{|c|}{\textbf{ ANT}} &
  \multicolumn{14}{c|}{}  \\ \hline

  \multicolumn{1}{|c|}{\cite{marfoq2020throughput}} &
  \multicolumn{1}{c|}{Synch} &
  \multicolumn{1}{c|}{87} &
  \multicolumn{1}{c|}{Decent} &
  \multicolumn{1}{c|}{\checkmark} &
  \multicolumn{1}{c|}{\checkmark} &
  \multicolumn{1}{c|}{\checkmark} &
  \multicolumn{1}{c|}{--} &
  \multicolumn{1}{c|}{--} &
  \multicolumn{1}{c|}{--} &
  \multicolumn{1}{c|}{--} &
  \multicolumn{1}{c|}{--} &
  \multicolumn{1}{c|}{--} &
  \multicolumn{1}{c|}{--} &
  \multicolumn{1}{c|}{--}  \\ 
  \hline

  \multicolumn{1}{|c|}{\cite{zhou2022communication}} & 
  \multicolumn{1}{c|}{Synch} &
  \multicolumn{1}{c|}{20} &
  \multicolumn{1}{c|}{Cent} &
  \multicolumn{1}{c|}{\checkmark} &
  \multicolumn{1}{c|}{--} &
  \multicolumn{1}{c|}{--} &
  \multicolumn{1}{c|}{--} &
  \multicolumn{1}{c|}{--} &
  \multicolumn{1}{c|}{--} &
  \multicolumn{1}{c|}{--} &
  \multicolumn{1}{c|}{\checkmark} &
  \multicolumn{1}{c|}{--} &
  \multicolumn{1}{c|}{--} &
  \multicolumn{1}{c|}{--}  \\ 
  \hline
  
\end{tabularx}

\begin{tablenotes}
      \footnotesize \item \textit{Synch}: Synchronization mode \{Synchronous, Asynchronous\}. \textit{Users}: Maximum active users used in experiments. \textit{Archi}: Architecture  \{Centralized, Decentralized(P2P), Hierarchical\}. \textit{Cm. E.}: Communication Efficiency. \textit{Cp. E.}: Computation Efficiency.\textit{ Conv}: Convergence Analysis. \textit{Pers}: Personalization. \textit{Gen}: Generalization. \textit{Fair}: Fairness. \textit{Heter}: Heterogeneity. \textit{Sec}: Security. \textit{Priv}: Privacy. \textit{Scal}: Scalability. \textit{ANT}: Adaptive Network Topology.
\end{tablenotes}
\end{threeparttable}

\end{table}

   \subsubsection{Communication Costs}
   
    An additional avenue of focus in the efficiency field pertains to reducing data transmission costs. Several factors amplify the increased expenses associated with FL communication. Although many environmental factors remain beyond one's control, other flexible elements related to the aggregation process offer opportunities for valuable contributions. For instance, the model size and the aggregation frequency present compelling areas for investigation. In Table \ref{tab-summary-FL-aggreg-comm-cost}, we offer a compilation of the surveyed papers that converge with this specific domain.

   \begin{itemize}
    
     \item{Model Size Reduction}
    
     Another captivating strategy to reduce communication overhead, deeply explored in Section \ref{FL_agg_solution_sec_priv}, is reducing the number of model parameters transmitted. Researchers argue that it is neither necessary nor efficient that each client must download and upload the complete in the federation process. Instead, they propose alternatives where only a relevant fraction of the global model is assigned to each client to be updated locally. One such configuration relies on \textit{model division}, whether it is layer-based or parameter-representation-based. The second configuration of \textit{update dropping} trims less significant neurons from the global model. Beyond model architecture changes, the \textit{compression} strategy has gained significant interest from the academic communities aiming to reduce energy consumption. Generally described, compression is the process by which the information is encoded using a smaller number of bits than that of the original representation, making it suitable for bandwidth-limited channels and lower energy requirements. Prior arts in this specific field fall into three lines as follows: 

    \textit{\textbf{Quantization.}}
    Quantization involves representing the neural network's weights and activations with lower precision (i.e., fewer bits). For instance, by reducing the number of colors in a digital image, the file becomes smaller and takes less space. Similarly, quantization maps the full-precision floating point to a smaller set, making the model lighter for transmission.
    In the pursuit of communication efficiency, QD-Compressor \cite{jin2023design} is a dedicated proposal to large-scale DNN snapshots, precisely for failure-prone clusters in FL. Due to variations in parameter value ranges across layers, the Local-Sensitive Quantization module employs a layer-specific quantization strategy to dynamically adjust quantizers and the number of quantization bits among layers. Moreover, the Error Feedback Mechanism helps maintain high-quality restored models by averting quantization errors.  In the same trajectory, the JoPEQ framework introduced in \cite{lang2023joint} jointly integrates lossy compression and privacy augmentation strategies. The core idea lies in employing vector quantization to exploit the incurred distortion in injecting noises into model updates, fortifying the FL system against privacy breaches. 
    
    \textit{\textbf{Sketching.}}
    The previous quantization mechanisms support only uniformly distributed data, which is not always the case for FL updates. The sketch algorithm is a probabilistic alternative that can retain key statistical properties and enable meaningful analysis. The sketching is a memory-saving solution that is used to estimate the model updates distribution in a single processing pass over the updates values. For instance, FetchSGD \cite{rothchild2020fetchsgd} incorporates Count Sketch with momentum and error accumulation, enabling efficient communication with good recovery guarantees.
    While the proposal in \cite{kollias2023sketch} leverages Locality Sensitive Hashing (LSH) sketching. This technique relies on the predicate that if two distant models are approximately “close,” there is no need to share them.

    \textit{\textbf{Sparsification.}}
    Intuitively, sparsification is the process by which a matrix becomes more sparse. A sparse matrix is a matrix whose zero elements are higher than those non-zero elements. The main advantage of having a sparse matrix is to save space by storing only the non-zero elements. Sparsification techniques are employed in federated learning to filter out and preserve only the most important parts of the locally trained models.
    Standard sparsification in FL can enhance communication efficiency but it may expose sensitive model parameters during aggregation. The work in \cite{ergun2021sparsified} tackles this concern with SparseSecAggn, an approach for secure sparsification in FL using pairwise sparsification. Shared random masks between devices ensure their aggregated masked models nullify each other. For model compression in analog FL, the authors in \cite{ahn2022model} suggested a novel lossless compression technique called Pattern-Shared Sparsification (PSS). Unlike prior methods where devices sparsify gradients independently (i.e., local top-k sparsification), PSS utilizes a collective sparsification pattern across all devices.
    Other scholars have demonstrated that the hybridization of the aforementioned compression mechanisms offers a potent resolution, as exemplified in \cite{malekijoo2021fedzip}.

    To summarize, a perfect compression method does not exist. It either compromises information loss or costs higher computation. From this computation-information trade-off lens, two primary approaches exist lossy and lossless compression. \textit{Lossy compression} methods prioritize efficiency by sacrificing some information fidelity \cite{liu2021hierarchical}. Conversely, \textit{lossless compression} assures no information loss but comes at the cost of higher computational requirements \cite{zhao2022cork}.
    
    \item \textit{Aggregation Frequency Reduction.}
    
    In a notable finding from the work in \cite{lin2017deep}, it was demonstrated that a remarkable 99\% of the gradients exchanged during the FL communication process are redundant. Furthermore, transmitting voluminous model parameters strains network resources and elongates the required time for convergence. To combat this, it proves advantageous to diminish the frequency at which clients forward their updates to the aggregator. In this regard, we bring two approaches, as found in the existing literature: \textit{periodic aggregation} and \textit{fixed communication rounds}.
    
    \textit{\textbf{Periodic Aggregation.}}
    As you may guess, this enables the client to perform more than one training iteration prior to sending their updates to the aggregator server. Therefore, the need for client-server communication will be lessened. Interestingly, the research in \cite{mohammadi2021differential} concurrently examined three mechanisms to enhance communication efficiency while maintaining privacy in FL. These mechanisms encompass period aggregation, model compression, and client participation scheduling. In a similar vein, the TAMUNA framework \cite{condat2023tamuna} employs infrequent communication on top of transmitting compressed models to alleviate the communication burden.

    \textit{\textbf{Fixed Communication Round.}} 
    It refers to a known approach of reducing aggregation frequency, in which the practice of collecting and aggregating model updates from active clients is only performed at regular intervals, generally defined by a fixed number of local iterations. For example, the efforts in \cite{mhaisen2021optimal} and \cite{li2023hfml} have emphasized this concept in the context of hierarchical FL. While the paper in \cite{hu2023scheduling} considered asynchronous FL setup. 
    
\end{itemize}
\addtocounter{table}{-1}

\setlength\rotFPtop{0pt plus 1fil} 
\begin{table} [htbp]
\tiny
\caption{Summary of Federated Learning Aggregation Methods for Tackling Communication Efficiency - Communication Costs.}
\label{tab-summary-FL-aggreg-comm-cost}
\begin{threeparttable}

\begin{tabularx}{\linewidth}{|X|XXX|XXXXXXXXXXX|}
\hline
\multicolumn{1}{|c|}{\multirow{3}{*}{\textbf{Related Work}}} &
  \multicolumn{3}{c|}{\multirow{2}{*}{\textbf{Environment}}} &
  \multicolumn{11}{c|}{\textbf{\textbf{Verified Goals}}} \\ \cline{5-15} 
 &
  \multicolumn{3}{c|}{} &
  \multicolumn{3}{c|}{\textbf{Process}} &
  \multicolumn{5}{c|}{\textbf{Model}} &
  \multicolumn{3}{c|}{\textbf{System}} \\ \cline{2-15} 
 &
  \multicolumn{1}{c|}{\textbf{Synch}} & 
  \multicolumn{1}{c|}{\textbf{Users}} & 
  \multicolumn{1}{c|}{\textbf{Archi}} & 
  
  \multicolumn{1}{c|}{\textbf{Cm. E.}} & 
  \multicolumn{1}{c|}{\textbf{Cp. E.}} & 
  \multicolumn{1}{c|}{\textbf{Conv}} & 
  \multicolumn{1}{c|}{\textbf{Pers}} & 
  \multicolumn{1}{c|}{\textbf{Gen}} & 
  \multicolumn{1}{c|}{\textbf{Reg}} & 
  \multicolumn{1}{c|}{\textbf{Fair}} & 
  \multicolumn{1}{c|}{\textbf{Heter}} & 
  \multicolumn{1}{c|}{\textbf{Sec}} & 
  \multicolumn{1}{c|}{\textbf{Priv}} & 
  \multicolumn{1}{c|}{\textbf{Scal}} \\  
  \hline

%
  \multicolumn{1}{|c|}{\textbf{Used Techniques}} &
  \multicolumn{14}{c|}{\textbf{Model Size Reduction - Compression}}  \\ \hline
  \multicolumn{1}{|c|}{\textbf{Quantization}} &
  \multicolumn{14}{c|}{}  \\ \hline

  \multicolumn{1}{|c|}{JoPEQ\cite{lang2023joint}} & 
  \multicolumn{1}{c|}{Synch} &
  \multicolumn{1}{c|}{10} &
  \multicolumn{1}{c|}{Cent} &
  \multicolumn{1}{c|}{\checkmark} &
  \multicolumn{1}{c|}{--} &
  \multicolumn{1}{c|}{\checkmark} &
  \multicolumn{1}{c|}{--} &
  \multicolumn{1}{c|}{--} &
  \multicolumn{1}{c|}{--} &
  \multicolumn{1}{c|}{--} &
  \multicolumn{1}{c|}{\checkmark} &
  \multicolumn{1}{c|}{--} &
  \multicolumn{1}{c|}{\checkmark} &
  \multicolumn{1}{c|}{--}  \\ 
  \hline
    
  \multicolumn{1}{|c|}{LAQ \cite{sun2020lazily}} & %
  \multicolumn{1}{c|}{Synch} &
  \multicolumn{1}{c|}{10} &
  \multicolumn{1}{c|}{Cent} &
  \multicolumn{1}{c|}{\checkmark} &
  \multicolumn{1}{c|}{--} &
  \multicolumn{1}{c|}{\checkmark} &
  \multicolumn{1}{c|}{--} &
  \multicolumn{1}{c|}{--} &
  \multicolumn{1}{c|}{--} &
  \multicolumn{1}{c|}{--} &
  \multicolumn{1}{c|}{\checkmark} &
  \multicolumn{1}{c|}{--} &
  \multicolumn{1}{c|}{--} &
  \multicolumn{1}{c|}{--}  \\ 
  \hline

  \multicolumn{1}{|c|}{\textbf{Sparsification}} &
  \multicolumn{14}{c|}{}  \\ \hline
   \multicolumn{1}{|c|}{PSS  \cite{ahn2022model}} &
  \multicolumn{1}{c|}{Synch} &
  \multicolumn{1}{c|}{100} & 
  \multicolumn{1}{c|}{Cent} &
  \multicolumn{1}{c|}{\checkmark} &
  \multicolumn{1}{c|}{\checkmark} &
  \multicolumn{1}{c|}{\checkmark} &
  \multicolumn{1}{c|}{--} &
  \multicolumn{1}{c|}{--} &
  \multicolumn{1}{c|}{--} &
  \multicolumn{1}{c|}{--} &
  \multicolumn{1}{c|}{--} &
  \multicolumn{1}{c|}{--} &
  \multicolumn{1}{c|}{--} &
  \multicolumn{1}{c|}{--}  \\ 
  \hline

  \multicolumn{1}{|c|}{SparseSecAgg \cite{ergun2021sparsified}} & 
  \multicolumn{1}{c|}{Synch} &
  \multicolumn{1}{c|}{100} &
  \multicolumn{1}{c|}{Cent} &
  \multicolumn{1}{c|}{\checkmark} &
  \multicolumn{1}{c|}{\checkmark} &
  \multicolumn{1}{c|}{\checkmark} &
  \multicolumn{1}{c|}{--} &
  \multicolumn{1}{c|}{--} &
  \multicolumn{1}{c|}{--} &
  \multicolumn{1}{c|}{--} &
  \multicolumn{1}{c|}{--} &
  \multicolumn{1}{c|}{\checkmark} &
  \multicolumn{1}{c|}{\checkmark} &
  \multicolumn{1}{c|}{--} \\ 
  \hline

  \multicolumn{1}{|c|}{\textbf{Sketching}} &
  \multicolumn{14}{c|}{}  \\ \hline

  \multicolumn{1}{|c|}{\cite{kollias2023sketch}} & 
  \multicolumn{1}{c|}{Synch} &
  \multicolumn{1}{c|}{10} & 
  \multicolumn{1}{c|}{Hier} &
  \multicolumn{1}{c|}{\checkmark} &
  \multicolumn{1}{c|}{\checkmark} &
  \multicolumn{1}{c|}{\checkmark} &
  \multicolumn{1}{c|}{--} &
  \multicolumn{1}{c|}{--} &
  \multicolumn{1}{c|}{--} &
  \multicolumn{1}{c|}{--} &
  \multicolumn{1}{c|}{\checkmark} &
  \multicolumn{1}{c|}{--} &
  \multicolumn{1}{c|}{--} &
  \multicolumn{1}{c|}{--}  \\ 
  \hline

  \multicolumn{1}{|c|}{Fetchsgd \cite{rothchild2020fetchsgd}} & 
  \multicolumn{1}{c|}{Synch} &
  \multicolumn{1}{c|}{10000} & 
  \multicolumn{1}{c|}{Cent} &
  \multicolumn{1}{c|}{\checkmark} &
  \multicolumn{1}{c|}{--} &
  \multicolumn{1}{c|}{\checkmark} &
  \multicolumn{1}{c|}{--} &
  \multicolumn{1}{c|}{--} &
  \multicolumn{1}{c|}{--} &
  \multicolumn{1}{c|}{--} &
  \multicolumn{1}{c|}{\checkmark} &
  \multicolumn{1}{c|}{--} &
  \multicolumn{1}{c|}{--} &
  \multicolumn{1}{c|}{\checkmark} \\ 
  \hline

  \multicolumn{1}{|c|}{\textbf{Hybrid}} &
  \multicolumn{14}{c|}{}  \\ \hline
  \multicolumn{1}{|c|}{LGC \cite{abrahamyan2021learned}} & 
  \multicolumn{1}{c|}{Synch} &
  \multicolumn{1}{c|}{8} &
  \multicolumn{1}{c|}{Cent} &
  \multicolumn{1}{c|}{\checkmark} &
  \multicolumn{1}{c|}{--} &
  \multicolumn{1}{c|}{\checkmark} &
  \multicolumn{1}{c|}{--} &
  \multicolumn{1}{c|}{--} &
  \multicolumn{1}{c|}{--} &
  \multicolumn{1}{c|}{--} &
  \multicolumn{1}{c|}{--} &
  \multicolumn{1}{c|}{--} &
  \multicolumn{1}{c|}{--} &
  \multicolumn{1}{c|}{--} \\ 
  \hline

  \multicolumn{1}{|c|}{FedZIP \cite{malekijoo2021fedzip}} & 
  \multicolumn{1}{c|}{Synch} &
  \multicolumn{1}{c|}{50} &
  \multicolumn{1}{c|}{Hierarch} &
  \multicolumn{1}{c|}{\checkmark} &
  \multicolumn{1}{c|}{\checkmark} &
  \multicolumn{1}{c|}{\checkmark} &
  \multicolumn{1}{c|}{--} &
  \multicolumn{1}{c|}{--} &
  \multicolumn{1}{c|}{--} &
  \multicolumn{1}{c|}{--} &
  \multicolumn{1}{c|}{\checkmark} &
  \multicolumn{1}{c|}{--} &
  \multicolumn{1}{c|}{--} &
  \multicolumn{1}{c|}{\checkmark} \\ 
  \hline

  \multicolumn{1}{|c|}{} &
  \multicolumn{14}{c|}{\textbf{Aggregation Frequency Reduction}}  \\ \hline
  \multicolumn{1}{|c|}{\textbf{P.A.}} &
  \multicolumn{14}{c|}{}  \\ \hline

  \multicolumn{1}{|c|}{FedPAQ  \cite{mohammadi2021differential}} & 
  \multicolumn{1}{c|}{Synch} &
  \multicolumn{1}{c|}{10} &
  \multicolumn{1}{c|}{Cent} &
  \multicolumn{1}{c|}{\checkmark} &
  \multicolumn{1}{c|}{\checkmark} &
  \multicolumn{1}{c|}{\checkmark} &
  \multicolumn{1}{c|}{--} &
  \multicolumn{1}{c|}{--} &
  \multicolumn{1}{c|}{\checkmark} &
  \multicolumn{1}{c|}{--} &
  \multicolumn{1}{c|}{\checkmark} &
  \multicolumn{1}{c|}{--} &
  \multicolumn{1}{c|}{\checkmark} &
  \multicolumn{1}{c|}{--}  \\ 
  \hline

  \multicolumn{1}{|c|}{TAMUNA \cite{condat2023tamuna}} & 
  \multicolumn{1}{c|}{Asynch} &
  \multicolumn{1}{c|}{1000} &
  \multicolumn{1}{c|}{Cent} &
  \multicolumn{1}{c|}{\checkmark} &
  \multicolumn{1}{c|}{--} &
  \multicolumn{1}{c|}{\checkmark} &
  \multicolumn{1}{c|}{--} &
  \multicolumn{1}{c|}{--} &
  \multicolumn{1}{c|}{--} &
  \multicolumn{1}{c|}{--} &
  \multicolumn{1}{c|}{\checkmark} &
  \multicolumn{1}{c|}{--} &
  \multicolumn{1}{c|}{--} &
  \multicolumn{1}{c|}{--} \\ 
  \hline
  
\end{tabularx}

\begin{tablenotes}
      \footnotesize \item \textit{Synch}: Synchronization mode \{Synchronous, Asynchronous\}. \textit{Users}: Maximum active users used in experiments. \textit{Archi}: Architecture  \{Centralized, Decentralized(P2P), Hierarchical\}. \textit{Cm. E.}: Communication Efficiency. \textit{Cp. E.}: Computation Efficiency.\textit{ Conv}: Convergence Analysis. \textit{Pers}: Personalization. \textit{Gen}: Generalization. \textit{Fair}: Fairness. \textit{Heter}: Heterogeneity. \textit{Sec}: Security. \textit{Priv}: Privacy. \textit{Scal}: Scalability. \textit{P.A.}: Periodic Aggregation.
\end{tablenotes}
\end{threeparttable}

\end{table}

\subsubsection{Client Participation} 
The objective of thoughtful client selection is to effectively make decisions on the aggregation level, enabling meaningful knowledge extraction from the selected clients. The challenges here stem from the significant difference in the data quality and quantity held by each device as well as their calculation powers. Besides, the present problem is aggravated by the size of the current deep neural networks, which can be in the range of millions of parameters, resulting in tremendous bandwidth consumption. This underscores the importance of designing a sophisticated client participation scheme that takes into consideration all of these factors to conduct more efficient and optimized federated aggregation protocols. In this section, we discuss three main state-of-the-art classes: reliability-based client selection, utility-based client selection, and client maximization. We summarize these insights in Table \ref{tab-summary-FL-aggreg-client-participation}.

\begin{itemize}
    \item \textit{Client Selection}
     
    \textit{\textbf{Reliability-based Client Selection.}}
    Reliability-based client selection focuses on selecting clients based on their reliability or trustworthiness. It considers their past performance, such as their consistency in joining the collaborative training, their ability to complete training tasks, and providing accurate updates. The goal is to prioritize clients that have proven reliable for not stumbling the smoothness of communications (e.g., that do not drop out from the training, surprisingly), ensuring FL system stability.
    To illustrate, we consider the approach in \cite{rjoub2022trust}. This strategy harnesses a trust-based Deep-RL mechanism tailored to select adequate clients in a resource-efficient and time-conscious manner.  In a similar wavelength, the scoring-aided FL framework in \cite{zhang2023scoring} chooses FL mobile clients based on their distinctive patterns, focusing on precision and efficiency.

    \textit{\textbf{Utility-based Client Selection.}} 
    In contrast, utility-based client selection emphasizes selecting clients based on their utility or usefulness for the specific learning task. It evaluates factors such as the potential impact of the client's data and updates on improving the model's generalization performance  \cite{egtgsv24iwcmc}. By strategically selecting clients with the most valuable contributions, this approach accelerates long-term convergence, requiring fewer communication rounds to achieve the target performance.
    To exemplify, the paper in \cite{shi2023efficient} enables the quantification of individual client contributions to the broader global model. The authors explore the Combinatorial Multi-Arm Bandit(MAB) strategies to propose the CU-CS scheme that intelligently allocates resources based on clients' impact. On a related reference \cite{huang2023active}, the ACFL method pinpoints highly informative clients within each cluster through Active Learning (AL) metrics, to enhance cluster-specific models.

    \item \textit{Client Number Maximization.}
    
    Intuitively, increasing the number of participating clients should decrease the time required to achieve convergence. This "time-to-converge" is essentially determined by the number of communication rounds needed in order to reach the desired FL performance. Although this might seem promising, it is worth noting that simply maximizing the number of clients will not necessarily achieve excellent FL system efficiency without undesirable costs (e.g., the impact on the overall non-IIDness and the model uploading waiting delays). 
    Due to the conflicting-objective nature of this problem, its proper formulation necessitates \textit{multi-objective optimization} functions.
    The authors in \cite{ji2022client} formulate a mixed-integer optimization for wireless FL, focusing on client selection and bandwidth allocation. The resulting Perround Energy Drift Plus Cost (PEDPC) algorithm translates the original offline problem into an online perspective for minimizing latency and accuracy degradation in the long run. In a related context, a recent study \cite{wu2023fedab} proposes FedAB, an incentive mechanism for FL that promotes user engagement, effectiveness, fairness, and reciprocity. It leverages a combination of multi-attribute reverse auction and combinatorial MAB strategies. Notably, FedAB incorporates Upper Confidence Bound (UCB)-based client selection, balancing the exploitation of past reliable clients with the exploration of promising new clients.
    
\end{itemize}
\addtocounter{table}{-1}

\setlength\rotFPtop{0pt plus 1fil} 

\begin{table} [htbp]
\tiny
\caption{Summary of Federated Learning Aggregation Methods for Tackling Communication Efficiency - Client Participation.}
\label{tab-summary-FL-aggreg-client-participation}
\begin{threeparttable}

\begin{tabularx}{\linewidth}{|X|XXX|XXXXXXXXXXX|}
\hline
\multicolumn{1}{|c|}{\multirow{3}{*}{\textbf{Related Work}}} &
  \multicolumn{3}{c|}{\multirow{2}{*}{\textbf{Environment}}} &
  \multicolumn{11}{c|}{\textbf{\textbf{Verified Goals}}} \\ \cline{5-15} 
 &
  \multicolumn{3}{c|}{} &
  \multicolumn{3}{c|}{\textbf{Process}} &
  \multicolumn{5}{c|}{\textbf{Model}} &
  \multicolumn{3}{c|}{\textbf{System}} \\ \cline{2-15} 
 &
  \multicolumn{1}{c|}{\textbf{Synch}} & 
  \multicolumn{1}{c|}{\textbf{Users}} & 
  \multicolumn{1}{c|}{\textbf{Archi}} & 
  
  \multicolumn{1}{c|}{\textbf{Cm. E.}} & 
  \multicolumn{1}{c|}{\textbf{Cp. E.}} & 
  \multicolumn{1}{c|}{\textbf{Conv}} & 
  \multicolumn{1}{c|}{\textbf{Pers}} & 
  \multicolumn{1}{c|}{\textbf{Gen}} & 
  \multicolumn{1}{c|}{\textbf{Reg}} & 
  \multicolumn{1}{c|}{\textbf{Fair}} & 
  \multicolumn{1}{c|}{\textbf{Heter}} & 
  \multicolumn{1}{c|}{\textbf{Sec}} & 
  \multicolumn{1}{c|}{\textbf{Priv}} & 
  \multicolumn{1}{c|}{\textbf{Scal}} \\  
  \hline

%
  \multicolumn{1}{|c|}{\textbf{Used Techniques}} &
  \multicolumn{14}{c|}{\textbf{Client Selection}}  \\ \hline
  \multicolumn{1}{|c|}{\textbf{Reliability}} &
  \multicolumn{14}{c|}{}  \\ \hline

  \multicolumn{1}{|c|}{\cite{rjoub2022trust}} & 
  \multicolumn{1}{c|}{Synch} &
  \multicolumn{1}{c|}{50} &
  \multicolumn{1}{c|}{Hier} &
  \multicolumn{1}{c|}{--} &
  \multicolumn{1}{c|}{\checkmark} &
  \multicolumn{1}{c|}{\checkmark} &
  \multicolumn{1}{c|}{\checkmark} &
  \multicolumn{1}{c|}{--} &
  \multicolumn{1}{c|}{--} &
  \multicolumn{1}{c|}{--} &
  \multicolumn{1}{c|}{\checkmark} &
  \multicolumn{1}{c|}{--} &
  \multicolumn{1}{c|}{--} &
  \multicolumn{1}{c|}{--}  \\ 
  \hline

 \multicolumn{1}{|c|}{Scoring FL \cite{zhang2023scoring}} &
  \multicolumn{1}{c|}{Synch} &
  \multicolumn{1}{c|}{8} &
  \multicolumn{1}{c|}{Cent} &
  \multicolumn{1}{c|}{\checkmark} &
  \multicolumn{1}{c|}{--} &
  \multicolumn{1}{c|}{\checkmark} &
  \multicolumn{1}{c|}{--} &
  \multicolumn{1}{c|}{--} &
  \multicolumn{1}{c|}{--} &
  \multicolumn{1}{c|}{--} &
  \multicolumn{1}{c|}{\checkmark} &
  \multicolumn{1}{c|}{--} &
  \multicolumn{1}{c|}{\checkmark} &
  \multicolumn{1}{c|}{--}  \\ 
  \hline
  \multicolumn{1}{|c|}{\textbf{Utility}} &
  \multicolumn{14}{c|}{}  \\ \hline

   \multicolumn{1}{|c|}{CU-CS \cite{shi2023efficient}} &
  \multicolumn{1}{c|}{Synch} &
  \multicolumn{1}{c|}{20} &
  \multicolumn{1}{c|}{Cent} &
  \multicolumn{1}{c|}{\checkmark} &
  \multicolumn{1}{c|}{\checkmark} &
  \multicolumn{1}{c|}{\checkmark} &
  \multicolumn{1}{c|}{--} &
  \multicolumn{1}{c|}{\checkmark} &
  \multicolumn{1}{c|}{--} &
  \multicolumn{1}{c|}{--} &
  \multicolumn{1}{c|}{--} &
  \multicolumn{1}{c|}{--} &
  \multicolumn{1}{c|}{--} &
  \multicolumn{1}{c|}{--} \\ 
  \hline

 \multicolumn{1}{|c|}{ACFL\cite{huang2023active}} & 
  \multicolumn{1}{c|}{Synch} &
  \multicolumn{1}{c|}{140} &
  \multicolumn{1}{c|}{Hier} &
  \multicolumn{1}{c|}{\checkmark} &
  \multicolumn{1}{c|}{\checkmark} &
  \multicolumn{1}{c|}{\checkmark} &
  \multicolumn{1}{c|}{\checkmark} &
  \multicolumn{1}{c|}{\checkmark} &
  \multicolumn{1}{c|}{--} &
  \multicolumn{1}{c|}{--} &
  \multicolumn{1}{c|}{--} &
  \multicolumn{1}{c|}{--} &
  \multicolumn{1}{c|}{--} &
  \multicolumn{1}{c|}{--}  \\ 
  \hline

    \multicolumn{1}{|c|}{\textbf{}} &
  \multicolumn{14}{c|}{\textbf{Client Maximization}}  \\ \hline

  \multicolumn{1}{|c|}{ITMCS \cite{ji2022client}} & 
  \multicolumn{1}{c|}{Synch} &
  \multicolumn{1}{c|}{40} &
  \multicolumn{1}{c|}{Cent} &
  \multicolumn{1}{c|}{--} &
  \multicolumn{1}{c|}{\checkmark} &
  \multicolumn{1}{c|}{\checkmark} &
  \multicolumn{1}{c|}{\checkmark} &
  \multicolumn{1}{c|}{--} &
  \multicolumn{1}{c|}{--} &
  \multicolumn{1}{c|}{--} &
  \multicolumn{1}{c|}{\checkmark} &
  \multicolumn{1}{c|}{--} &
  \multicolumn{1}{c|}{--} &
  \multicolumn{1}{c|}{--}  \\ 
  \hline

 \multicolumn{1}{|c|}{FedAB \cite{wu2023fedab}} &
  \multicolumn{1}{c|}{Synch} &
  \multicolumn{1}{c|}{20} &
  \multicolumn{1}{c|}{Cent} &
  
  \multicolumn{1}{c|}{\checkmark} &
  \multicolumn{1}{c|}{\checkmark} &
  \multicolumn{1}{c|}{\checkmark} &
  \multicolumn{1}{c|}{--} &
  \multicolumn{1}{c|}{\checkmark} &
  \multicolumn{1}{c|}{--} &
  \multicolumn{1}{c|}{\checkmark} &
  \multicolumn{1}{c|}{\checkmark} &
  \multicolumn{1}{c|}{--} &
  \multicolumn{1}{c|}{--} &
  \multicolumn{1}{c|}{--}  \\ 
  \hline
\end{tabularx}

\begin{tablenotes}
      \footnotesize \item \textit{Synch}: Synchronization mode \{Synchronous, Asynchronous\}. \textit{Users}: Maximum active users used in experiments. \textit{Archi}: Architecture  \{Centralized, Decentralized(P2P), Hierarchical\}. \textit{Cm. E.}: Communication Efficiency. \textit{Cp. E.}: Computation Efficiency.\textit{ Conv}: Convergence Analysis. \textit{Pers}: Personalization. \textit{Gen}: Generalization. \textit{Fair}: Fairness. \textit{Heter}: Heterogeneity. \textit{Sec}: Security. \textit{Priv}: Privacy. \textit{Scal}: Scalability.
\end{tablenotes}
\end{threeparttable}

\end{table}

\subsection{\textbf{Strategies for Security and Privacy Concerns}}
\label{FL_agg_solution_sec_priv}
\begin{figure}[htbp]
    \centering
    \includegraphics[scale=0.4]{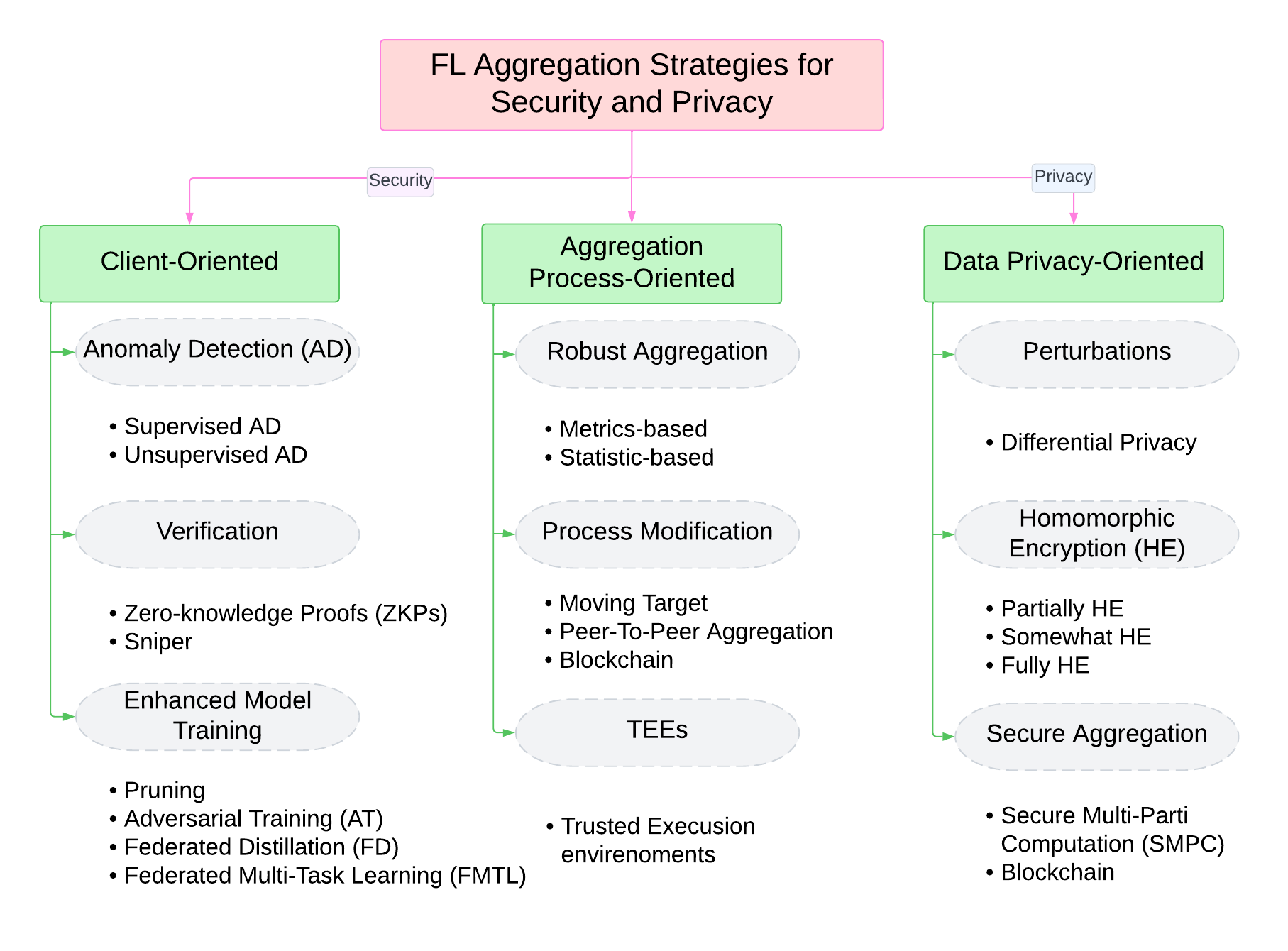}
    \caption{Classification of Federated Learning Aggregation Strategies for Security and Privacy.}
    \label{FL-security-privacy}
\end{figure}

Given the significant risks associated with FL security and privacy attack surfaces, as discussed in Section \ref{security_privacy_in_FL}, it is crucial to survey the defense mechanisms to attract attention to this line of interest and inspire further research. In this section, we will furnish an exhaustive breakdown of various detection and defense measures, particularly emphasizing those that directly impact FL aggregation. Although attack detection and defense are two different phases, we treat them as the same in this survey since they usually work together to protect FL. Our proposed taxonomy for FL security and privacy mechanisms synthesizes multiple perspectives for a comprehensive view, as illustrated in Fig. \ref{FL-security-privacy}. 

\subsubsection{Client-Oriented}
    
    A straightforward and intuitive solution to defend against aggregation attacks is to analyze the components involved in FL aggregation, such as the clients' local updates and training rules. The central idea behind this class of solutions is that, even without direct access to the client's data or complete control over their behaviors, for privacy reasons, the central aggregator in an FL system can identify abnormalities and mitigate them by implementing appropriate evaluation mechanisms for reliability verification. Table \ref{tab-summary-FL-aggreg-security-client} outlines the proposed taxonomy of security and privacy solutions in FL, highlighting relevant contemporary research.

\begin{itemize}
    \item \textit{Anomaly Detection.} 
    
    Anomaly detection mechanisms play an active role in identifying and mitigating the impact of suspicious behavior from individual users, thereby safeguarding the FL aggregation process. The main perspectives of anomaly detection are twofold. First, it involves detecting outliers that significantly deviate from expected behavior or exhibit patterns indicative of potential attacks or data anomalies. Second, it considers monitoring the communication and network activities of clients. For example, unusual communication patterns, excessive data transfer, or suspicious network behavior can serve as signals for detecting malicious activity or unauthorized access attempts. For a comprehensive overview, we categorize proactive anomaly detectors in to two families: 
    
    \textit{\textbf{Supervised Anomaly Detection.}}
    Supervised anomaly detection techniques require labeled data during the classifier training. These methods utilize historical data that encompasses both normal and deviant instances, enabling the classifier to learn and differentiate between them. \textit{Spectral-based methods} are widely employed in anomaly detection tasks. These techniques embed both benign and malicious data into a lower-dimensional space using spectral analysis. For instance, both studies in \cite{gu2022fedcut} and \cite{zeng2022never} leverage spectral analysis as a means to detect and alleviate Byzantine misbehavior and backdoors.
    Likewise, the \textit{entropy-based methods} offer insights into the unusual patterns within the data. However, it relies on entropy, a measure of uncertainty or randomness in the data. By calculating the entropy of different features or data distributions, these methods can detect instances with abnormally high or low entropy values. The Sageflow \cite{park2021sageflow} and ELITE \cite{wang2021elite} frameworks compare the entropy of clients' local updates to a threshold value, excluding potential outliers from aggregation. This process prevents harmful updates from impacting the final global model.
 
    \textit{\textbf{Unsupervised Anomaly Detection.}}
    In contrast to the previous detectors, unsupervised anomaly detection techniques perform independently of labeled data. They aim to identify anomalies by learning the underlying structure or patterns in the data that deviate significantly from expected behavior. These methods assume that anomalous data are infrequent and distinct from benign ones.  
    For instance, DL-based anomaly detection often operates using \textit{autoencoders}, a specialized type of neural network architecture. An autoencoder comprises two main components: an encoder and a decoder. In the encoding phase, the encoder compresses input data into a lower-dimensional representation, aiming to capture essential features of the data. Subsequently, the decoder reconstructs the original input from this compressed format, striving for minimal error in the reconstruction process. This process facilitates anomaly identification by discerning variations between the reconstructed data and the original data. The data instance is then classified as an anomaly if the computed error surpasses a predetermined threshold. Notably, autoencoders have proven their superiority over traditional linear algebra-based methods like Principal Component Analysis (PCA) when handling intricate and nonlinear data problems.
    Extensive research has been conducted in this area \cite{liu2022privacy, vucovich2022anomaly}. For instance, the work in \cite{liu2022privacy}, introduces an autoencoder to estimate the anomaly score before each aggregation round. 
    Another commonly employed strategy in this class involves the use of \textit{DNN-based detectors} \cite{wang2023federated}. The binary implementation of this technique operates in two steps. First, the DNN is trained on normal training data to recognize the patterns associated with expected behavior. Then, during the testing phase, each new instance is fed into the trained DNN. If the DNN accepts the input, it is considered non-anomalous, whereas if the DNN network rejects the instance, it is identified as an anomaly. The proposal in \cite{ma2021federated} exemplifies this approach. Similarly, Through DNN utilization in privacy-preserving recommender systems \cite{wang2022fast}, the proposed PrivRec and DP-PrivRec models attain intricate user and item representations, facilitating swift adaptation to new users while safeguarding privacy.

    \item \textit{Verification.}
     
    \textit{\textbf{Zero-knowledge Proofs.}} 
    ZKPs offer a powerful tool for verifiability on private data. In essence, they enable a "prover" to demonstrate knowledge of specific information without disclosing the underlying details to a "verifier." In federated learning, probabilistic ZKP assessments enable participants to prove that their submitted model updates adhere to predefined criteria, such as specified ranges and authentic features. This validation ensures the accuracy and correctness of their computations while yielding their sensitive data preserved \cite{yang2023fedzkp, xing2023zero}.

    \textit{\textbf{Sniper.}} 
    To address the problem of distributed poisoning attacks, where multiple attackers collude to inject malicious samples into the training data, potentially causing significant harm to FL, the work in \cite{cao2019understanding} presents the Sniper paradigm. The Sniper scheme uses a two-phase verification approach based on a validation dataset to detect and eliminate potentially poisoned models. The authors advocate that their proposal dramatically decreases the success rate of poisoning attacks even when more than one attacker is involved.

    \item \textit{Enhanced Model Training.}
    
    The defense paradigms that fall into this class strive to improve the training of FL models using innovative methods that surpass conventional methodologies and surmount unresolved limitations. Researchers have actively sought solutions from diverse disciplines, adapting them to the unique challenges of FL. 

     \textit{\textbf{Pruning.}}
    The pruning technique reduces the size of neural networks by eliminating unnecessary connections or parameters without sacrificing performance. Parameters are evaluated based on their impact on the network's accuracy, and those deemed less crucial or present negligible contributions are pruned. This results in a more lightweight and efficient learned model. Pruning is a suitable technique in FL, where users often have limited resources incapable of training large neural networks. To offer a global overview of this method, the authors in \cite{wei2023securing} describe two simple pruning approaches: \textit{Threshold-Based Pruning} and \textit{Random Pruning}. For enhanced model training, the authors in \cite{wei2023securing} describe two simple pruning approaches: \textit{Threshold-Based Pruning} and \textit{Random Pruning}. In a similar vein, the Large Gradient Pruning (LGP) method \cite{zhang2023preserving} involves setting a threshold for gradient magnitudes and removing gradients below this
    threshold. By doing so, the LGP framework retains informative updates while reducing the risk of privacy breaches through gradient inversion attacks. Similar work can be found in \cite{zhao2022cork}\cite{stripelis2022towards} to defend against both security and privacy breaches. 
       
    \textit{\textbf{Adversarial Training (AT).}}
    Adversarial training incorporates specially manipulated adversarial examples alongside the regular data during the model training process. The goal is to expose the model to various attacks during training, teaching it to be more cautious and resilient when dealing with such attacks in real-world scenarios. For example, in \cite{hallaji2023label}, this training policy is applied to strengthen the robustness of neural network models against label poisoning attacks. Similarly, \cite{zhang2023secure} presents a novel vertical federated learning approach explicitly designed to thwart label inference attacks as one of the significant privacy threats in vertical FL systems.
    
    \textit{\textbf{Federated Distillation (FD).}}
    As opposed to standard FL exchanges of model parameters, Federated Distillation (FD) \cite{seo202216} only requires the transmission of model outputs, which are significantly smaller in size. The concept of FD draws inspiration from Knowledge Distillation (KD), which focuses on transferring knowledge from a fully trained, large, and complex model (teacher) to a smaller and simpler model (student). In communication resource-constrained settings, federated distillation emerges as a highly appealing solution for sharing knowledge among participants.
    Previous proposals in this regard are presented in FedMD \cite{li2019fedmd}, FedKD \cite{seo202216}, and FedGEN \cite{zhu2021data}. However, recent investigations have revealed that solely relying on gradient hiding leaves the FL system prone to threats like single-point-of-failure and membership information leakage \cite{yang2023fd}. In light of this, researchers have sought to combine knowledge distillation with emerging solutions like blockchain \cite{li2023hbmd} and edge computing \cite{yang2023edge}. One notable work \cite{shao2023selective} takes FD to the next level through a selective knowledge-sharing mechanism that is able to handle heterogeneous data. Selective-FD employs client-side and server-side selectors to identify and filter misleading or unreliable contributions during knowledge sharing.
    
    \textit{\textbf{Federated Multi-Task Learning (FMTL).}}
    In multi-task learning (MTL), the objective is to train a single model that can effectively handle multiple related tasks simultaneously, rather than building separate models for each task. Federated multi-task learning combines the benefits of MLT with the advantages of FL to address the challenges of learning models for numerous related activities on non-iid data. For instance, the authors in \cite{yi2022hfedmtl} designed HFedMTL, an FMTL system using a primal-dual method for task reduction, enhancing MTL flexibility on massive terminals. Another FMTL scheduling mechanism is developed in \cite{liu2023federated} harnesses a trusted computing sandbox within the blockchain framework. 
\end{itemize} 
\addtocounter{table}{-1}

\setlength\rotFPtop{0pt plus 1fil} 

\begin{table} [htbp]
\tiny
\caption{Summary of Federated Learning Aggregation Methods for Tackling Security - Client-Oriented.}
\label{tab-summary-FL-aggreg-security-client}
\begin{threeparttable}
\begin{tabularx}{\linewidth}{|X|XXX|XXXXXXXXXXX|}
\hline
\multicolumn{1}{|c|}{\multirow{3}{*}{\textbf{Related Work}}} &
  \multicolumn{3}{c|}{\multirow{2}{*}{\textbf{Environment}}} &
  \multicolumn{11}{c|}{\textbf{\textbf{Verified Goals}}} \\ \cline{5-15} 
 &
  \multicolumn{3}{c|}{} &
  \multicolumn{3}{c|}{\textbf{Process}} &
  \multicolumn{5}{c|}{\textbf{Model}} &
  \multicolumn{3}{c|}{\textbf{System}} \\ \cline{2-15} 
 &
  \multicolumn{1}{c|}{\textbf{Synch}} & 
  \multicolumn{1}{c|}{\textbf{Users}} & 
  \multicolumn{1}{c|}{\textbf{Archi}} & 
  
  \multicolumn{1}{c|}{\textbf{Cm. E.}} & 
  \multicolumn{1}{c|}{\textbf{Cp. E.}} & 
  \multicolumn{1}{c|}{\textbf{Conv}} & 
  \multicolumn{1}{c|}{\textbf{Pers}} & 
  \multicolumn{1}{c|}{\textbf{Gen}} & 
  \multicolumn{1}{c|}{\textbf{Reg}} & 
  \multicolumn{1}{c|}{\textbf{Fair}} & 
  \multicolumn{1}{c|}{\textbf{Heter}} & 
  \multicolumn{1}{c|}{\textbf{Sec}} & 
  \multicolumn{1}{c|}{\textbf{Priv}} & 
  \multicolumn{1}{c|}{\textbf{Scal}} \\  
  \hline

    
  \multicolumn{1}{|c|}{\textbf{}} &
  \multicolumn{14}{c|}{\textbf{Anomaly Detection}} \\ \hline
  \multicolumn{1}{|c|}{\textbf{Supervised AD}} &
  \multicolumn{14}{c|}{} \\ \hline

  \multicolumn{1}{|c|}{FedCut \cite{gu2022fedcut}} & 
  \multicolumn{1}{c|}{Asynch} &
  \multicolumn{1}{c|}{100} &
  \multicolumn{1}{c|}{Cent} &
  \multicolumn{1}{c|}{--} &
  \multicolumn{1}{c|}{\checkmark} &
  \multicolumn{1}{c|}{--} &
  \multicolumn{1}{c|}{--} &
  \multicolumn{1}{c|}{--} &
  \multicolumn{1}{c|}{--} &
  \multicolumn{1}{c|}{--} &
  \multicolumn{1}{c|}{--} &
  \multicolumn{1}{c|}{\checkmark} &
  \multicolumn{1}{c|}{--} &
  \multicolumn{1}{c|}{\checkmark} \\ 
  \hline

  \multicolumn{1}{|c|}{Sageflow \cite{park2021sageflow} } & 
  \multicolumn{1}{c|}{Asynch} &
  \multicolumn{1}{c|}{100} &
  \multicolumn{1}{c|}{Cent} &
  \multicolumn{1}{c|}{\checkmark} &
  \multicolumn{1}{c|}{\checkmark} &
  \multicolumn{1}{c|}{\checkmark} &
  \multicolumn{1}{c|}{--} &
  \multicolumn{1}{c|}{--} &
  \multicolumn{1}{c|}{--} &
  \multicolumn{1}{c|}{--} &
  \multicolumn{1}{c|}{\checkmark} &
  \multicolumn{1}{c|}{\checkmark} &
  \multicolumn{1}{c|}{--} &
  \multicolumn{1}{c|}{--} \\ 
  \hline
  \multicolumn{1}{|c|}{\textbf{Unsupervised AD}} &
  \multicolumn{14}{c|}{} \\ \hline
  
  \multicolumn{1}{|c|}{FL-RAEC \cite{liu2022privacy}} & 
  \multicolumn{1}{c|}{Synch} &
  \multicolumn{1}{c|}{10} &
  \multicolumn{1}{c|}{Cent} &
  \multicolumn{1}{c|}{\checkmark} &
  \multicolumn{1}{c|}{\checkmark} &
  \multicolumn{1}{c|}{--} &
  \multicolumn{1}{c|}{--} &
  \multicolumn{1}{c|}{--} &
  \multicolumn{1}{c|}{--} &
  \multicolumn{1}{c|}{--} &
  \multicolumn{1}{c|}{--} &
  \multicolumn{1}{c|}{\checkmark} &
  \multicolumn{1}{c|}{\checkmark} &
  \multicolumn{1}{c|}{--} \\ 
  \hline

  \multicolumn{1}{|c|}{DeepSA \cite{ma2021federated}} & 
  \multicolumn{1}{c|}{Synch} &
  \multicolumn{1}{c|}{50} &
  \multicolumn{1}{c|}{Cent} &
  \multicolumn{1}{c|}{--} &
  \multicolumn{1}{c|}{\checkmark} &
  \multicolumn{1}{c|}{\checkmark} &
  \multicolumn{1}{c|}{--} &
  \multicolumn{1}{c|}{--} &
  \multicolumn{1}{c|}{--} &
  \multicolumn{1}{c|}{--} &
  \multicolumn{1}{c|}{--} &
  \multicolumn{1}{c|}{\checkmark} &
  \multicolumn{1}{c|}{--} &
  \multicolumn{1}{c|}{--} \\ 
  \hline
  
  \multicolumn{1}{|c|}{\textbf{}} &
  \multicolumn{14}{c|}{\textbf{Verification}} \\ \hline
  \multicolumn{1}{|c|}{\textbf{ZKPs}} &
  \multicolumn{14}{c|}{} \\ \hline
  
  \multicolumn{1}{|c|}{ \cite{xing2023zero}} & 
  \multicolumn{1}{c|}{Synch} &
  \multicolumn{1}{c|}{--} &
  \multicolumn{1}{c|}{Cent} &
  \multicolumn{1}{c|}{--} &
  \multicolumn{1}{c|}{\checkmark} &
  \multicolumn{1}{c|}{\checkmark} &
  \multicolumn{1}{c|}{--} &
  \multicolumn{1}{c|}{--} &
  \multicolumn{1}{c|}{--} &
  \multicolumn{1}{c|}{--} &
  \multicolumn{1}{c|}{--} &
  \multicolumn{1}{c|}{\checkmark} &
  \multicolumn{1}{c|}{\checkmark} &
  \multicolumn{1}{c|}{--} \\ 
  \hline
  \multicolumn{1}{|c|}{\textbf{Sniper}} &
  \multicolumn{14}{c|}{} \\ \hline
  
  \multicolumn{1}{|c|}{ \cite{cao2019understanding}} &
  \multicolumn{1}{c|}{Synch} &
  \multicolumn{1}{c|}{10} &
  \multicolumn{1}{c|}{Cent} &
  \multicolumn{1}{c|}{--} &
  \multicolumn{1}{c|}{--} &
  \multicolumn{1}{c|}{--} &
  \multicolumn{1}{c|}{--} &
  \multicolumn{1}{c|}{--} &
  \multicolumn{1}{c|}{--} &
  \multicolumn{1}{c|}{--} &
  \multicolumn{1}{c|}{--} &
  \multicolumn{1}{c|}{\checkmark} &
  \multicolumn{1}{c|}{--} &
  \multicolumn{1}{c|}{--} \\ 
  \hline

  \multicolumn{1}{|c|}{\textbf{}} &
  \multicolumn{14}{c|}{\textbf{Enhanced Model Training}} \\ \hline
  \multicolumn{1}{|c|}{\textbf{Pruning}} &
  \multicolumn{14}{c|}{} \\ \hline

  \multicolumn{1}{|c|}{CORK \cite{zhao2022cork}} & 
  \multicolumn{1}{c|}{Synch} &
  \multicolumn{1}{c|}{100} &
  \multicolumn{1}{c|}{Cent} &
  \multicolumn{1}{c|}{\checkmark} &
  \multicolumn{1}{c|}{\checkmark} &
  \multicolumn{1}{c|}{\checkmark} &
  \multicolumn{1}{c|}{--} &
  \multicolumn{1}{c|}{--} &
  \multicolumn{1}{c|}{--} &
  \multicolumn{1}{c|}{--} &
  \multicolumn{1}{c|}{\checkmark} &
  \multicolumn{1}{c|}{\checkmark} &
  \multicolumn{1}{c|}{\checkmark} &
  \multicolumn{1}{c|}{--} \\ 
  \hline
  \multicolumn{1}{|c|}{ \cite{zhang2023preserving}} &
  \multicolumn{1}{c|}{Synch} &
  \multicolumn{1}{c|}{10} &
  \multicolumn{1}{c|}{Cent} &
  \multicolumn{1}{c|}{\checkmark} &
  \multicolumn{1}{c|}{--} &
  \multicolumn{1}{c|}{--} &
  \multicolumn{1}{c|}{--} &
  \multicolumn{1}{c|}{--} &
  \multicolumn{1}{c|}{\checkmark} &
  \multicolumn{1}{c|}{--} &
  \multicolumn{1}{c|}{--} &
  \multicolumn{1}{c|}{--} &
  \multicolumn{1}{c|}{\checkmark} &
  \multicolumn{1}{c|}{--} \\ 
  \hline

  \multicolumn{1}{|c|}{\textbf{AT}} &
  \multicolumn{14}{c|}{} \\ \hline
  
  \multicolumn{1}{|c|}{RobustFL \cite{zhang2021robustfl}} & 
  \multicolumn{1}{c|}{Synch} &
  \multicolumn{1}{c|}{100} &
  \multicolumn{1}{c|}{Cent} &
  \multicolumn{1}{c|}{--} &
  \multicolumn{1}{c|}{--} &
  \multicolumn{1}{c|}{--} &
  \multicolumn{1}{c|}{--} &
  \multicolumn{1}{c|}{--} &
  \multicolumn{1}{c|}{--} &
  \multicolumn{1}{c|}{--} &
  \multicolumn{1}{c|}{--} &
  \multicolumn{1}{c|}{\checkmark} &
  \multicolumn{1}{c|}{\checkmark} &
  \multicolumn{1}{c|}{--} \\ 
  \hline
  \multicolumn{1}{|c|}{ \cite{hallaji2023label}} & 
  \multicolumn{1}{c|}{Synch} &
  \multicolumn{1}{c|}{50} &
  \multicolumn{1}{c|}{Cent} &
  \multicolumn{1}{c|}{--} &
  \multicolumn{1}{c|}{--} &
  \multicolumn{1}{c|}{\checkmark} &
  \multicolumn{1}{c|}{--} &
  \multicolumn{1}{c|}{--} &
  \multicolumn{1}{c|}{--} &
  \multicolumn{1}{c|}{--} &
  \multicolumn{1}{c|}{--} &
  \multicolumn{1}{c|}{\checkmark} &
  \multicolumn{1}{c|}{--} &
  \multicolumn{1}{c|}{--} \\ 
  \hline

  \multicolumn{1}{|c|}{\textbf{FD}} &
  \multicolumn{14}{c|}{} \\ \hline
  
  \multicolumn{1}{|c|}{Selective-FD \cite{shao2023selective}} & 
  \multicolumn{1}{c|}{Synch} &
  \multicolumn{1}{c|}{4} &
  \multicolumn{1}{c|}{Cent} &
  \multicolumn{1}{c|}{\checkmark} &
  \multicolumn{1}{c|}{--} &
  \multicolumn{1}{c|}{\checkmark} &
  \multicolumn{1}{c|}{--} &
  \multicolumn{1}{c|}{--} &
  \multicolumn{1}{c|}{--} &
  \multicolumn{1}{c|}{--} &
  \multicolumn{1}{c|}{\checkmark} &
  \multicolumn{1}{c|}{--} &
  \multicolumn{1}{c|}{\checkmark} &
  \multicolumn{1}{c|}{--} \\ 
  \hline

  \multicolumn{1}{|c|}{HBMD-FL \cite{li2023hbmd}} & 
  \multicolumn{1}{c|}{Synch} &
  \multicolumn{1}{c|}{10} &
  \multicolumn{1}{c|}{Decent} &
  \multicolumn{1}{c|}{\checkmark} &
  \multicolumn{1}{c|}{--} &
  \multicolumn{1}{c|}{--} &
  \multicolumn{1}{c|}{--} &
  \multicolumn{1}{c|}{--} &
  \multicolumn{1}{c|}{--} &
  \multicolumn{1}{c|}{--} &
  \multicolumn{1}{c|}{\checkmark} &
  \multicolumn{1}{c|}{\checkmark} &
  \multicolumn{1}{c|}{\checkmark} &
  \multicolumn{1}{c|}{--} \\ 
  \hline
  
  \multicolumn{1}{|c|}{\textbf{FMTL}} &
  \multicolumn{14}{c|}{} \\ \hline

  \multicolumn{1}{|c|}{ \cite{liu2023federated}} & 
  \multicolumn{1}{c|}{Synch} &
  \multicolumn{1}{c|}{10} &
  \multicolumn{1}{c|}{Decent} &
  \multicolumn{1}{c|}{--} &
  \multicolumn{1}{c|}{\checkmark} &
  \multicolumn{1}{c|}{\checkmark} &
  \multicolumn{1}{c|}{--} &
  \multicolumn{1}{c|}{--} &
  \multicolumn{1}{c|}{--} &
  \multicolumn{1}{c|}{--} &
  \multicolumn{1}{c|}{\checkmark} &
  \multicolumn{1}{c|}{\checkmark} &
  \multicolumn{1}{c|}{--} &
  \multicolumn{1}{c|}{--} \\ 
  \hline
  
  \multicolumn{1}{|c|}{HFedMTL \cite{yi2022hfedmtl}} & 
  \multicolumn{1}{c|}{Synch} &
  \multicolumn{1}{c|}{10} &
  \multicolumn{1}{c|}{Hier} &
  \multicolumn{1}{c|}{\checkmark} &
  \multicolumn{1}{c|}{--} &
  \multicolumn{1}{c|}{\checkmark} &
  \multicolumn{1}{c|}{--} &
  \multicolumn{1}{c|}{--} &
  \multicolumn{1}{c|}{--} &
  \multicolumn{1}{c|}{\checkmark} &
  \multicolumn{1}{c|}{--} &
  \multicolumn{1}{c|}{--} &
  \multicolumn{1}{c|}{--} &
  \multicolumn{1}{c|}{--} \\ 
  \hline
\end{tabularx}

\begin{tablenotes}
      \footnotesize \item \textit{Synch}: Synchronization mode \{Synchronous, Asynchronous\}. \textit{Users}: Maximum active users used in experiments. \textit{Archi}: Architecture  \{Centralized, Decentralized(P2P), Hierarchical\}. \textit{Cm. E.}: Communication Efficiency. \textit{Cp. E.}: Computation Efficiency.\textit{ Conv}: Convergence Analysis. \textit{Pers}: Personalization. \textit{Gen}: Generalization. \textit{Fair}: Fairness. \textit{Heter}: Heterogeneity. \textit{Sec}: Security. \textit{Priv}: Privacy. \textit{Scal}: Scalability. \textit{AT}: Adversarial Training. \textit{FD}: Federated Distillation. \textit{FMTL}: Federated Multi-Task Learning
\end{tablenotes}
\end{threeparttable}
\end{table}

\subsubsection{Aggregation Process-Oriented} 
    Securing the FL aggregation procedure is of utmost importance. The proposals in this category aim for a resilient pipeline against communication issues, client dropouts, and malicious actors. Upon our literature review, we have organized these efforts into three primary categories: robust and secure aggregation, a modified version of the standard aggregation process, and a well-protected execution environment. These categories as well as their sub-categories are summarized in Table \ref{tab-summary-FL-aggreg-security-agg} to form a selection of relevant work in the literature.

\begin{figure}[htbp]
    \centering
    \includegraphics[width=0.5\textwidth]{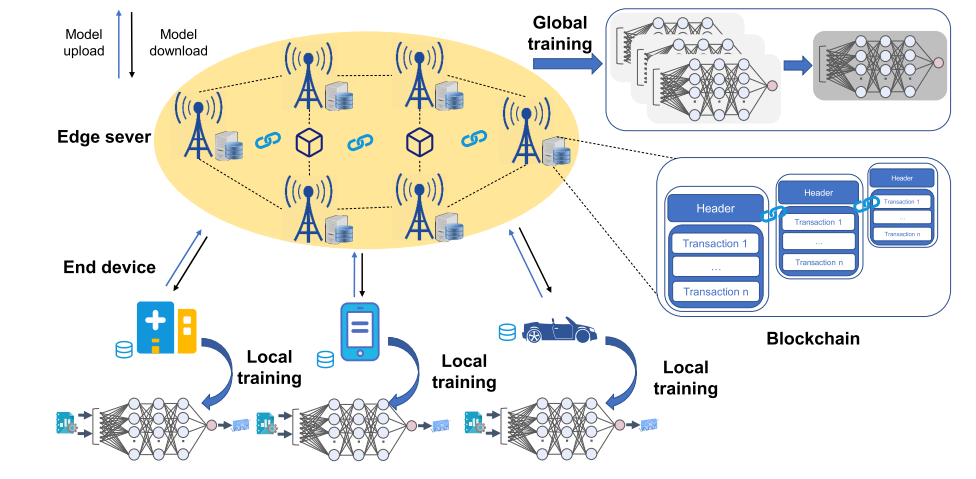}
    \caption{Securing Cloud-Edge Federated Learning System with Blockchain \cite{mao2023security}.}
    \label{FL-with-bloackchain}
\end{figure}

    \begin{itemize}
        \item \textit{Robust Aggregation.}
        
        Robust aggregation remediates the FL system against security and privacy attacks by detecting and mitigating malicious or inaccurate client models. These practices operate under the assumption that poisoned and benign models exhibit distinguishable features. Their objective is to accomplish this task without jeopardizing performance or introducing communication bottlenecks. The literature explores two main approaches:
        
        \textit{\textbf{Metrics-based Aggregation Methods.}} The term "metrics" here refers to specific criteria used to evaluate the quality of local updates, such as trust, reliability, and similarity. The FL server scrutinizes individual models provided by each client $c$ and compare their performance metrics to a validation dataset using the aggregated model derived from all updates except for that of client $c$. Consequently, the FL server can flag anomalous updates that diminish the model's utility based on predefined rules or thresholds and potentially discard them. It is worth noting that here, the FL server requires access to a validation dataset, which may not always be feasible in FL. For instance, the FLTrust \cite{cao2020fltrust} exemplifies a trust-based mechanism. It treats client updates as vectors and employs ReLU cosine similarity to assess their alignment with the server model, potentially flagging dissimilar updates. Unlike prior cosine-based methods \cite{ma2022shieldfl}, the CosDefense algorithm \cite{yaldiz2023secure} utilizes cosine similarity scores and requires only the updates themselves without requiring additional information. However, the authors in \cite{mao2021romoa} argue that simply calculating distance or similarity is not suﬃcient, proposing hybrid similarity measures with a look-ahead strategy. A similar voting-based strategy is presented in \cite{yue2022federated}.

        \textit{\textbf{Statistic-based Aggregation Methods.}}
        An alternative (or a supplementary) approach to metrics-based verification methods consists of evaluating the statistical properties of update magnitudes such as geometric median \cite{pillutla2022robust}, trimmed mean \cite{yin2018byzantine}, Krum/multi-Krum \cite{blanchard2017machine}, Bulyan \cite{guerraoui2018hidden}, and FoolsGold \cite{fung2020limitations}. Such indicators filter only valid updates. Nonetheless, it's important to highlight that these straightforward statistical techniques have demonstrated their vulnerabilities \cite{zizzo2020fat}. Consequently, this has led researchers to explore sophisticated methodologies, such as blockchain \cite{liu2022federated}, clustering \cite{li2022enhancing}, and fair detection \cite{singh2023fair} to boost the FL security.

        \item \textit{Process Modification}

        \textit{\textbf{Moving Target.}}
            Moving Target Defense (MTD) mechanisms orchestrate perturbal adjustments continuously to the system components and parameters, making it more challenging and costly for potential attackers. These proactive measures offer robust protection against intrusions at the server, the network, and the application levels, effectively bolstering the system's resilience. Inspired by MTD \cite{zhou2021augmented}, researchers in \cite{zhou2021augmented} proposed ADS-MTD to achieve CIA security (confidentiality, integrity, and availability) in FL. ADS-MTD uses hierarchical dual shuffling: primary model shuffling and augmented client shuffling. The former anonymizes client contributions, making it difficult for attackers to identify which client contributed to which part of the final global model. While the later \textit{augmented shuffling} dynamically removes malicious clients to ensure model integrity and availability.

        \textit{\textbf{Peer-To-Peer Aggregation.}}
            Other research has emphasized the advantages of an alternative decentralized solution in federated learning, as it eliminates vulnerabilities to attacks targeting the central hub of the FL system. Therefore, there has been a growing interest in exploring peer-to-peer (or decentralized) aggregation. Nevertheless, it is crucial to consider the implications of such a decentralized aggregation scheme. As such, delegating the monitoring responsibility to individual clients leads to limiting global monitoring capabilities. 
            Casting light on traditional FL limitations in mobile robotics, a recent paper \cite{zhou2023decentralized} offers PPAFL, a P2P FL approach in 5G and beyond networks. This secure solution incorporates a Secret Sharing-based communication, a Secure Stochastic Gradient Descent scheme integrated with an Autoencoder, and a Gaussian mechanism to address data leakage. In the same spirit, the study \cite{lu2022privacy} presents a privacy-preserving consensus-based algorithm for decentralized  FL, in which learners share their local models exclusively with their one-hub peers. This area of research was extensively explored due to its potential for strengthening FL security and efficiency \cite{wang2023sparsfa, piotrowski2023towards, chen2022cfl}.

        \textit{\textbf{Blockchain.}}
            As defined by \cite{javed2022integration}, blockchain technology functions as a distributed and accessible database that acts as a verifiable and tamper-proof ledger. Its application in securing the aggregation process in federated learning yields numerous advantages. Firstly, it cultivates trust among clients and incentivizes the best contributors by rewarding their valuable involvement. Secondly, it reinforces the resilience of the aggregation process, minimizing the risk of vulnerabilities and potential failures, thanks to its remarkable properties of verifiability, traceability, and privacy preservation. Fig. \ref{FL-with-bloackchain} showcases a seamless integration of blockchain technology within an edge-based FL system \cite{mao2023security}. For example, The VFChain proposal \cite{peng2021vfchain} engages blockchain for enhanced security through verifiability and auditability. VFChain replaces central servers with a blockchain-selected verifier committee for aggregation. Also, it introduces an authenticated data structure for efficient verification and secure committee rotation. 
            Several recent studies have capitalized on this promising approach for various objectives and domains, including healthcare services \cite{samuel2022iomt}, Industrial Internet of Things (IIoT) networks \cite{yazdinejad2022block}.

        \item {Trusted Execution Environment (TEE).}
        
            TEEs refer to trusted and secure FL ecosystems that allow only authorized parties to perform attested operations and protect their communications. More specifically, TEE provides a cryptographic and isolated environment where authenticated code can run safely. Additionally, the TEE service is in charge of checking the authenticity and managing the access rights of the participant clients. Thus, the TEE ensures the learning process's integrity and confidentiality, preventing any tampering or manipulation. For instance, in \cite{mo2021ppfl}, the TEEs strategy plays a vital role at both the client and server levels. On the client side, the authors employ a robust greedy layer-wise training approach to keep sensitive information hidden from adversaries. Meanwhile, on the server side, TEEs enable secure aggregation using a cryptographic protocol. 
            The paper presented in \cite{chen2020training} also leverages the isolated enclave of TEEs to ensure the learning process integrity on both the client and server sides. In contrast, the work described in \cite{rieger2022close} focuses, specifically, on the client side and introduces CrowdGuard to combat targeted attacks. 
          
    \end{itemize}

\addtocounter{table}{-1}

\setlength\rotFPtop{0pt plus 1fil} 

\begin{table} [htbp]
\tiny
\caption{Summary of Federated Learning Aggregation Methods for Tackling Security - Aggregation Process-Oriented.}
\label{tab-summary-FL-aggreg-security-agg}
\begin{threeparttable}

\begin{tabularx}{\linewidth}{|X|XXX|XXXXXXXXXXX|}
\hline
\multicolumn{1}{|c|}{\multirow{3}{*}{\textbf{Related Work}}} &
  \multicolumn{3}{c|}{\multirow{2}{*}{\textbf{Environment}}} &
  \multicolumn{11}{c|}{\textbf{\textbf{Verified Goals}}} \\ \cline{5-15} 
 &
  \multicolumn{3}{c|}{} &
  \multicolumn{3}{c|}{\textbf{Process}} &
  \multicolumn{5}{c|}{\textbf{Model}} &
  \multicolumn{3}{c|}{\textbf{System}} \\ \cline{2-15} 
 &
  \multicolumn{1}{c|}{\textbf{Synch}} & 
  \multicolumn{1}{c|}{\textbf{Users}} & 
  \multicolumn{1}{c|}{\textbf{Archi}} & 
  
  \multicolumn{1}{c|}{\textbf{Cm. E.}} & 
  \multicolumn{1}{c|}{\textbf{Cp. E.}} & 
  \multicolumn{1}{c|}{\textbf{Conv}} & 
  \multicolumn{1}{c|}{\textbf{Pers}} & 
  \multicolumn{1}{c|}{\textbf{Gen}} & 
  \multicolumn{1}{c|}{\textbf{Reg}} & 
  \multicolumn{1}{c|}{\textbf{Fair}} & 
  \multicolumn{1}{c|}{\textbf{Heter}} & 
  \multicolumn{1}{c|}{\textbf{Sec}} & 
  \multicolumn{1}{c|}{\textbf{Priv}} & 
  \multicolumn{1}{c|}{\textbf{Scal}} \\  
  \hline

  \multicolumn{1}{|c|}{ \textbf{Used Techniques }} &
  \multicolumn{14}{c|}{\textbf{Robust Aggregation}} \\ \hline
  \multicolumn{1}{|c|}{\textbf{Metrics-based }} &
  \multicolumn{14}{c|}{} \\ \hline

  \multicolumn{1}{|c|}{Romoa \cite{mao2021romoa}} & 
  \multicolumn{1}{c|}{Asynch} &
  \multicolumn{1}{c|}{100} & 
  \multicolumn{1}{c|}{Cent} &
  \multicolumn{1}{c|}{--} &
  \multicolumn{1}{c|}{--} &
  \multicolumn{1}{c|}{--} &
  \multicolumn{1}{c|}{--} &
  \multicolumn{1}{c|}{--} &
  \multicolumn{1}{c|}{--} &
  \multicolumn{1}{c|}{--} &
  \multicolumn{1}{c|}{--} &
  \multicolumn{1}{c|}{\checkmark} &
  \multicolumn{1}{c|}{\checkmark} &
  \multicolumn{1}{c|}{\checkmark} \\ 
  \hline

  \multicolumn{1}{|c|}{FedVote \cite{yue2022federated}} & 
  \multicolumn{1}{c|}{Synch} &
  \multicolumn{1}{c|}{16} & 
  \multicolumn{1}{c|}{Cent} &
  \multicolumn{1}{c|}{\checkmark} &
  \multicolumn{1}{c|}{\checkmark} &
  \multicolumn{1}{c|}{\checkmark} &
  \multicolumn{1}{c|}{--} &
  \multicolumn{1}{c|}{--} &
  \multicolumn{1}{c|}{--} &
  \multicolumn{1}{c|}{--} &
  \multicolumn{1}{c|}{--} &
  \multicolumn{1}{c|}{\checkmark} &
  \multicolumn{1}{c|}{\checkmark} &
  \multicolumn{1}{c|}{--} \\ 
  \hline
  
  \multicolumn{1}{|c|}{\textbf{Statistic-based }} &
  \multicolumn{14}{c|}{} \\ \hline
  
  \multicolumn{1}{|c|}{RFA \cite{li2022robust}} & 
  \multicolumn{1}{c|}{Synch} &
  \multicolumn{1}{c|}{1000} &
  \multicolumn{1}{c|}{Cent} &
  \multicolumn{1}{c|}{\checkmark} &
  \multicolumn{1}{c|}{--} &
  \multicolumn{1}{c|}{\checkmark} &
  \multicolumn{1}{c|}{\checkmark} &
  \multicolumn{1}{c|}{--} &
  \multicolumn{1}{c|}{--} &
  \multicolumn{1}{c|}{--} &
  \multicolumn{1}{c|}{\checkmark} &
  \multicolumn{1}{c|}{\checkmark} &
  \multicolumn{1}{c|}{\checkmark} &
  \multicolumn{1}{c|}{\checkmark} \\ 
  \hline

  \multicolumn{1}{|c|}{ \cite{singh2023fair}} & 
  \multicolumn{1}{c|}{Synch} &
  \multicolumn{1}{c|}{50} &
  \multicolumn{1}{c|}{Hier} &
  \multicolumn{1}{c|}{--} &
  \multicolumn{1}{c|}{--} &
  \multicolumn{1}{c|}{--} &
  \multicolumn{1}{c|}{--} &
  \multicolumn{1}{c|}{\checkmark} &
  \multicolumn{1}{c|}{--} &
  \multicolumn{1}{c|}{\checkmark} &
  \multicolumn{1}{c|}{\checkmark} &
  \multicolumn{1}{c|}{\checkmark} &
  \multicolumn{1}{c|}{--} &
  \multicolumn{1}{c|}{--} \\ 
  \hline

  \multicolumn{1}{|c|}{\textbf{}} &
  \multicolumn{14}{c|}{\textbf{Process Modification}} \\ \hline
  \multicolumn{1}{|c|}{\textbf{Moving Target }} &
  \multicolumn{14}{c|}{} \\ \hline
  
  \multicolumn{1}{|c|}{LSSM \cite{zhou2020endogenous}} & 
  \multicolumn{1}{c|}{Synch} &
  \multicolumn{1}{c|}{100} & 
  \multicolumn{1}{c|}{Hier} &
  \multicolumn{1}{c|}{--} &
  \multicolumn{1}{c|}{--} &
  \multicolumn{1}{c|}{--} &
  \multicolumn{1}{c|}{--} &
  \multicolumn{1}{c|}{--} &
  \multicolumn{1}{c|}{--} &
  \multicolumn{1}{c|}{--} &
  \multicolumn{1}{c|}{\checkmark} &
  \multicolumn{1}{c|}{\checkmark} &
  \multicolumn{1}{c|}{--} &
  \multicolumn{1}{c|}{\checkmark} \\ 
  \hline

  \multicolumn{1}{|c|}{\textbf{Peer-To-Peer Agg}} & 
  \multicolumn{14}{c|}{} \\ \hline
  
  \multicolumn{1}{|c|}{ \cite{lu2022privacy}} &
  \multicolumn{1}{c|}{Synch} &
  \multicolumn{1}{c|}{100} &
  \multicolumn{1}{c|}{Decent} &
  \multicolumn{1}{c|}{--} &
  \multicolumn{1}{c|}{\checkmark} &
  \multicolumn{1}{c|}{\checkmark} &
  \multicolumn{1}{c|}{--} &
  \multicolumn{1}{c|}{--} &
  \multicolumn{1}{c|}{--} &
  \multicolumn{1}{c|}{--} &
  \multicolumn{1}{c|}{--} &
  \multicolumn{1}{c|}{--} &
  \multicolumn{1}{c|}{\checkmark} &
  \multicolumn{1}{c|}{\checkmark} \\ 
  \hline

  \multicolumn{1}{|c|}{CFL \cite{chen2022cfl}} & 
  \multicolumn{1}{c|}{Synch} &
  \multicolumn{1}{c|}{200} &
  \multicolumn{1}{c|}{Dec-Hier} &
  \multicolumn{1}{c|}{\checkmark} &
  \multicolumn{1}{c|}{\checkmark} &
  \multicolumn{1}{c|}{\checkmark} &
  \multicolumn{1}{c|}{--} &
  \multicolumn{1}{c|}{\checkmark} &
  \multicolumn{1}{c|}{--} &
  \multicolumn{1}{c|}{--} &
  \multicolumn{1}{c|}{\checkmark} &
  \multicolumn{1}{c|}{\checkmark} &
  \multicolumn{1}{c|}{\checkmark} &
  \multicolumn{1}{c|}{--} \\ 
  \hline

  \multicolumn{1}{|c|}{\textbf{Blockchain}} &
  \multicolumn{14}{c|}{} \\ \hline
  
  \multicolumn{1}{|c|}{VFChain \cite{peng2021vfchain}} & 
  \multicolumn{1}{c|}{Synch} &
  \multicolumn{1}{c|}{10} &
  \multicolumn{1}{c|}{Decent} &
  \multicolumn{1}{c|}{--} &
  \multicolumn{1}{c|}{--} &
  \multicolumn{1}{c|}{\checkmark} &
  \multicolumn{1}{c|}{--} &
  \multicolumn{1}{c|}{--} &
  \multicolumn{1}{c|}{--} &
  \multicolumn{1}{c|}{--} &
  \multicolumn{1}{c|}{--} &
  \multicolumn{1}{c|}{\checkmark} &
  \multicolumn{1}{c|}{\checkmark} &
  \multicolumn{1}{c|}{--} \\ 
  \hline

  \multicolumn{1}{|c|}{FedMedChain \cite{samuel2022iomt}} & 
  \multicolumn{1}{c|}{Synch} &
  \multicolumn{1}{c|}{100} &
  \multicolumn{1}{c|}{Decent} &
  \multicolumn{1}{c|}{--} &
  \multicolumn{1}{c|}{--} &
  \multicolumn{1}{c|}{--} &
  \multicolumn{1}{c|}{--} &
  \multicolumn{1}{c|}{--} &
  \multicolumn{1}{c|}{--} &
  \multicolumn{1}{c|}{--} &
  \multicolumn{1}{c|}{--} &
  \multicolumn{1}{c|}{\checkmark} &
  \multicolumn{1}{c|}{\checkmark} &
  \multicolumn{1}{c|}{--} \\ 
  \hline

  \multicolumn{1}{|c|}{\textbf{}} &
  \multicolumn{14}{c|}{\textbf{Trusted Execution Environments}} \\ \hline
  
  \multicolumn{1}{|c|}{PPFL \cite{mo2021ppfl}} & 
  \multicolumn{1}{c|}{Synch} &
  \multicolumn{1}{c|}{10} &
  \multicolumn{1}{c|}{Cent} &
  \multicolumn{1}{c|}{\checkmark} &
  \multicolumn{1}{c|}{\checkmark} &
  \multicolumn{1}{c|}{\checkmark} &
  \multicolumn{1}{c|}{\checkmark} &
  \multicolumn{1}{c|}{--} &
  \multicolumn{1}{c|}{--} &
  \multicolumn{1}{c|}{--} &
  \multicolumn{1}{c|}{--} &
  \multicolumn{1}{c|}{\checkmark} &
  \multicolumn{1}{c|}{\checkmark} &
  \multicolumn{1}{c|}{--} \\ 
  \hline

  \multicolumn{1}{|c|}{CrowdGuard \cite{rieger2022close}} & 
  \multicolumn{1}{c|}{Cent} &
  \multicolumn{1}{c|}{20} &
  \multicolumn{1}{c|}{Hier} &
  \multicolumn{1}{c|}{--} &
  \multicolumn{1}{c|}{--} &
  \multicolumn{1}{c|}{--} &
  \multicolumn{1}{c|}{--} &
  \multicolumn{1}{c|}{--} &
  \multicolumn{1}{c|}{--} &
  \multicolumn{1}{c|}{--} &
  \multicolumn{1}{c|}{\checkmark} &
  \multicolumn{1}{c|}{\checkmark} &
  \multicolumn{1}{c|}{\checkmark} &
  \multicolumn{1}{c|}{--} \\ 
  \hline

\end{tabularx}
\begin{tablenotes}
      \footnotesize \item \textit{Synch}: Synchronization mode \{Synchronous, Asynchronous\}. \textit{Users}: Maximum active users used in experiments. \textit{Archi}: Architecture  \{Centralized, Decentralized(P2P), Hierarchical\}. \textit{Cm. E.}: Communication Efficiency. \textit{Cp. E.}: Computation Efficiency.\textit{ Conv}: Convergence Analysis. \textit{Pers}: Personalization. \textit{Gen}: Generalization. \textit{Fair}: Fairness. \textit{Heter}: Heterogeneity. \textit{Sec}: Security. \textit{Priv}: Privacy. \textit{Scal}: Scalability.
\end{tablenotes}
\end{threeparttable}
\end{table}

\subsection{Data Privacy-Oriented}
    Amidst the conflict between the privacy-by-design aspect of FL and the inherent privacy hurdles associated with its distributed nature, a pressing demand arises: extracting insights while upholding the privacy of data holders. Existing research centers on securing users' private data, local model updates, and the final deployed model to guarantee that no sensitive information has been revealed. These solutions foster trust among the diverse actors within the federated network, carefully addressing the crucial challenge of harmonizing privacy preservation with knowledge sharing. We summarize in Table \ref{tab-summary-FL-aggreg-privacy-data} a comprehensive array of strategies and scholarly papers published in this regard. 

    \begin{itemize}
        \item \textit{Perturbations.}
        
        The perturbation mechanisms in FL strategically inject controlled noise to the data or the local model before sharing it with other participants at the expense of a slight degradation in model accuracy. The goal is to protect data privacy and prevent unauthorized access or disclosure of sensitive information. \textit{\textbf{Differential Privacy (DP)}} stands as one of the most prominent perturbation methods in this domain. Its rigorous mathematical-based proof and flexible privacy-utility trade-offs made it relevant to many privacy-preserving applications in FL. According to \cite{zhang2023systematic}, the DP techniques in FL can be divided into three categories: (i) FL with central differential privacy: a trusted central server injects noise into the global parameters. (ii) FL with local differential privacy: each client incorporates noise into their local parameters. (iii) federated learning with distributed differential privacy: address the shortcomings of the two classes above by introducing only a small amount of noise locally and using a secure aggregation technique. From another perspective \cite{zhang2023systematic}, the field of DP research can be classified into two distinct categories: DP analysis \cite{zhang2022understanding} and DP proposals \cite{ho2022fedsgdcovid, zhang2023two}. Building on DP, the Poisson Binomial Mechanism (BPM) algorithm \cite{chen2022poisson} adds calibrated noise to FL updates, shielding device data from even in the presence of an untrusted aggregator. In particular, the injected noise is meticulously fine-tuned using quantified parameters for privacy (privacy epsilon, privacy budget), resulting in an unbiased yet optimal privacy-accuracy framework. Similarly, \cite{zhang2023two} leverages this strategy within an end-edge-cloud system. Researchers established a flexible two-stage DP policy that can be applied individually or synergistically. Initially, a randomized response algorithm perturbs the feature data. Then, edge servers infuse noise into the model, bolstering security.

        \item \textit{Homomorphic Encryption.}
        
        Homomorphic Encryption (HE) is an advanced cryptographic technique that facilitates computation on encrypted data without prior decryption. Within the context of FL, HE permits active devices to encrypt their local models, building a formidable barrier against sensitive information breaches. Hence, the aggregator can confidently execute arithmetic operations directly on the encrypted data in a seamlessly emulated fashion as if they were performed on plaintext data, preserving the privacy of individual contributors. Moreover, homomorphic encryption unfolds into three distinctive classes depending on the type and number of ciphertext operations configured: (\textit{i}) \textit{Partially Homomorphic Encryption (PHE)} supports one single operation, either addition or multiplication, for infinite times. (\textit{ii}) \textit{Somewhat Homomorphic Encryption (SHE)} enables both addition and multiplication, albeit with limitations on the number of computations. (\textit{iii}) \textit{Fully Homomorphic Encryption (FHE)} can perform any number of operations in the ciphertext space but requires advanced techniques and it is known to be computationally intensive. Numerous studies in the existing literature have utilized this intricate encryption method to establish robust federated learning frameworks, as shown in \cite{zhang2022homomorphic}, ShieldFL \cite{ma2022shieldfl}, and FedML-HE \cite{jin2023fedml}.

        \item \textit{Secure Aggregation.}
        
        Secure aggregation plays a pivotal role in upholding the privacy of FL users, guaranteeing that no party discloses its model update in the clear, even to the aggregator. The concern of secure aggregation has been extensively researched, with scholars employing diverse strategies to protect the aggregation process through the lens of privacy, including data perturbations, encryption, secure multi-party aggregation, and blockchain solutions.

        \textit{\textbf{Secure Multi-Party Computation (SMPC).}} 
        SMPC is a sub-field of cryptography wherein a consortium of data owners who do not mutually trust each other jointly perform computations for a specific task. This collaborative effort is underpinned by the strict condition that the confidentiality of the entities' data remains unbreached throughout the process. The main relevant techniques for implementing the SMPC include Secret Sharing (SS), Garbled Circuit (CC), and Oblivious Transfer (OT). To showcase this potent tool, we reference \cite{tian2022flvoogd}, which introduced FLVoogd to enhance efficiency and reduce resource-intensive operations. FLVoogd effectively rejects malicious uploads while protecting sensitive data, utilizing SMPC coupled with DP and Density-based Spatial Clustering of Applications with Noise (DBSCAN). Refer also to \cite{piotrowski2023towards} for a similar study.

\end{itemize}

\addtocounter{table}{-1}

\setlength\rotFPtop{0pt plus 1fil} 
\begin{table} [htbp]
\tiny
\caption{Summary of Federated Learning Aggregation Methods for Tackling Privacy - Data Privacy-Oriented.}
\label{tab-summary-FL-aggreg-privacy-data}
\begin{threeparttable}

\begin{tabularx}{\linewidth}{|X|XXX|XXXXXXXXXXX|}
\hline
\multicolumn{1}{|c|}{\multirow{3}{*}{\textbf{Related Work}}} &
  \multicolumn{3}{c|}{\multirow{2}{*}{\textbf{Environment}}} &
  \multicolumn{11}{c|}{\textbf{\textbf{Verified Goals}}} \\ \cline{5-15} 
 &
  \multicolumn{3}{c|}{} &
  \multicolumn{3}{c|}{\textbf{Process}} &
  \multicolumn{5}{c|}{\textbf{Model}} &
  \multicolumn{3}{c|}{\textbf{System}} \\ \cline{2-15} 
 &
  \multicolumn{1}{c|}{\textbf{Synch}} & 
  \multicolumn{1}{c|}{\textbf{Users}} & 
  \multicolumn{1}{c|}{\textbf{Archi}} & 
  
  \multicolumn{1}{c|}{\textbf{Cm. E.}} & 
  \multicolumn{1}{c|}{\textbf{Cp. E.}} & 
  \multicolumn{1}{c|}{\textbf{Conv}} & 
  \multicolumn{1}{c|}{\textbf{Pers}} & 
  \multicolumn{1}{c|}{\textbf{Gen}} & 
  \multicolumn{1}{c|}{\textbf{Reg}} & 
  \multicolumn{1}{c|}{\textbf{Fair}} & 
  \multicolumn{1}{c|}{\textbf{Heter}} & 
  \multicolumn{1}{c|}{\textbf{Sec}} & 
  \multicolumn{1}{c|}{\textbf{Priv}} & 
  \multicolumn{1}{c|}{\textbf{Scal}} \\  
  \hline


  \multicolumn{1}{|c|}{\textbf{Used Techniques}} &
  \multicolumn{14}{c|}{\textbf{Perturbations}} \\ \hline
  \multicolumn{1}{|c|}{\textbf{DP}} &
  \multicolumn{14}{c|}{} \\ \hline

  \multicolumn{1}{|c|}{PBM \cite{chen2022poisson}} & 
  \multicolumn{1}{c|}{Synch} &
  \multicolumn{1}{c|}{--} & 
  \multicolumn{1}{c|}{Cent} &
  \multicolumn{1}{c|}{\checkmark} &
  \multicolumn{1}{c|}{--} &
  \multicolumn{1}{c|}{--} &
  \multicolumn{1}{c|}{--} &
  \multicolumn{1}{c|}{--} &
  \multicolumn{1}{c|}{--} &
  \multicolumn{1}{c|}{--} &
  \multicolumn{1}{c|}{--} &
  \multicolumn{1}{c|}{\checkmark} &
  \multicolumn{1}{c|}{\checkmark} &
  \multicolumn{1}{c|}{--} \\ 
  \hline

  \multicolumn{1}{|c|}{\cite{ho2022fedsgdcovid}} & 
  \multicolumn{1}{c|}{Synch} &
  \multicolumn{1}{c|}{3} &
  \multicolumn{1}{c|}{Cent} &
  \multicolumn{1}{c|}{--} &
  \multicolumn{1}{c|}{--} &
  \multicolumn{1}{c|}{\checkmark} &
  \multicolumn{1}{c|}{--} &
  \multicolumn{1}{c|}{--} &
  \multicolumn{1}{c|}{--} &
  \multicolumn{1}{c|}{\checkmark} &
  \multicolumn{1}{c|}{--} &
  \multicolumn{1}{c|}{\checkmark} &
  \multicolumn{1}{c|}{\checkmark} &
  \multicolumn{1}{c|}{--} \\ 
  \hline

  \multicolumn{1}{|c|}{\cite{zhang2023two}} & 
  \multicolumn{1}{c|}{Synch} &
  \multicolumn{1}{c|}{8} &
  \multicolumn{1}{c|}{Hier} &
  \multicolumn{1}{c|}{--} &
  \multicolumn{1}{c|}{--} &
  \multicolumn{1}{c|}{\checkmark} &
  \multicolumn{1}{c|}{--} &
  \multicolumn{1}{c|}{--} &
  \multicolumn{1}{c|}{--} &
  \multicolumn{1}{c|}{\checkmark} &
  \multicolumn{1}{c|}{--} &
  \multicolumn{1}{c|}{--} &
  \multicolumn{1}{c|}{\checkmark} &
  \multicolumn{1}{c|}{--} \\ 
  \hline
   
  \multicolumn{1}{|c|}{} &
  \multicolumn{14}{c|}{\textbf{Homomorphic Encryption}} \\ \hline

  \multicolumn{1}{|c|}{ \cite{zhang2022homomorphic}} &
  \multicolumn{1}{c|}{Synch} &
  \multicolumn{1}{c|}{30} &
  \multicolumn{1}{c|}{Cent} &
  \multicolumn{1}{c|}{\checkmark} &
  \multicolumn{1}{c|}{\checkmark} &
  \multicolumn{1}{c|}{\checkmark} &
  \multicolumn{1}{c|}{--} &
  \multicolumn{1}{c|}{--} &
  \multicolumn{1}{c|}{--} &
  \multicolumn{1}{c|}{--} &
  \multicolumn{1}{c|}{--} &
  \multicolumn{1}{c|}{\checkmark} &
  \multicolumn{1}{c|}{\checkmark} &
  \multicolumn{1}{c|}{--} \\ 
  \hline

  \multicolumn{1}{|c|}{\cite{jin2023fedml}} & 
  \multicolumn{1}{c|}{Synch} &
  \multicolumn{1}{c|}{200} &
  \multicolumn{1}{c|}{Cent} &
  \multicolumn{1}{c|}{\checkmark} &
  \multicolumn{1}{c|}{\checkmark} &
  \multicolumn{1}{c|}{\checkmark} &
  \multicolumn{1}{c|}{\checkmark} &
  \multicolumn{1}{c|}{--} &
  \multicolumn{1}{c|}{--} &
  \multicolumn{1}{c|}{--} &
  \multicolumn{1}{c|}{--} &
  \multicolumn{1}{c|}{\checkmark} &
  \multicolumn{1}{c|}{\checkmark} &
  \multicolumn{1}{c|}{\checkmark} \\ 
  \hline

  \multicolumn{1}{|c|}{} &
  \multicolumn{14}{c|}{\textbf{Secure Aggregation}} \\ \hline
  \multicolumn{1}{|c|}{\textbf{SMPC}} &
  \multicolumn{14}{c|}{} \\ \hline

  \multicolumn{1}{|c|}{FLVoogd \cite{tian2022flvoogd}} &
  \multicolumn{1}{c|}{Synch} &
  \multicolumn{1}{c|}{100} &
  \multicolumn{1}{c|}{Cent} &
  \multicolumn{1}{c|}{--} &
  \multicolumn{1}{c|}{--} &
  \multicolumn{1}{c|}{\checkmark} &
  \multicolumn{1}{c|}{--} &
  \multicolumn{1}{c|}{--} &
  \multicolumn{1}{c|}{--} &
  \multicolumn{1}{c|}{--} &
  \multicolumn{1}{c|}{--} &
  \multicolumn{1}{c|}{\checkmark} &
  \multicolumn{1}{c|}{\checkmark} &
  \multicolumn{1}{c|}{--} \\ 
  \hline
  \multicolumn{1}{|c|}{ \cite{piotrowski2023towards}} & 
  \multicolumn{1}{c|}{Asynch} &
  \multicolumn{1}{c|}{4} &
  \multicolumn{1}{c|}{Decent} &
  \multicolumn{1}{c|}{\checkmark} &
  \multicolumn{1}{c|}{--} &
  \multicolumn{1}{c|}{--} &
  \multicolumn{1}{c|}{--} &
  \multicolumn{1}{c|}{--} &
  \multicolumn{1}{c|}{--} &
  \multicolumn{1}{c|}{--} &
  \multicolumn{1}{c|}{\checkmark} &
  \multicolumn{1}{c|}{\checkmark}&
  \multicolumn{1}{c|}{\checkmark} &
  \multicolumn{1}{c|}{--} \\ 
  \hline
  
  \multicolumn{1}{|c|}{\textbf{Blockchain}} &
  \multicolumn{14}{c|}{} \\ \hline

  \multicolumn{1}{|c|}{\cite{kumar2021blockchain}} &
  \multicolumn{1}{c|}{Synch} &
  \multicolumn{1}{c|}{3} &
  \multicolumn{1}{c|}{Decent} &
  \multicolumn{1}{c|}{--} &
  \multicolumn{1}{c|}{--} &
  \multicolumn{1}{c|}{\checkmark} &
  \multicolumn{1}{c|}{--} &
  \multicolumn{1}{c|}{\checkmark} &
  \multicolumn{1}{c|}{--} &
  \multicolumn{1}{c|}{--} &
  \multicolumn{1}{c|}{\checkmark} &
  \multicolumn{1}{c|}{--} &
  \multicolumn{1}{c|}{\checkmark} &
  \multicolumn{1}{c|}{--} \\ 
  \hline

\end{tabularx}
\begin{tablenotes}
      \footnotesize \item \textit{Synch}: Synchronization mode \{Synchronous, Asynchronous\}. \textit{Users}: Maximum active users used in experiments. \textit{Archi}: Architecture  \{Centralized, Decentralized(P2P), Hierarchical\}. \textit{Cm. E.}: Communication Efficiency. \textit{Cp. E.}: Computation Efficiency.\textit{ Conv}: Convergence Analysis. \textit{Pers}: Personalization. \textit{Gen}: Generalization. \textit{Fair}: Fairness. \textit{Heter}: Heterogeneity. \textit{Sec}: Security. \textit{Priv}: Privacy. \textit{Scal}: Scalability.
\end{tablenotes}
\end{threeparttable}
\end{table}

To summarize, we provide in Table \ref{tab_attacks_defenses}, a summary of the common FL attacks along with their corresponding defense mechanisms, emphasizing the key findings of this section. 

\addtocounter{table}{-1}

\begin{table}[htbp]
\tiny
\caption{FL Popular Attacks and Corresponding Defenses.}
\label{tab_attacks_defenses}
\begin{threeparttable}
\begin{tabularx}{\linewidth}{| >{\hsize=.29\hsize\linewidth=\hsize}X|
>{\hsize=.03\hsize\linewidth=\hsize}X|
 >{\hsize=.04\hsize\linewidth=\hsize}X|
  >{\hsize=.05\hsize\linewidth=\hsize}X|
  >{\hsize=.05\hsize\linewidth=\hsize}X|
  >{\hsize=.03\hsize\linewidth=\hsize}X|
  >{\hsize=.03\hsize\linewidth=\hsize}X|
  >{\hsize=.04\hsize\linewidth=\hsize}X|
  >{\hsize=.03\hsize\linewidth=\hsize}X|
  >{\hsize=.03\hsize\linewidth=\hsize}X|
  >{\hsize=.04\hsize\linewidth=\hsize}X|
  >{\hsize=.03\hsize\linewidth=\hsize}X|
  >{\hsize=.04\hsize\linewidth=\hsize}X|
  >{\hsize=.03\hsize\linewidth=\hsize}X|
  >{\hsize=.03\hsize\linewidth=\hsize}X|
  >{\hsize=.05\hsize\linewidth=\hsize}X|}
\hline
 & \multicolumn{15}{|c|}{\textbf{Defenses}}\\
\hline
\textbf{Attacks} & \scriptsize{AD} &\scriptsize{ZKPs} &\scriptsize{Snipper} &\scriptsize{Pruning} & \scriptsize{AT} & \scriptsize{FD} & \scriptsize{FMT} & \scriptsize{RA} & \scriptsize{MT} & \scriptsize{P2P} &\scriptsize {BC} &\scriptsize {TEE} & \scriptsize{DP} & \scriptsize{HE} & \scriptsize{SMPC} \\ [0.75ex] \hline
\scriptsize{Data Poisoning}           
 & \checkmark   
 & \checkmark   
 & \checkmark      
 &  \text{--}       
 & \text{--}     
 & \text{--}     
 & \text{--}       
 & \checkmark   
 &  \text{--}    
 & \text{--}    
 & \text{--}   
 & \text{--}    
 & \checkmark 
 & \text{--}   
 & \text{--}                       
 \\[0.75ex] \hline

\scriptsize{Model Poisoning}            
& \checkmark   
& \checkmark   
& \checkmark      
&  \text{--}       
&  \text{--}    
& \text{--}     
& \checkmark     
& \checkmark   
&  \text{--}    
&  \text{--}  
& \checkmark 
&  \text{--}   
& \checkmark 
&  \text{--}  
&  \text{--}                        
\\[0.75ex] \hline

\scriptsize {Backdoors}                   
&      
& \checkmark   
& \checkmark      
& \checkmark      
&  \text{--}      
&  \text{--}      
&  \text{--}        
& \checkmark   
&   \text{--}     
&  \text{--}     
&  \text{--}   
&  \text{--}    
&  \text{--}    
&  \text{--}      
&  \text{--}                      
\\ [0.75ex] \hline

\scriptsize {Training Rules Manipulation} 
& \text{--}       
& \text{--}       
& \text{--}          
& \text{--}          
& \text{--}       
& \text{--}       
& \text{--}        
& \text{--}       
& \text{--}       
& \text{--}      
& \text{--}     
& \checkmark  
&  \text{--}    
&  \text{--}    
& \checkmark                     
\\ [0.75ex] \hline

\scriptsize {GAN Reconstruction}        
& \text{--}        
& \text{--}       
& \text{--}          
& \text{--}          
& \text{--}       
& \checkmark   
& \text{--}         
& \text{--}       
& \checkmark   
& \text{--}      
& \text{--}     
& \checkmark  
& \text{--}     
& \text{--}     
& \checkmark                      
\\ [0.75ex] \hline

\scriptsize {Evasion Attacks}             
& \text{--}       
& \text{--}       
& \text{--}          
& \text{--}           
& \checkmark   
&  \text{--}      
& \text{--}         
& \text{--}       
& \text{--}       
& \text{--}      
& \text{--}     
& \checkmark  
& \checkmark 
& \text{--}     
& \checkmark                      
\\ [0.75ex] \hline

\scriptsize {Single Point of Failure}     
& \text{--}       
& \text{--}       
& \text{--}          
& \text{--}          
& \text{--}       
& \text{--}       
& \text{--}         
& \checkmark   
& \text{--}       
& \checkmark  
& \checkmark 
& \checkmark  
&  \text{--}    
& \text{--}      
& \text{--}                         
\\ [0.75ex] \hline

\scriptsize {Non Robust Aggregation}      
&  \text{--}     
&  \text{--}      
& \text{--}          
& \text{--}          
& \text{--}      
& \text{--}       
& \text{--}         
& \checkmark   
& \text{--}       
& \text{--}      
& \checkmark 
& \text{--}      
& \text{--}     
& \text{--}     
& \text{--}                           
\\  [0.75ex] \hline

\scriptsize{Clients Unavailability}      
& \text{--}       
& \text{--}       
& \text{--}          
& \checkmark      
&  \text{--}      
& \checkmark   
& \checkmark     
& \checkmark   
&  \text{--}      
& \checkmark  
& \text{--}     
& \text{--}      
& \text{--}     
& \text{--}     
& \text{--}                          
\\ [0.75ex] \hline

\scriptsize {Information Leakage}         
&  \text{--}      
&  \text{--}      
&  \text{--}         
&  \text{--}         
&  \text{--}      
&  \text{--}      
&  \text{--}        
& \text{--}       
& \text{--}       
& \checkmark  
& \checkmark 
& \checkmark  
& \checkmark 
& \checkmark 
& \checkmark                      
\\ [0.75ex] \hline

\scriptsize {Inference Attacks}           
&  \text{--}      
&  \text{--}      
& \text{--}          
&  \text{--}         
& \text{--}       
& \checkmark   
& \checkmark     
& \text{--}       
& \checkmark   
& \text{--}      
& \checkmark 
& \text{--}      
& \checkmark 
& \checkmark 
& \checkmark                     
\\ [0.75ex] \hline

\scriptsize {Free-Rider Attacks}         
& \checkmark   
&  \text{--}      
&  \text{--}  
&  \text{--}         
&  \text{--}      
&  \text{--}      
&   \text{--}       
& \checkmark   
& \text{--}       
&  \text{--}     
&  \text{--}    
& \checkmark  
&  \text{--}    
&  \text{--}  
& \text{--}                          
\\ [0.75ex] \hline

\scriptsize{Evasdroppers}                
& \text{--}       
& \text{--}       
& \text{--}          
& \text{--}          
& \text{--}       
& \text{--}       
& \text{--}         
& \text{--}       
& \text{--}       
& \text{--}     
& \checkmark 
& \checkmark  
& \text{--}     
& \checkmark 
& \checkmark                      
\\ [0.75ex] \hline

\scriptsize {Man in The Middle}          
&      
& \checkmark   
&  \text{--}         
&  \text{--}         
& \text{--}       
& \checkmark   
&  \text{--}        
&  \text{--}      
& \checkmark   
&  \text{--}     
& \checkmark 
&  \text{--}     
&  \text{--}  
&  \text{--}  
& \checkmark 
\\ [0.75ex] \hline

\end{tabularx}
\begin{tablenotes}
    \footnotesize 
    \item \textit{AD}: Anomaly Detection. \textit{ZKPs}: Zero Knowledge Proofs. \textit{AT}: Adversarial Training. \textit{FD}: Federated Distillation. \textit{FMTL}: Federated Multi-Task Learning. \textit{RA}: Robust Aggregation. \textit{MT}: Moving Target. \textit{P2P} Peer-to-Peer Aggregation. \textit{BC}: Blockchain. \textit{TEE}: Trusted Execution Environment. \textit{DP}: Differential Privacy. \textit{HE}: Homomorphic Encryption. \textit{SMPC}: Secure Multi-Party Computation.
\end{tablenotes}

\end{threeparttable}
\clearpage
\end{table} 
\section{Experiments and Results}
\label{FL_simulations}
\subsection{Experimental Goals}

   In this section, we unveil the experimental findings of our study, where we assessed a range of aggregation methods using benchmarking datasets and leveraging PyTorch for implementation. Our primary aim is to provide valuable insights for fellow researchers into evaluating novel FL aggregation proposals in real-world settings.

    Initially, we explore the relevant parameters crucial for validating the correctness and efficiency of proposed solutions. These include the hyperparameters of DL models, benchmark datasets, real-world data distributions, and other unique FL parameters like the number of clients and various heterogeneity types. Following this, we examine the responsiveness trends of different aggregation algorithms, each from a distinct class of solution, in order to understand how various variables in the FL ecosystem impact overall performance. Simultaneously, we discuss how each algorithm, representing various classes, responds to variations in the considered variables.
    In other words, we objectively compare the performance results of the selected algorithms under these relevant settings.

    In essence, we endeavor to offer a nuanced perspective that aids in the holistic understanding of FL practical implementation, providing researchers with a valuable resource for evaluating the aggregation performance of their proposals within realistic simulation scenarios.

\subsection{Experimental Baseline}
\subsubsection{\textbf{Datasets}}

For the sake of simplicity and due to the absence of universally recognized benchmark datasets, we chose to use some of the widely adopted datasets for image classification in the research community. Our selection criteria focused on datasets with a rich number of samples, allowing us to partition the data into hundreds of clients while ensuring each client had sufficient data for adequate training. By employing diverse datasets for the same task (image classification), we aimed to explore varying levels of complexity exhibited in terms of shapes, textures, and patterns. Table \ref{datasets_specs} summarizes the characteristics of the employed datasets.

\begin{itemize}
    \item CIFAR-10 and CIFAR-100 \cite{krizhevsky2009learning} are computer vision datasets widely used in the field. They consist of 60,000 color images, each with a dimension of 32x32 pixels. CIFAR-10 is composed of images representing 10 different classes, including ubiquitous objects like animals, vehicles, household items, and other categories. CIFAR-100 expands the scope with 100 different classes covering a wider range of objects for more challenging multi-task classification. 
    \item FashionMNIST is a popular dataset \cite{xiao2017fashion} that comprises 60,000 grayscale images, each with a resolution of 28x28 pixels, covering 10 distinct classes. However, instead of depicting general objects, FashionMNIST specializes in clothing and accessories items. 
    \item MedMNIST \cite{yang2021medmnist, yang2023medmnist} is large-scale MNIST-like biomedical images, including 12 datasets for 2D and six datasets for 3D. All images are 28 × 28 (2D) or 28 × 28 × 28 (3D). For our experiments, we select three datasets: OrganAMNIST, OrganCMNIST, and OrganSMNIST, as in \cite{lu2022personalized}. These three datasets are all about Abdominal CT images illustrating 11 distinct classes with 58,850, 23,660, and 25,221 samples, respectively.
\end{itemize}
\addtocounter{table}{-1}

\begin{table} [htbp]
\scriptsize
\caption{Summary of Different Dataset Characteristics}
\label{datasets_specs}
\begin{tabularx}{\textwidth}{|>{\hsize=.3\hsize\textwidth=\hsize}X|  >{\hsize=.2\hsize\textwidth=\hsize}X|  >{\hsize=.2\hsize\textwidth=\hsize}X |>{\hsize=.2\hsize\textwidth=\hsize}X| }
\hline  
\textbf{Data Name} & \textbf{CIFAR-10}  & \textbf{FMNIST} & \textbf{Med-MNIST}  \\ 

\hline
\textbf{Type} 
& Object Images 
& Fashion Images
& Biomedical Images\\
\hline
\textbf{\# of Train Samples} 
&  45K
&  54K
&  53,866\\
\hline
\textbf{\# of Val Samples} 
& 5k 
& 6k 
& 21,546\\
\hline
\textbf{\# of Test Samples} 
& 10k
& 10k
& 16,159\\
\hline
\textbf{Distinct Classes} 
&  10
&  10
& 11\\
\hline
\textbf{Size} 
&  32x32
&  28x28
&  28x28\\
\hline
\textbf{\# of Clients} 
& 50
& 50
& 10\\
\hline
\textbf{Availability} 
&  torchvision
&  torchvision
&  \href{https://medmnist.com/}{medmnist}\\
\hline
\end{tabularx}
\end{table}

\subsubsection{\textbf{Data Distributions}}

\begin{figure}[htbp]
    \centering
\subfloat[\scriptsize{IID-Balanced data partition given 100 clients \cite{JMLR:v24:22-0440}.}]
    {\includegraphics[scale=0.4]{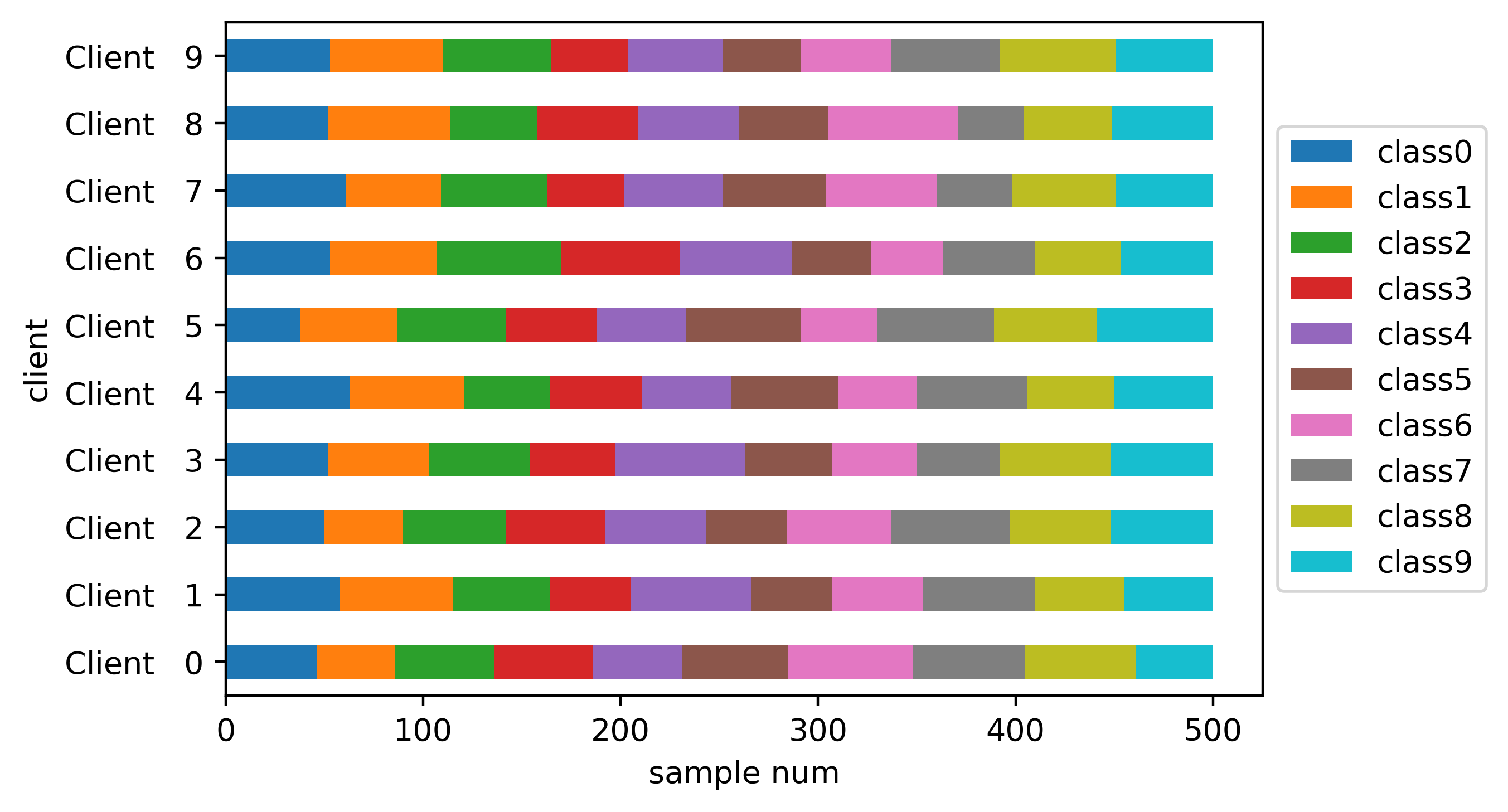}
    \label{balanced_iid_data_dist}}
    \hspace{0.5em}%
\subfloat[\scriptsize{Unbalanced-Dirichlet data partition given 100 clients,$\sigma = 0.3$, $\alpha = 0.3$ \cite{JMLR:v24:22-0440}.}]
    {\includegraphics[scale=0.4]{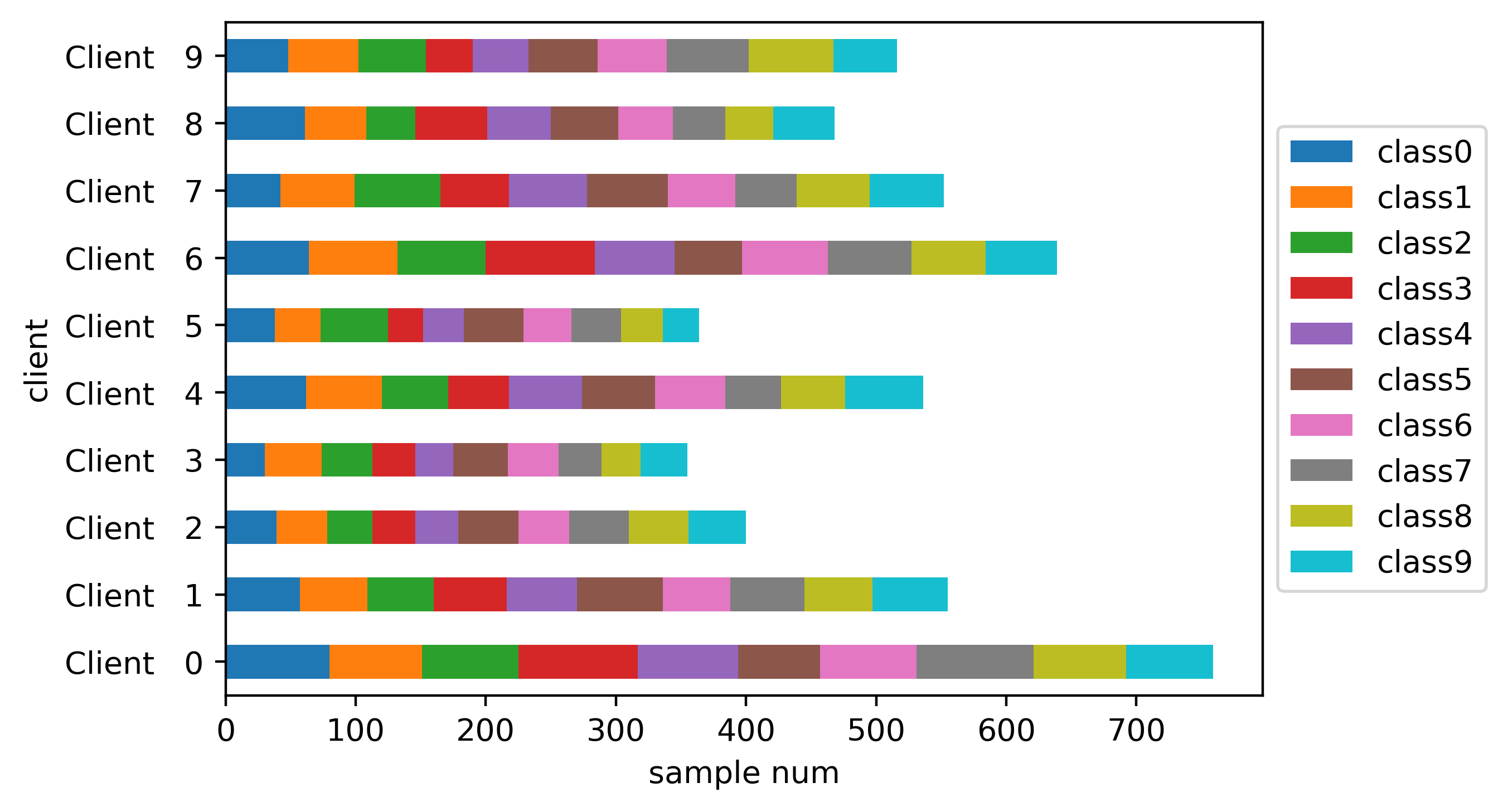}
    \label{unbalanced_noniid_dirichelet}}
    \hspace{0.5em}%
\caption{Data Distribution for CIFAR-10 Dataset}
\label{data_dist_cifar}
\end{figure}

\begin{figure}[htbp]
    \centering
    \includegraphics[width=0.5\textwidth,height=3.5cm]{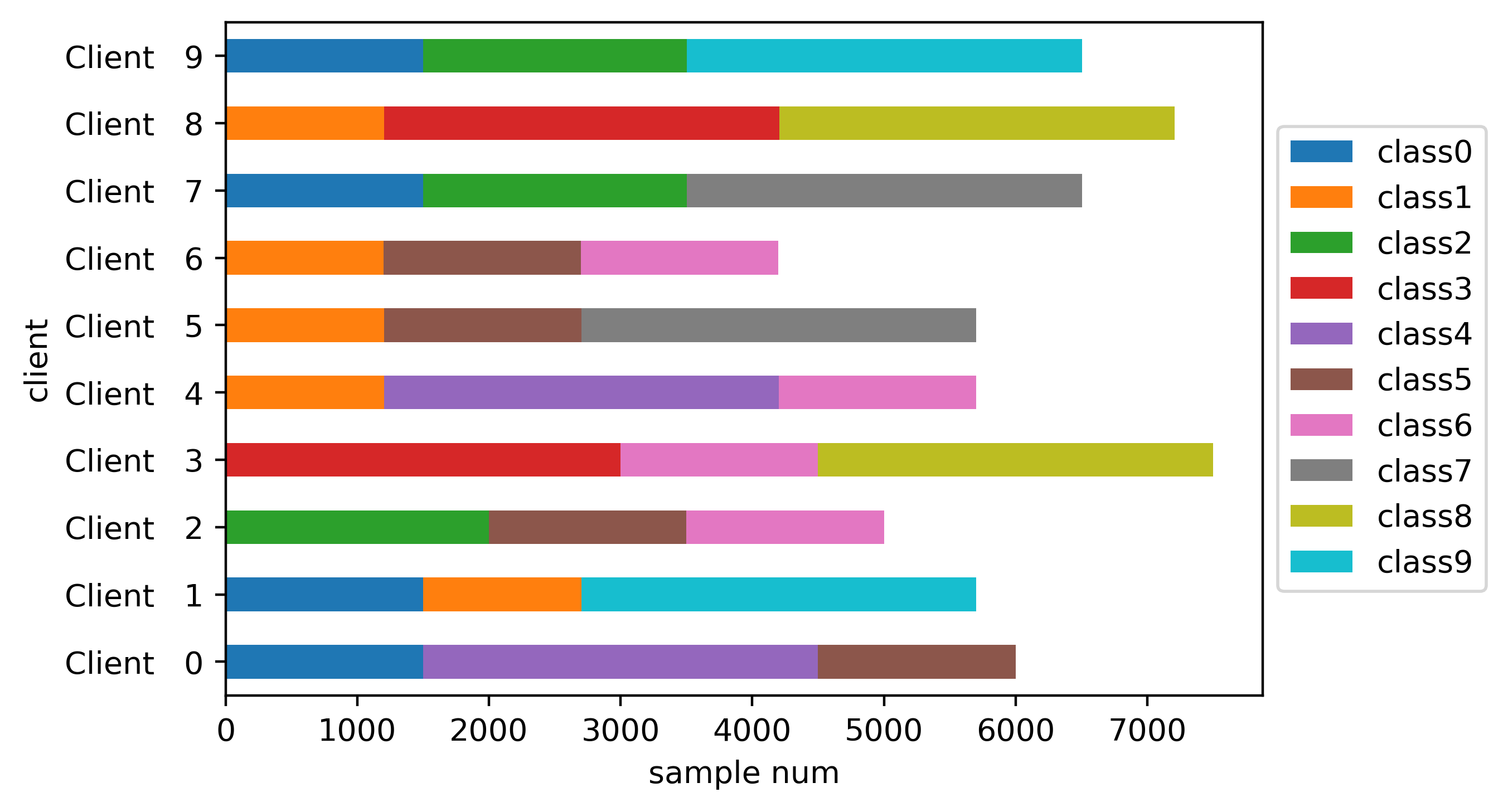}
    \caption{Quantity-based Label Distribution Skew data partition for FashionMNIST, given 100 clients and $class\_num = 3$ for each client \cite{JMLR:v24:22-0440}.}
    \label{quantity_based_label_skew}
\end{figure}

The evaluation of the FL ecosystem must encompass diverse data distribution scenarios, accounting for heterogeneity among clients. Therefore, it is vital to explore non-IID partitions that more accurately mirror real-world conditions besides the simplistic IID distribution that serves as a baseline. However, it is worth noting that numerous research works often lack explicit details regarding the specific non-IID partitions they use despite the various possible schemes for creating such diverse setups. In our simulations, we drew inspiration from the FedLab framework \cite{JMLR:v24:22-0440}, which provides eight (8) distinct classes of data distribution for FL settings, as summarized in Table \ref{FedLAb_data_dist}. From this set, we selected the most suitable categories, including \textit{IID-Balanced} for baseline comparison, \textit{Unbalanced-Dirichlet}, and \textit{Quantity-based Label Distribution Skew} for varying levels of statistical heterogeneity. Fig. \ref{balanced_iid_data_dist} illustrates the data samples assigned to the first 10 clients under the IID-Balanced partition when involving 100 clients and the CIFAR-10 dataset. While Fig. \ref{unbalanced_noniid_dirichelet} and Fig. \ref{quantity_based_label_skew} depict the data samples resulting from the Unbalanced-Dirichlet and Quantity-based Label Distribution Skew partitions, for CIFAR-10 and FashionMNIST, respectively.
\addtocounter{table}{-1}
\begin{table} [htbp]
\scriptsize
\caption{Summary of the Distribution Functions of Different Data Partition Scenarios in FL.}
\label{FedLAb_data_dist}
\begin{threeparttable}

\begin{tabularx}{\linewidth}{| >{\hsize=.3\hsize\linewidth=\hsize}X|  >{\hsize=.35\hsize\linewidth=\hsize}X|  >{\hsize=.35\hsize\linewidth=\hsize}X |}
\hline  
\textbf{Data Partition Class} & \textbf{P1}  & \textbf{P2}  \\ 
\hline
\textbf{Balanced IID} 
&  Same number
& Same distribution\\
\hline
\textbf{Unbalanced IID} 
& Log-Normal ($Log\_N(0,\sigma^2)$)
& Same distribution \\
\hline
\textbf{Non-iid Dirichlet \cite{yurochkin2019bayesian, wang2020federated} } 
&  Dirichlet ($Dir (\alpha)$)
&  Unbalanced ($P_{k,j}$ class $k$, client $j$) \\
\hline
\textbf{Shards \cite{mcmahan2017communication}} 
&  -
&  -\\
\hline
\textbf{Balanced Dirichlet \cite{acar2021federated}} 
&  Same number
&  Dirichlet (Dir($\alpha$)) \\
\hline
\textbf{Unbalanced Dirichlet \cite{acar2021federated}} 
&  Log-Normal ($Log\_N(0,\sigma^2)$)
&  Dirichlet (Dir($\alpha$)) \\
\hline
\textbf{Label Distribution Skew \cite{li2022federated}} 
&  Unbalanced
&  Only a specific number of sample classes.\\
\hline
\textbf{Feature Distribution Skew \cite{li2022federated}} 
&  -
&  Sample feature with Gaussian noise \\
\hline
\end{tabularx}
    \begin{tablenotes}
        \footnotesize \item 
        \textit{P1}: Number of samples for each client, \textit{P2}: distribution for different class samples at each client.
    \end{tablenotes}
\end{threeparttable}

\end{table}

\subsubsection{\textbf{Model Architecture and Hyper-parameters}}
We have employed CNNs as our deep learning models. Specifically, we used AlexNet \cite{krizhevsky2012imagenet} for the CIFAR dataset and LeNet-5 \cite{lecun1998gradient} for the FashionMNIST and MedMNIST datasets. AlexNet and LeNet-5 are renowned architectures tailored for computer vision tasks. AlexNet consists of eight layers, while LeNet-5 consists of seven layers. In all our experiments, we standardize the number of epochs to 5 after fine-tuning from options of [1,3,5], the batch size to 16 or 32 depending on the specific configuration, chosen from a range of [1,16,32,64]. We conducted the training for 300 iterations (number of rounds) after considering several options [100,200,300,400] with a fixed learning rate of $10^{-3}$ (refer to Table \ref{cnn_model_params}). It's important to note that our fine-tuning simulations were based only on the best performance achieved using FedAvg. 

\addtocounter{table}{-1}

\begin{table} [htbp]
\scriptsize
\caption{ Model Hyper-parameter Values Used for Experiments}
\label{cnn_model_params}
\begin{tabularx}{0.5\linewidth}{| >{\hsize=.5\hsize\linewidth=\hsize}X|  >{\hsize=.5\hsize\linewidth=\hsize}X| }
\hline  
\textbf{Param} & \textbf{Value} \\ 
\hline
\textbf{\# of Iterations} 
& 300\\
\hline
\textbf{\# of Epochs} 
& 5 \\
\hline
\textbf{Batch Size} 
& 16 / 32 \\
\hline
\textbf{Learning Rate} 
& 0.001\\
\hline
\textbf{Loss Function} 
& Cross-Entropy \\
\hline
\textbf{Optimizer} 
& SGD \\
\hline
\end{tabularx}

\end{table}

\subsubsection{\textbf{Federated Learning Aggregation Algorithms}}

Recognizing the impracticality of assessing all available aggregation methods, we have limited our evaluation of FL aggregation algorithms to four distinct approaches. While the literature boasts a multitude of pioneering aggregation strategies, our intention is to furnish a comprehensive yet focused evaluation of representative approaches that encapsulate the fundamental characteristics and trends. Specifically, for our selection criteria, we have focused on two key factors. Firstly, we aimed to encompass a wide range of contributions families provided by the picked set of algorithms, prioritizing the selection of only one algorithm per solution family. Secondly, we considered the availability of their implementation. As a result, we determined the following set of algorithms for our study:
\begin{itemize}
    \item \textbf{FedAvg \cite{mcmahan2017communication}} The server aggregates all client models through a basic averaging process.
    \item \textbf{FedDyn \cite{acar2021federated}} The algorithm uses regularization terms to speed up convergence and prevent local models from deviating significantly from the global model. Notably, we assess FedDyn over other well-established regularization-based aggregation methods, such as FedProx and Scaffold, given their extensive usage in prior evaluation experiments documented in the literature.
    \item \textbf{FedBN \cite{li2021fedbn}} It employs batch normalization to improve generalization.
    \item \textbf {Power-of-Choice \cite{cho2020client}} The aggregation process relies on client selection determined by loss.
\end{itemize}
In Table \ref{test_algo_functions}, we present the functions employed on both the client and server sides for each algorithm.


\addtocounter{table}{-1}

\begin{table} [htbp]
\scriptsize
\caption{Algorithm Functions}
\label{test_algo_functions}
\begin{tabularx}{\linewidth}{| >{\hsize=.09\hsize\linewidth=\hsize}X|  >{\hsize=.44\hsize\linewidth=\hsize}X| >{\hsize=.47\hsize\linewidth=\hsize}X|}
\hline  
\textbf{Algo} & \textbf{Server Side}  & \textbf{Client Side} \\ 
\hline
\textbf{FedAvg}
& 
\begin{math}
   w = \frac{1}{N} \sum_{i=1}^{N} w_i 
\end{math}

& 
\begin{math}
 w_i' = \arg\min_{w_i} \left( L(w_i, D_i)  + \lambda \lVert w_i - w_{\text{avg}} \rVert^2 \right)  
\end{math} 

\\
\hline
\textbf{FedDyn}
& 

\begin{math}
R_i' = \text{Update}(w_i, w_{i,t-1},\alpha ) 
\end{math}

\begin{math}
w_t = \arg\min_w \left\{ \text{FedAvg}(w) + \lambda R(w) \right\}
\end{math}

& 
\begin{math}
    w_i' = \arg\min_w \left\{ L(w_i) + R_i' ( w_i, w_{\text{avg}} )  \right\}
\end{math}
 \\
\hline

\textbf{FedBN}
&
\begin{math}
w_{\text{avg}}' = \text{FedAvg}\left(w_t, \gamma_t, \beta_t \right)    
\end{math}

\begin{math}
    \gamma \leftarrow \frac{\sum_{k=1}^{K} n_k \gamma_k}{\sum_{k=1}^{K} n_k}
\end{math}
\begin{math}
    \beta \leftarrow \frac{\sum_{k=1}^{K} n_k \beta_k}{\sum_{k=1}^{K} n_k}
\end{math}

& 
\begin{math}
    w_i' = w_i - \eta \frac{1}{|\mathcal{B}|} \sum_{i \in \mathcal{B}} \frac{\partial L_i}{\partial w}
\end{math}

\begin{math}
    \gamma \leftarrow \gamma - \eta \frac{1}{|\mathcal{B}|} \sum_{i \in \mathcal{B}} \frac{\partial L_i}{\partial w}
\end{math}

\begin{math}
    \beta \leftarrow \beta - \eta \frac{1}{|\mathcal{B}|} \sum_{i \in \mathcal{B}} \frac{\partial L_i}{\partial \beta}
\end{math}
\\
\hline
\textbf{{Power-of-Choice}}
&
\begin{math}
   w_{\text{avg}}' = \text{FedAvg}\left(w_t, S(t) \right) 
\end{math}

\begin{math}
    S(t)=\left\{ k : \left( t \mod \tau \right) + 1 \leq j \right\},
\end{math}

\begin{math}
    \left\{j \leq \left( t \mod \tau \right) + m\right\},
\end{math}

\begin{math}
    j \equiv k \pmod{K}
\end{math}
&
\begin{math}
   w_i' = \arg\min_{w} \left\{F_i(w) + \frac{\lambda}{2} \|w_i - w_{\text{avg}}\|^2 
   \right\}
\end{math}
 \\
\hline

\end{tabularx}
\end{table}

\subsection{Experimental Results}

\subsubsection{Impact of Number of Clients}
In our initial investigation of parameters, we focus on understanding how the number of clients affects the aggregation process. To isolate this factor and exclusively observe its effect on FL performance, we conducted experiments using an IID data distribution with the CIFAR10 dataset for the FedAvg, Fedbn, and FedDyn algorithms.

\begin{figure}[htbp]

\begin{subfigure}{0.35\textwidth}
    \includegraphics[width=\textwidth]{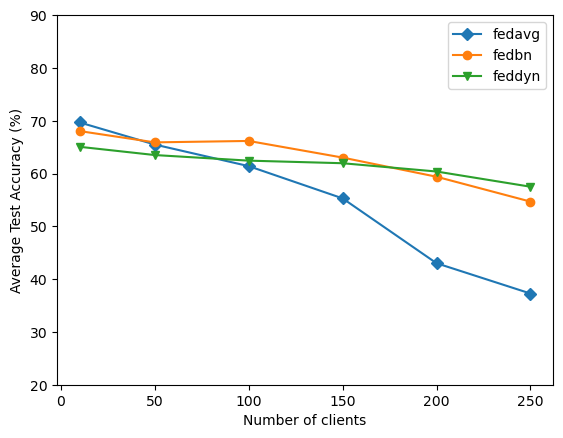}
    \caption{Average Accuracy Variation}
    \label{n_clients_acc}
\end{subfigure}
\begin{subfigure}{0.35\textwidth}
    \includegraphics[width=\textwidth]{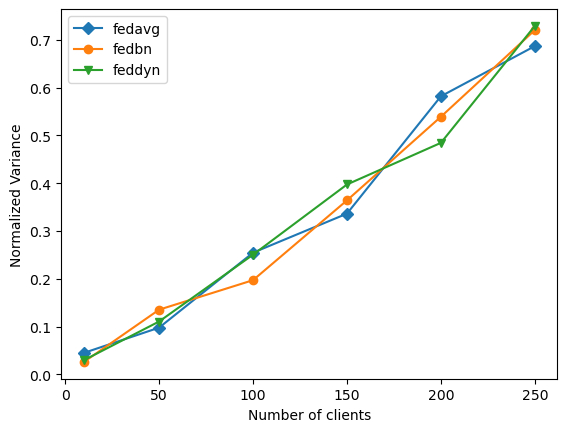}
    \caption{Normalized Variance}
    \label{n_clients_var}
\end{subfigure}
\caption{Impact of Number of Clients on FL Aggregation Performance}
\label{n_clients_performance}
\end{figure}

Increasing the number of clients, from 10 to 250, as depicted in Fig. \ref{n_clients_acc}, significantly reduces accuracy. In our specific scenario, this decrement leads to a performance drop of up to 70\% with 250 clients. This illustration emphasizes the pivotal role played by the number of clients, which reflects the scale of aggregation a given algorithm can handle effectively. Another vital metric to consider when assessing FL aggregation strategies is the variation in accuracy across participant clients. This measure helps to verify if the global model exhibits robust generalization capabilities, ensuring fair performance without favoring specific clients. Lower accuracy variance indicates a more consistent and fair model. In our scenario, we observed that increasing the number of clients led to higher variance, but there were no significant differences among the three aggregation methods for each client count, as shown in Fig. \ref{n_clients_var}. Notably, our data distribution was non-heterogeneous in this setup. However, the concerns of accuracy variance typically arise when clients possess heterogeneous data and varied computational resources.

In essence, the factor of client number necessitates careful caution when designing new FL aggregation proposals. Hence, researchers must exercise precision when discussing the suitability of an aggregation method initially designed for cross-silo federated learning in the cross-device context. Neglecting this adaptation can result in a pronounced decline in performance, as highlighted by our empirical findings.


\subsubsection{Impact of Data Heterogeneity}
Addressing the issue of data heterogeneity, we subjected the four chosen algorithms to evaluation under non-IID distribution. To rigorously assess severe cases of data heterogeneity and discern how each algorithm navigates this intricate obstacle, we employed a quantity-based label distribution skew partition across the FashionMNIST dataset. This approach allowed us to create scenarios where the total sample number and the number of samples per class on each client showcased significant imbalances, as visually represented in Fig. \ref{quantity_based_label_skew}. Given that each client possessed only a specific number of sample classes, we structured our experimentation around the \textit{number of major classes} parameter, which dictates the number of distinct labels each client could hold, spanning from 2 to 8. Additionally, we set the alpha parameter to 0.3, controlling the extent of imbalance in the Dirichlet distribution across 40 clients.

\begin{figure}[htbp]
\begin{subfigure}{0.33\textwidth}
    \includegraphics[width=\textwidth]{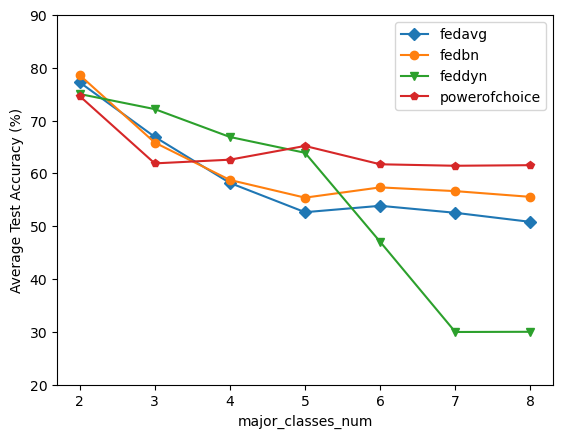}
    \caption{Average Accuracy Variation}
    \label{n_data_dist_acc}
\end{subfigure}
\begin{subfigure}{0.33\textwidth}
    \includegraphics[width=\textwidth]{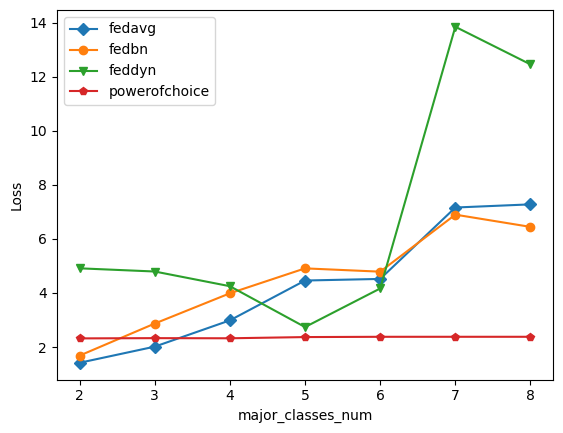}
    \caption{Average Loss Variation}
    \label{n_data_dist_loss}
\end{subfigure}
\begin{subfigure}{0.33\textwidth}
    \includegraphics[width=\textwidth]{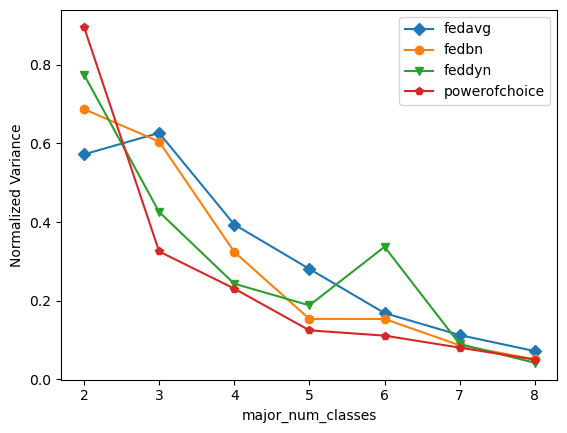}
    \caption{Normalized Variance}
    \label{n_data_variance}
\end{subfigure}
\caption{Impact of Heterogeneous Data Distribution (Quantity-Based Label Distribution Skew) on FL Aggregation Performance}
\label{n_data_dist_performance}
\end{figure}

Fig. \ref{n_data_dist_performance} displays the performance of all assessed algorithms when confronted with escalating levels of data heterogeneity. Visibly, both Fedbn and Powerofchoice demonstrate a more adept ability to navigate this type of heterogeneity, even when the task complexity increases. Conversely, Fedavg and Feddyn experience a significant decline in accuracy, a trend that becomes more pronounced when the \textit{number of major classes} parameter exceeds 5. This disparity in performance can be ascribed to the effectiveness of simple yet influential modifications, such as refined client selection policies (as exemplified in Powerofchoice) or batch normalization technique (as seen in Feddyn). While such enhancements may appear relatively modest in terms of what they bring to the standard Fedavg, they undeniably play a pivotal role in overcoming the formidable challenge posed by data heterogeneity.

As illustrated in Fig. \ref{n_data_variance}, the observed disparity in trends between variance and accuracy stems from variations in data distribution among clients, particularly evident in highly imbalanced class scenarios. For instance, when a client's dataset contains only 2 or 3 classes, data distribution tends to be more diverse, with uneven distribution among those limited classes, resulting in divergent performance levels among clients. Conversely, as the data distribution becomes more balanced with an increase in the number of labels per client, the variance in training accuracy tends to decrease despite potential imbalances in sample distribution across clients. In summary, while high accuracy results may indicate overall performance, they do not account for fairness in individual client contributions. Hence, considering metrics such as variance is essential for a more nuanced understanding of each client's participation in the collaborative process.

\subsubsection{Impact of System Heterogeneity}

To simulate system heterogeneity, with a specific focus on the diversity of device resources, we adopted a methodology similar to that outlined in \cite{li2020federatedprox}. We kept a fixed number of epochs, denoted as E. Then, we created instances where some devices executed fewer updates than the specified E epoch, considering the restrictions imposed by each device's current system capabilities. Specifically, In our experimentation,  we allocate a certain number of epochs, determined uniformly at random from the range of [1, E], and assigned it to a percentage of active devices. To introduce different levels of resource heterogeneity, we progressively varied this percentage from 0\% to 90\% in increments of 10\%. Scenarios where 0\% of devices performed fewer than E epochs represented environments devoid of system heterogeneity. On the other hand, instances where up to 90\% of devices executed partial training conveyed highly heterogeneous settings. 

\begin{figure}[htbp]
\begin{subfigure}{0.35\textwidth}
    \includegraphics[width=\textwidth]{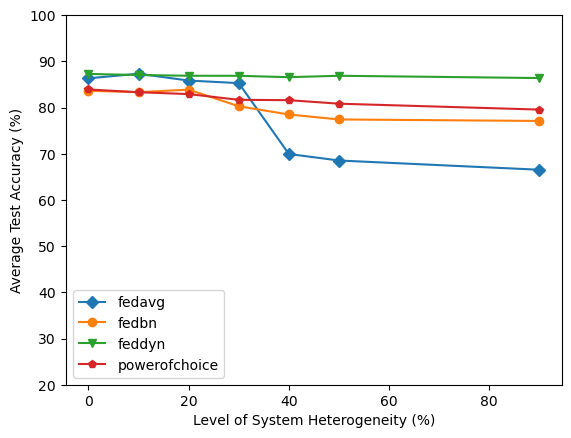}
    \caption{Average Accuracy Variation}
    \label{sys_heterog_acc}
\end{subfigure}
\begin{subfigure}{0.35\textwidth}
    \includegraphics[width=\textwidth]{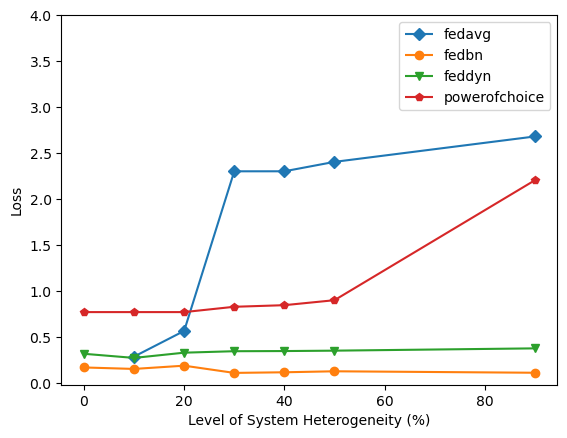}
    \caption{Average Loss Variation}
    \label{sys_heterog_loss}
\end{subfigure}
\caption{Impact of System Heterogeneity on FL Aggregation Performance}
\label{sys_heterog_performance}
\end{figure}

Upon analyzing the outcomes presented in Fig. \ref{sys_heterog_performance}, it becomes apparent that all the evaluated aggregation methods maintain strong performance as we incrementally introduce system heterogeneity from 0\% to 40\%. Nevertheless, a notable decline in Fedavg's performance is observed beyond this threshold, whereas the other algorithms show a modest reduction in average accuracy. This behavior is attributed to Fedavg's limited capacity to handle significant device heterogeneity. Notably, recent aggregation techniques introduced in the literature, such as Fedbn, Feddyn, and Powerofchoice, have proven their ability to accommodate such disparities.

\subsubsection{Impact of DL Hyperparameters}
\begin{figure}[htbp]
\begin{subfigure}{0.35\textwidth}
    \includegraphics[width=\textwidth]{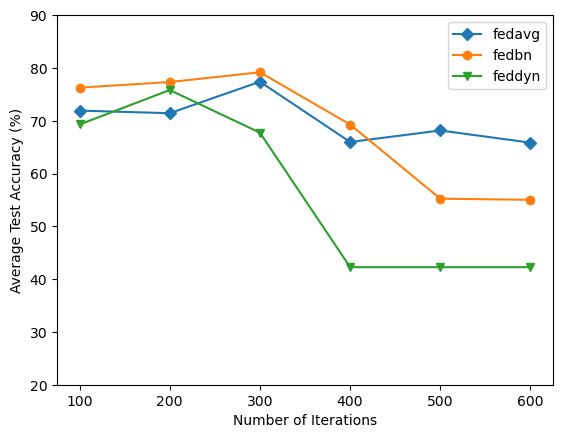}
    \caption{Number of Iterationn}
    \label{n_rounds_performance_acc}
\end{subfigure}
\begin{subfigure}{0.35\textwidth}
    \includegraphics[width=\textwidth]{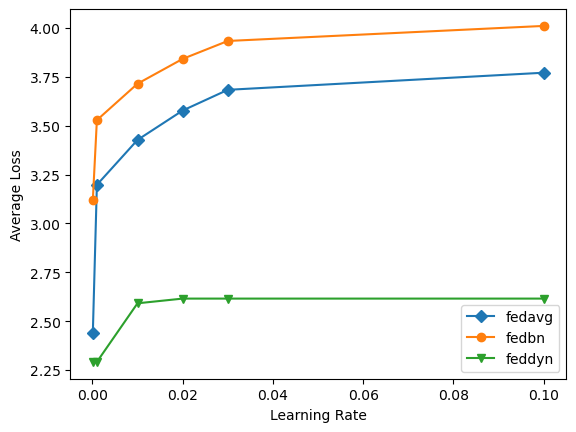}
    \caption{Learning Rate}
    \label{n_lr_performance_loss}
\end{subfigure}
\caption{Impact DL Hyperparameters Tuning on FL Aggregation Performance}
\label{DL_hyperparameter_performance}
\end{figure}

\paragraph{Number of Iteration}
The choice of hyperparameters significantly influences the federated learning process. One critical parameter we put under study is the number of communication rounds or iterations. In our experiment, we utilized the MedMNIST dataset with ten (10) clients, partitioning the data according to a Dirichlet distribution with an alpha parameter of 0.3. We ranged the number of iterations from 100 to 600 and analyzed the results in Fig. \ref{n_rounds_performance_acc}.

The tendencies observed for FedAvg, FedBN, and FedDyn are quite similar. These three algorithms exhibit improved accuracy when increasing the number of iterations from 100 to 200 and subsequently from 200 to 300, which is logical as more iterations allow the local models to learn better from the data. However, after surpassing 300 rounds, performance begins to decline due to overfitting. In other words, the models tend to perform exceptionally well on the training data but struggle with unseen data (test data).

\paragraph{Learning Rate}
The second hyperparameter under investigation was the learning rate, with the same experimental setup as previously described. We varied the learning rate across the range [0.0001, 0.001, 0.01, 0.02, 0.03, 0.1]. As illustrated in Fig. \ref{n_lr_performance_loss}, it becomes evident that as the learning rate increases, the average loss also rises for all algorithms. This finding implies that for this specific scenario, the optimal learning rate tends to be below 0.001. Typically, this value should be carefully tuned based on the task's requirements and dataset characteristics.

\section{Future Directions}
\label{FL_future_directions}

\subsection{Federated Learning-Empowred Large Language Models}
In recent years, the Large Language Models (LLMs) field has witnessed remarkable advancements. Notable LLM models, such as Google's BERT \cite{devlin2018bert}, OpenAI's ChatGPT-3 and ChatGPT-4 \cite{openGPT4}, TII's FlaconLLM \cite{almazrouei2023falcon}, and Meta's LLaMA \cite{touvron2023llama}, excel in understanding human language, engaging in life-like conversations, and generating coherent, contextually relevant responses. These models have left an indelible mark on a multitude of AI applications. However, their development and deployment face significant challenges. On the one hand, access to large-scale and high-quality public data is a bottleneck, often impeded by privacy concerns and fierce commercial competition. Additionally, the demand for outstanding computational resources is insatiable. On the other hand, research has shown that amplifying the scale of LLMs in terms of both model parameters and training data yields substantial performance gains, especially for handling complex tasks \cite{ferrag2023securefalcon}.

In light of these constraints, FL emerges as a beacon of promise, introducing a decentralized AI paradigm that could revolutionize the LLM domain. FL offers a spectrum of benefits spanning over the entire lifecycle of LLMs, from pre-training and fine-tuning to deployment and downstream applications \cite{chen2023federated}. Thanks to its contemporary techniques like parameter-efficient training methods, prompt tuning, and advanced model compression, FL facilitates the distribution of the computational workload across multiple participants. Moreover, ensuring the fluid embodiment of trustworthy FL mechanisms within the LLM ecosystem is pivotal for encountering limitations tied to privacy, security, robustness, and bias concerns \cite{zhuang2023foundation}. Therefore, these hot research topics, at the crossroads of FL and LLMs, hold tantalizing prospects for exploration and practical implementation. Unleashing the potential of this harmonious collaboration promises to unlock a trove of impressive advantages \cite{zhang2023gpt}.

\subsection{Federated Learning for 6G and Beyond Technologies}
In the wake of 5G's transformative impact on the networking landscape, as it continues to be deployed globally, the spotlight is now on the next-gen wireless revolution, i.e., Sixth-generation (6G) \cite{letaief2019roadmap}. As its core, 6G promises to deliver ubiquitous, seamless, and high-performance connectivity, all while prioritizing security and privacy. The ultimate aim is to mark a shift from "\textit{connected things}" to "\textit{connected intelligence}" in modern network systems \cite{dang2020should}. Furthermore, driven by Industry 5.0 principles and advanced AI technologies, 6G is poised to introduce groundbreaking benefits in the ears of new media, new services, and new infrastructure \cite{liu2020federated}. 

In this context, FL emerges as a captivating avenue of exploration, bearing immense potential for empowering 6G with AI capabilities, yet FL is still in the nascent stages and confronts novel challenges when embedded into 6G scenarios. One such issue lies in communication efficiency, as the current infrastructure strains to support FL-enabled 6G across a wider geographic area, necessitating the development of innovative, communication-efficient methodologies, practically, for cross-device FL. Besides, the security and privacy of FL within the 6G ecosystem demand vigilant attention. Robust aggregation techniques, tailored defense strategies, and the creation of heterogeneity-tolerant environments are imperative to ensure the success of FL in the 6G era, guarding against numerous vulnerabilities exploited by malicious FL actors.
\subsection{Federated Learning Integrated Digital Twin Systems}
A digital twin (DT) is a virtual emulation of a physical system, mirroring its elements and dynamics in real-time. The fundamental idea behind DT is to forge a digital system replica, seamlessly connected to its real-world counterpart through bidirectional links. Consequently, this simulation ensures that the DT remains a precise and up-to-date representation of its physical system \cite{khan2022digital}. In contexts such as IoT-enabled 6G systems, DT plays a pivotal role in achieving low latency, reliable connectivity, high performance, and energy efficiency. Moreover, when paired with data analytics and ML techniques, DT facilitates the adept management of complex systems and enhances decision-making processes for flight systems, the military, smart cities, healthcare, and others. However, integrating DT modeling into IoT and edge computing applications poses notable challenges. Firstly, DT relies on extensive distributed data, a feat hindered by privacy concerns, rendering data amalgamation from diverse devices nearly unattainable. Secondly, the imperative real-time interplay between DTs and their corresponding entities requires frequent device communication. Yet, conventional centralized ML network architectures often falter in this regard \cite{yang2022optimizing}.

Acknowledging the distinctive merits of FL, including its decentralization, privacy preservation, data wealth, efficiency, and security aspects, some researchers have ventured into employing FL to construct DT models. Nevertheless, the current literature lacks a comprehensive examination of this promising fusion of FL and DT. As DT is emerging as one of the most influential technologies in the coming decades \cite{al2023edge}, researchers are encouraged to investigate the potential synergy between FL and DT, unveiling numerous possibilities for mutual empowerment.
\subsection{Federated Meta-Learning}
In response to the shortcomings of data-hungry ML techniques, researchers are actively exploring novel approaches, particularly few-shot learning, to reduce data requirements while enhancing model performance. In effect, few-shot learning presents a compelling method that trains models to rapidly adapt to new tasks using minimal data and fewer training iterations \cite{griva2023model}. Meta-learning, often known as "\textit{learning how to learn}",  has gained popularity in pursuit of this goal. In contrast to standard AI approaches that solve tasks from scratch using fixed learning algorithms, meta-learning seeks to improve the learning algorithm itself, leveraging insights gained from previous learning experiences. In other words, it involves distilling knowledge from multiple learning episodes, often spanning various related tasks, to improve future learning performance. This results in improved data efficiency, better knowledge transfer, and enhanced unsupervised learning when training DL models \cite{gharoun2023meta}. 

Integrating meta-learning into the FL framework brings numerous gains. Remarkably, federated meta-learning swiftly adapts to new heterogenous tasks even with small datapoints, all while conserving computational resources and training time. Moreover, as exemplified in \cite{finn2017model}, the emergence of optimization meta-learning algorithms, designed to learn and fine-tune FL-related components like selected clients and regularization terms, represents a promising area of research. To fully harness the potential of this dynamic field and its related areas, such as domain adaptation and domain generalization \cite{hospedales2021meta}, further research is essential to unlock the myriad opportunities they offer within the FL.
\subsection{Multimodal and Dynamic Federated Learning}
In current federated learning approaches, there is an implicit assumption that client data remains static and unaltered. However, this premise frequently diverges from reality. Users often operate in dynamic environments where local data continuously evolve due to sensor observations, resulting in the incremental inclusion of new classes into the training data. When confronted with this scenario, conventional training and aggregation methods struggle with a challenge known as "\textit{catastrophic forgetting}." This phenomenon results in a significant decline in overall performance as new classes are introduced incrementally. Traditional NN models require the entire dataset, including old and new class samples, to be available during training. Wearers, this requirement becomes rapidly impractical as the number of classes expands. An ideal approach to address this issue is \textit{incremental learning}, as referred to in the literature. The core concept of incremental learning is to train incrementally on an infinitely expanding set of classes while maintaining accuracy and the same number of model parameters \cite{castro2018end}. Consequently, the combination of federated learning and incremental learning becomes an intriguing area for investigation \cite{liu2023recent}.

Another set of assumptions about datasets used in FL settings are unimodal (containing a singular data type: image, audio, or tabular inputs) and completely labeled (comprising readily available labeled data in the standard format). However, many practical application domains involve clients with multimodal data \cite{lin2023federated} and limited access to ground-truth labels \cite{ding2022federated}. Dealing with data incompleteness, its dynamic and multimodal aspects in FL remains largely unexplored but holds significant promise. Therefore, research efforts in these directions can bring innovative solutions that benefit the broader federated learning community.

\section{Conclusions}
\label{FL_conclusions}
In this survey, we concentrate on three primary clusters of contributions around the federated learning paradigm: personalization, optimization, and robustness. Within this high-level classification, we dissect our investigation to delve into the challenges and potential solutions found in the literature. Consequently, we propose multi-level and well-structured classification schemes that better organize the content within the classes and sub-classes of each cluster, resulting in six (06) distinct taxonomy schemes with up to four (04) layers each. 
Specifically, we examine various facets of federated learning, including diverse types of heterogeneity, efficiency, security, and privacy concerns. Then, we down-break the contemporary strategies addressing these topics into three-level taxonomies, mainly focused on aggregation methods. Additionally, we illustrate the exploration of these promising strategies with recent work published over the past three years. Distinguishing our survey from others is our hybrid methodology for selecting state-of-the-art papers. We combined bibliometric analysis using CiteSpace to understand the trends and dynamics of FL with systematic scrutiny to include only the most relevant and exceptional work. These scholarly publications illustrate the myriad application areas of FL, including healthcare, industry, robotics, and recommendation systems. Additionally, they demonstrate the emergence of FL coupling with other cutting-edge technologies, such as blockchain, edge, fog, IoT, and many more. To facilitate efficient content navigation, we have evaluated and summarised the features of most surveyed literature in informative yet concise tables with a total of 14 criteria corresponding to two distinct families (i.e., proposal environment and verified goals). The resulting nine (09) tables encompass 85 carefully selected papers. We further perform extensive simulations to evaluate the performance of four popular aggregation algorithms, including FedAvg, FedBn, FedDyn, and Powerofchoice, across six (06) real-world scenarios. Our experimental findings align closely with our theoretical analysis regarding scalability, generalization, fairness, statistical heterogeneity, system heterogeneity, and DL hyperparameter tuning. To conclude our navigation in the FL landscape, we present a compelling set of future research directions, encouraging fellow researchers to dive deeper into these captivating areas of investigation.

In summary, we believe that our survey is the most comprehensive in its coverage of challenges and techniques. Simultaneously, it carefully structures the content with a high degree of hierarchical organization to assist researchers in navigating specific topics of their interest. Moreover, it pioneers an aggregation-centric approach to the FL domain and furnishes detailed guidelines for evaluating novel proposals within practical settings.

\bibliographystyle{unsrt}
\bibliography{sample-base}

\end{document}